\documentclass[final]{cvpr}

\usepackage{times}
\usepackage{epsfig}
\usepackage{graphicx}
\usepackage{amsmath}
\usepackage{amssymb}
\usepackage{Inputs/packages}

\def\confYear{CVPR 2021}

\begin{document}

\title{\SHORTTITLE: \TITLE}

\author{Andr\'es Romero\textsuperscript{1} \qquad Luc Van Gool\textsuperscript{1,2} \qquad Radu Timofte\textsuperscript{1} \\
\textsuperscript{1}Computer Vision Lab, ETH Z\"urich \qquad \textsuperscript{2}KU Leuven\\
{\tt\small \{roandres, vangool, timofter\}@vision.ee.ethz.ch}
}

\maketitle
\newcommand{\aref}[1]{Please refer to Appendix~{#1}}
\begin{abstract}
Attribute image manipulation has been a very active topic since the introduction of Generative Adversarial Networks (GANs). Exploring the disentangled attribute space within a transformation is a very challenging task due to the multiple and mutually-inclusive nature of the facial images, where different labels (eyeglasses, hats, hair, identity, etc.) can co-exist at the same time. Several works address this issue either by exploiting the modality of each domain/attribute using a conditional random vector noise, or extracting the modality from an exemplary image. However, existing methods cannot handle both random and reference transformations for multiple attributes, which limits the generality of the solutions. In this paper, we successfully exploit a multimodal representation that handles all attributes, be it guided by random noise or exemplar images, while only using the underlying domain information of the target domain. We present extensive qualitative and quantitative results for facial datasets and several different attributes that show the superiority of our method. Additionally, our method is capable of adding, removing or changing either fine-grained or coarse attributes by using an image as a reference or by exploring the style distribution space, and it can be easily extended to head-swapping and face-reenactment applications without being trained on videos.
\end{abstract}
\section{Introduction}

In this paper we tackle the problem of adding, removing or manipulating facial attributes for either exemplar images or random manipulations, using a single model. For instance, given a person A, our system could aim at imposing the haircut of person B, eyeglasses of person C, hat of person D, earrings of person E, and randomly changing the background and the color of the hair. Particularly, the problem of manipulating multiple attributes has been coined `multi-domain image-to-image (I2I) translation'~\cite{liu2019stgan,choi2017stargan,romero2019smit}.

Image-to-image translation methods have been traditionally categorized in two groups: latent and exemplar approaches. Latent approaches~\cite{zhu2017bicycle,almahairi2018augmentedcyclegan,chang2018pairedcyclegan} require sampling from a distribution in order to perform a cross-domain mapping, that is, to explore the underlying latent distribution and produce multimodal representations given a single input. Conversely, exemplar-based approaches~\cite{wang2019controlling,xiao2018elegant,guo2019mulgan} require an additional image to condition the generation. There have been some efforts~\cite{huang2018munit,DRIT++,choi2019starganv2} that tried to reconcile the latent and exemplar approaches in a single and unified system. However, they consider independent and inter-class diverse domains.

\begin{figure}[t]
\begin{center}
    \includegraphics[width=\linewidth]{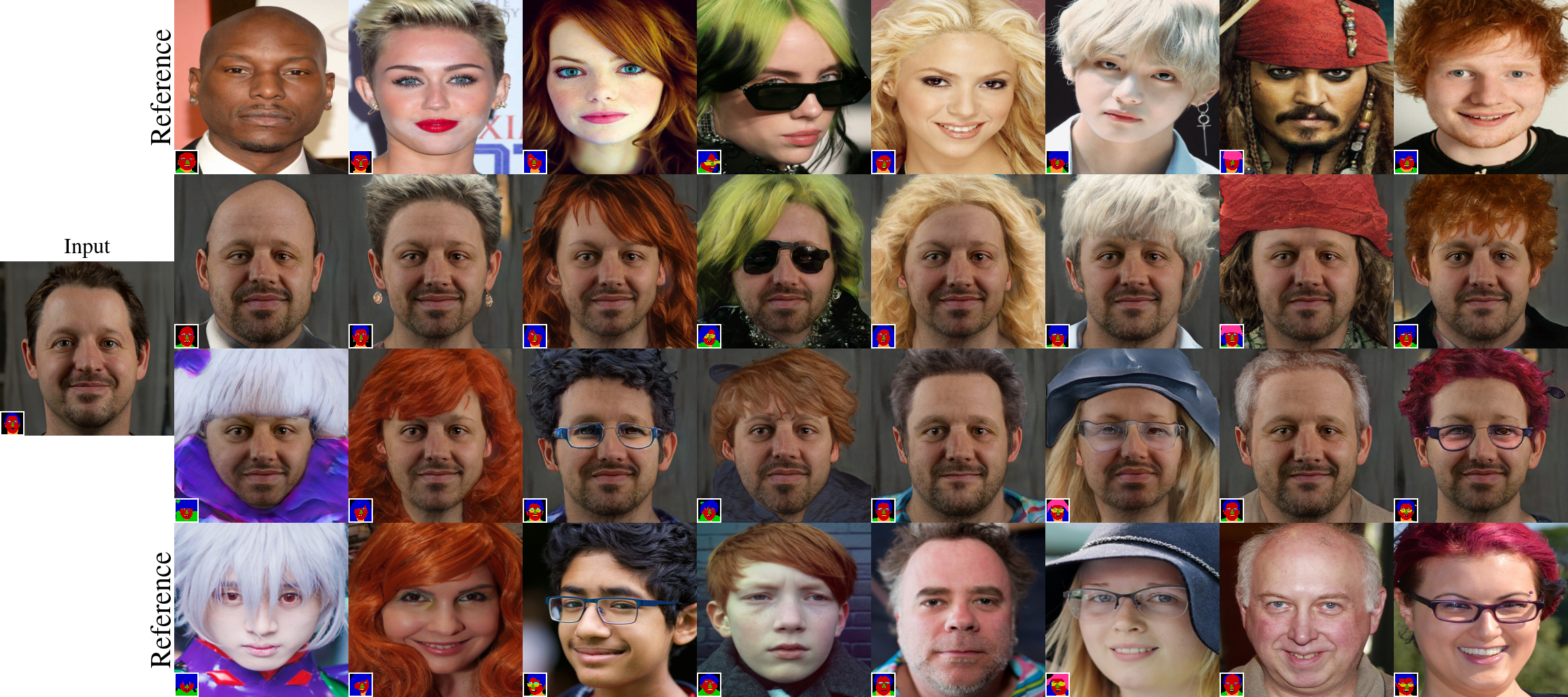}
\end{center}
   \caption{Our model learns a full diverse manipulation for multiple attributes using a single generator and keeping the identity of the input. Note that the reference image does not require to be aligned with respect to the input. Please zoom in for better details.}
\label{figure:teaser}
\end{figure}

Regarding facial manipulation, I2I translation approaches come with the additional constraint that some regions of the image (\eg, background, clothes, etc) or fixed characteristics of the face (\eg, eyeglasses, hats, etc) should remain unaltered during the transformation. Vanilla CycleGAN~\cite{zhu2017cyclegan}-based approaches traditionally alter the general content and shift the colors of the input. To overcome this undesired property, latent generative approaches~\cite{liu2019stgan,he2019attgan,wu2019relgan} have proposed attention mechanisms~\cite{pumarola2018ganimation}, performing architectural changes and introducing tailored loss functions into the training framework, thus obtaining impressive results. Nevertheless, the transformations are mostly fine-grained and do not perform well for more global transformations such as a change of identity or from short to long hair. Reference guided methods~\cite{zhou2017genegan,benaim2019domain,xiao2018dna}, on the other hand, either work on low resolution scales or focus on the same local texture transformation as in latent approaches. Recently, StarGANv2~\cite{choi2019starganv2} was proposed as a variant for multi-domain I2I translation. Nonetheless, it requires that the multiple domains are not activated at the same time and it does not perform well for fine-grained transformations.

In order to solve the aforementioned issues, we propose \TITLE (\SHORTTITLE). With \SHORTTITLE, we split the solution of this problem into two stages. Instead of dealing with complex general and local transformations in the RGB space, we first simplify the attribute manipulation by performing it in the semantic segmentation space. Second, as the manipulation happens there, an additional stage consists of driving the image synthesis via semantics in order to produce photo-realistic RGB faces.

We enumerate our contributions as follows:
\begin{enumerate}
    \item We propose a multi-attribute I2I transformation method for both fine-grained and more global attributes in the semantic space for both random and exemplar-based synthesis.
    \item We propose an extended version of StyleGAN2~\cite{karras2019style} to deal with semantic masks and per-region-styles either to perform random or exemplar-based synthesis.
    \item Even though we train our system with still images, we qualitatively show that by only extracting the style matrix of a single target image, it suffices to create a controlled-by-attribute head-swapping or face-reenactment video using a video in the wild frame by frame.
\end{enumerate}

\begin{figure}[t]
\begin{center}
    \includegraphics[width=\linewidth]{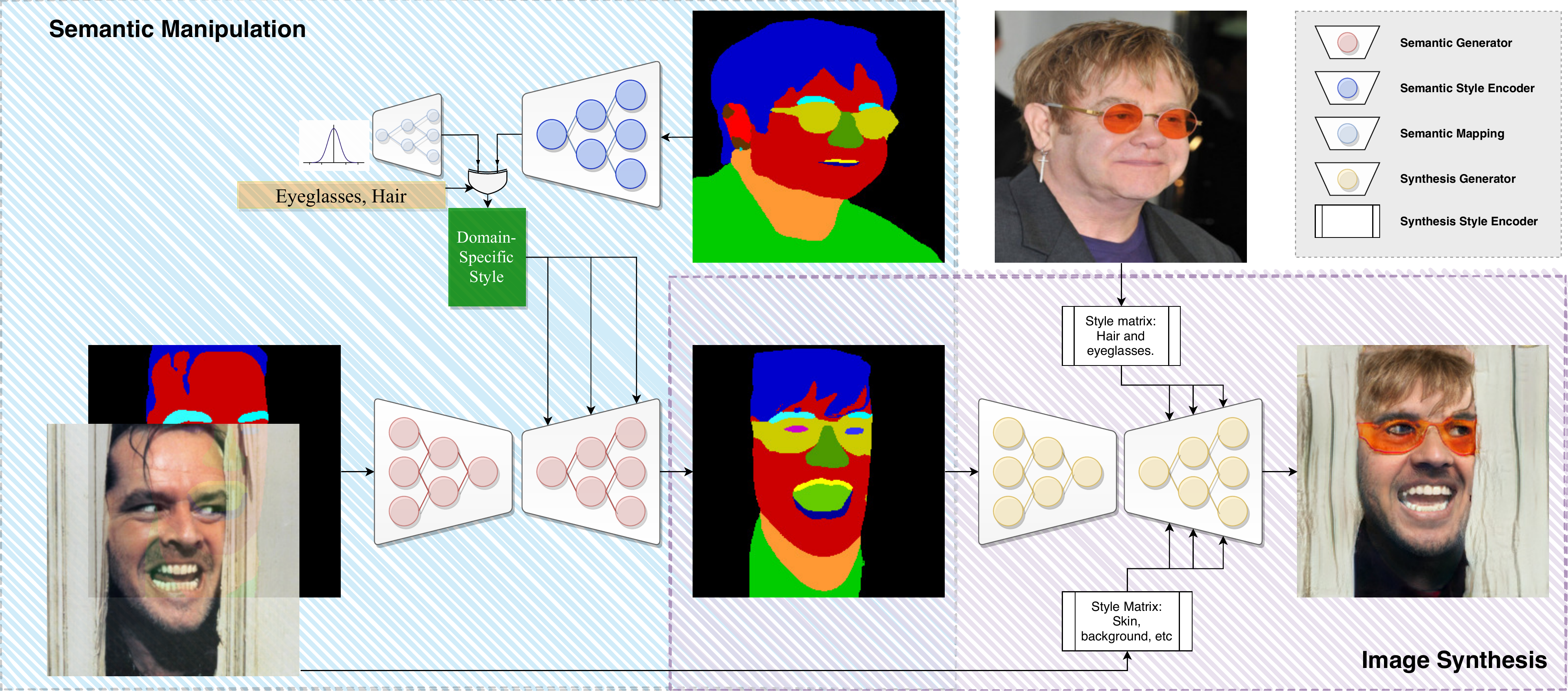}
\end{center}
   \caption{\textbf{Overview of \SHORTTITLE.} We translate an image by either taking as input a random style or target attributes into the generator (we use a reference image in this example). We first manipulate the semantic segmentation map, and then we synthesize the output using both input and reference styles to produce a photo-realistic merge of the two images.}
\label{figure:overview}
\end{figure}

We depict diverse facial manipulations in~\fref{figure:teaser}, and an overview of our system in~\fref{figure:overview}. Code source and pre-trained models can be found in \url{https://github.com/affromero/SMILE}.

\section{Related Work}
\label{section:related_work}
Recently, Image-to-image (I2I) translation has become a very active topic thanks to the impressive advances in generative modeling methods, and in particular, Generative Adversarial Networks (GANs)~\cite{goodfellow2014generative}. Several novel and challenging problems have been successfully tackled with this technique, \eg, multi-domain manipulation~\cite{choi2017stargan,romero2019smit}, style transferring~\cite{huang2017arbitraryadain,lin2020distribution}, image inpainting~\cite{yu2019free,ntavelis2020sesame}, image synthesis using semantic segmentation~\cite{park2019SPADE,zhu2020sean,lee2020maskgan}, image content manipulation
~\cite{park2020swapping}, exploratory image super-resolution~\cite{menon2020pulse,buhler2020deepsee}, etc.

\subsection{Facial Attribute Manipulation}
Since the face is one of the most common, yet interesting models, facial image editing has gained traction over the years~\cite{viazovetskyi2020stylegan2,shen2020interpreting,bau2020rewriting}. 
Different works~\cite{choi2017stargan,pumarola2018ganimation,lample2017fader,guo2019mulgan} have included facial attribute information as condition in GANs to manipulate eyeglasses, mouth expression, hair and other attributes with remarkable results. Due to the not mutually-exclusive representation of facial attributes, multi-domain methods have received quite some attention as a unified and flexible way to deal with several domains. Nevertheless, modeling each attribute as a domain requires having a fully disentangled understanding for each attribute. We developed a system that (\textit{i}) explores the modal representation in the latent space and (\textit{ii}) allows for exemplar imposition, \eg, imposing someone else's eyeglasses or hair. Recently, there have been some efforts~\cite{he2019attgan,romero2019smit,yu2019dmit} elucidating the former, or the latter~\cite{guo2019mulgan}, yet the combination had not been achieved yet. 

We refer to facial object transfiguration as the problem of extracting some specific information of person A's face and transplant it on person B's face, \eg, make-up, eyeglasses, smile, hair, etc. There are traditionally two different groups of methods doing this: makeup transferring~\cite{chang2018pairedcyclegan,wang2019controlling} and attribute manipulation~\cite{zhou2017genegan,benaim2019domain,mokady2019mask,xiao2018dna,xiao2018elegant,guo2019mulgan}. Make-up transferring methods focus on localized texture mappings, whereas attribute imposing methods are traditionally modeled as binary problems using the presence or absence of a selected feature. While the former allow for high resolution transformations and require exemplar images, the latter normally operate at low resolution due to the intricate representations of multiple attributes in the RGB space. 

Recently, StarGANv2~\cite{choi2019starganv2} was introduced as an alternative for multi-domain facial attribute manipulation for both random sampling and reference guidance. Nonetheless, as we discuss in~\sref{sec:results} the generalization capabilities of StarGANv2 are compromised when training with different and/or additional domains to Male/Female, and it also wanting when it comes to fine-grained transformations. Moreover, it was designed for mutually-exclusive domains such as Male/Female and different kinds of animals.

Our method combines the best of the two worlds by using specific attribute imposition from exemplar images, or exploring the latent space using a fully disentangled representation.

\subsection{Semantically-guided manipulation}

Using semantic information for image synthesis is an emerging field, in which using the semantic segmentation as input, it aims at producing an RGB image that perfectly resembles the semantic regions in the input. Semantic manipulation allows finer control of the resulting image just by adjusting the input. To this end, inspired by pix2pixHD~\cite{wang2018pix2pixHD}, SPADE~\cite{park2019SPADE} introduced a specialized and spatially-driven normalization block in order to deal with the different masks in an up-sampling manner, producing impressive results for high-resolution synthesis. However, one critical issue about SPADE is the lack of control for each resulting semantic region. Recently, SEAN~\cite{zhu2020sean} and MaskGAN~\cite{lee2020maskgan} modeled independent style representations for each semantic region, and in the same vein as SPADE, they introduce the semantic and style information combined in a $W$ space distribution through adaptive normalization layers in the generator. 
It is worth to mention that both SEAN~\cite{zhu2020sean} and MaskGAN~\cite{lee2020maskgan} require an exemplar image to perform the generation, and this is particularly critical for attribute imposition as we would like to generate new content (\eg, hat, eyeglasses, etc) that can be hard to find in a dataset.

To the best of our knowledge, there are two methods that aim at performing image manipulation via semantic mapping: MaskGAN~\cite{lee2020maskgan} and SegVAE~\cite{cheng2020controllable}. MaskGAN uses a Variational AutoEncoder~\cite{larsen2015vaegan} approach to perform manipulations in an unsupervised way, \ie, to perform interpolation between two semantic faces to strengthen the image synthesis. Similarly, using a VAE-based approach SegVAE performs more complex transformations such as adding and removing attributes. It starts from an empty canvas that it progressively fills in the desired regions of attributes. This happens in a cascade fashion, that is, it generates semantic images rather than manipulating existing ones. Our method is very different from MaskGAN and SegVAE, as we can manipulate attributes of real semantic maps with both a latent space or images as reference. Moreover, we see SegVAE as orthogonal to ours, since it could generate a semantic map, which can be consequently manipulated by our proposed system.

\SHORTTITLE{} is akin to both SEAN and MaskGAN, yet by leveraging StyleGAN2~\cite{karras2020analyzing} we extend the $W$ latent distribution towards virtually any kind of style per semantic region. We accomplish this by replacing the normalization layers in the generator with semantically adaptive convolutions (SACs), and by using an alternative training scheme for both random generation (similar to StyleGAN2) and exemplar-guided generation (similar to SEAN).

In summary, \tref{table:related_work} provides a comparison of \SHORTTITLE{} with prior image-to-image translation methods.

\begin{table}[t]
\begin{center}
\resizebox{\linewidth}{!}{
\begin{tabular}{ccccc}
\hline
\multirow{ 2}{*}{Methods}           & Latent-guided  & Image-guided & Mutually-inclusive & Fine-grained \\
              & synthesis & synthesis & domains & mapping \\
\hline
\hline
CycleGAN~\cite{zhu2017cyclegan}    &  \xmark  & \xmark  & \xmark & \xmark \\
StarGAN~\cite{choi2017stargan}     &  \xmark  & \xmark  & \cmark & \cmark \\
DRIT++\textsubscript{\& alike}~\cite{DRIT++,huang2018munit,ma2019exemplar}     & \cmark & \cmark & \xmark  & \xmark  \\
GeneGAN\textsubscript{\& alike}~\cite{zhou2017genegan,benaim2019domain,mokady2019mask}  & \xmark & \cmark & \xmark & \cmark \\
MulGAN\textsubscript{\& alike}~\cite{guo2019mulgan,xiao2018elegant}    & \xmark & \cmark & \cmark & \cmark  \\
StarGANv2~\cite{choi2019starganv2}  & \cmark & \cmark & \xmark & \xmark \\
\hline
\hline
\textbf{\SHORTTITLE{}~(ours)} & \color{green!40!black}{\cmark} & \color{green!40!black}{\cmark} & \color{green!40!black}{\cmark} & \color{green!40!black}{\cmark} \\
\hline
\end{tabular}
}
\caption{Feature comparison with state-of-the-art approaches in I2I translation. \SHORTTITLE~successfully performs both latent-guide and image-guide attribute transformations for fine-grained or more global mappings in a mutually inclusive domain manner. 
}
\label{table:related_work}
\end{center}
\end{table}

\begin{figure*}[t]
\resizebox{\linewidth}{!}{
\begin{tabular}{c||cccc}
    \includegraphics[width=0.2\textwidth]{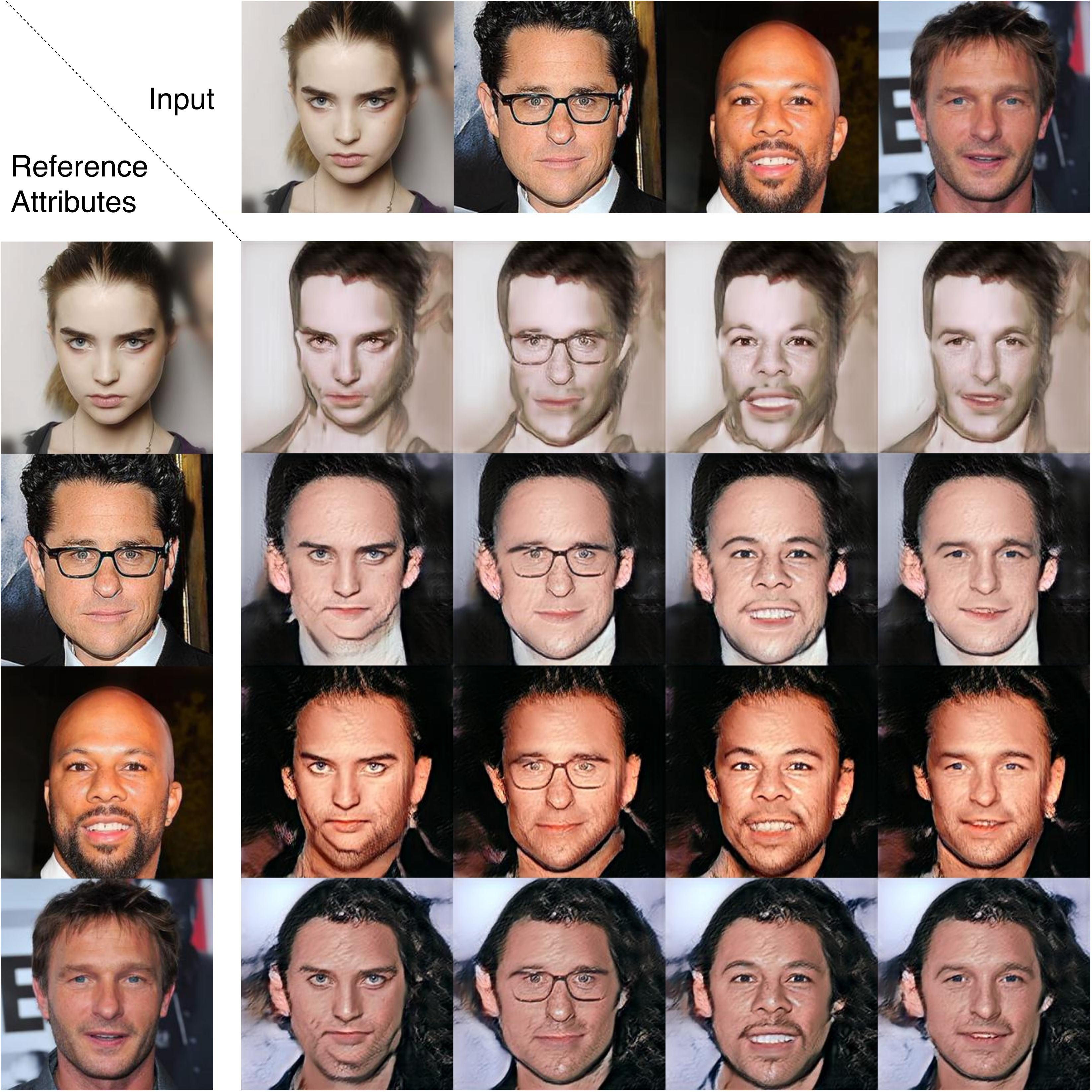} &
    \includegraphics[width=0.2\textwidth]{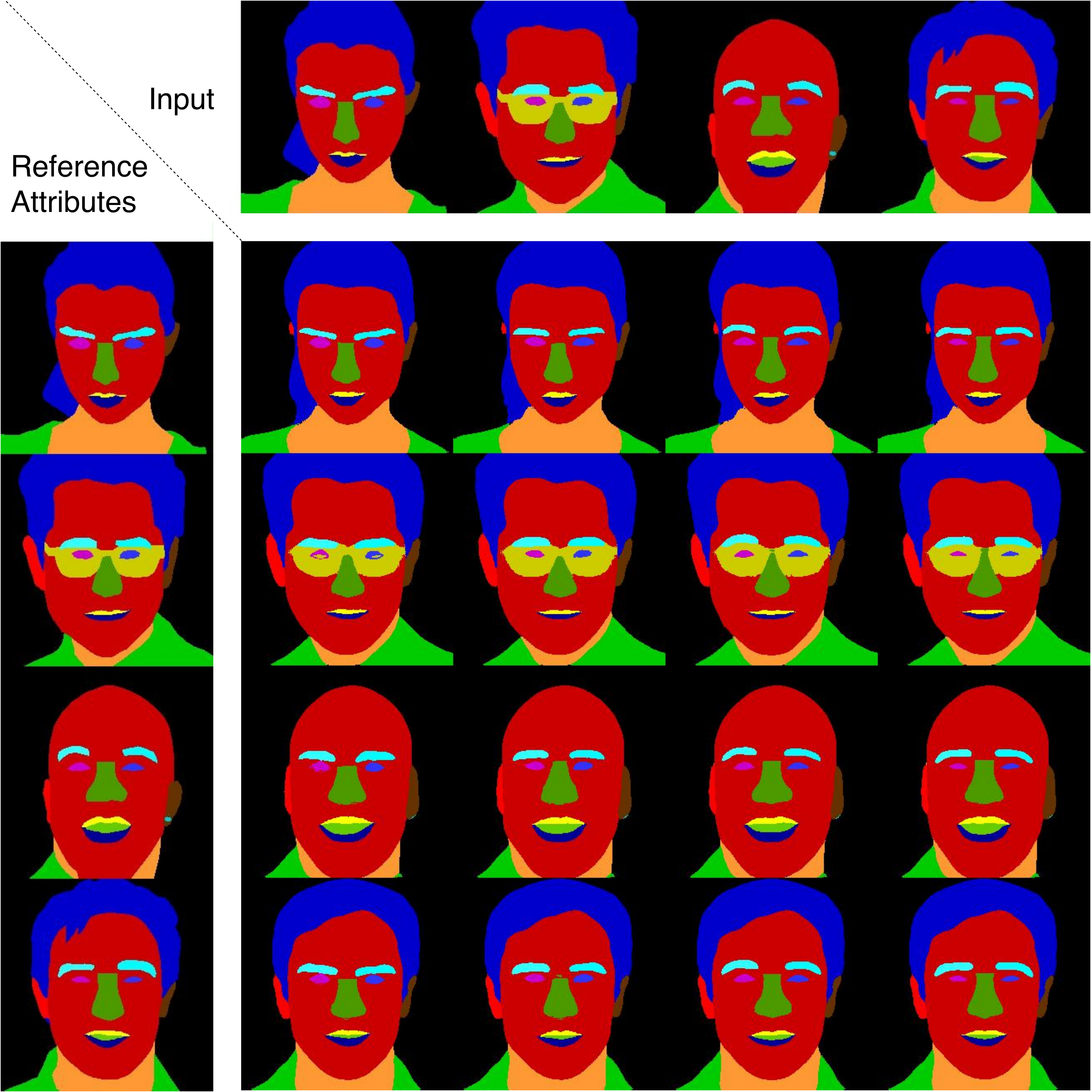} & 
    \includegraphics[width=0.2\textwidth]{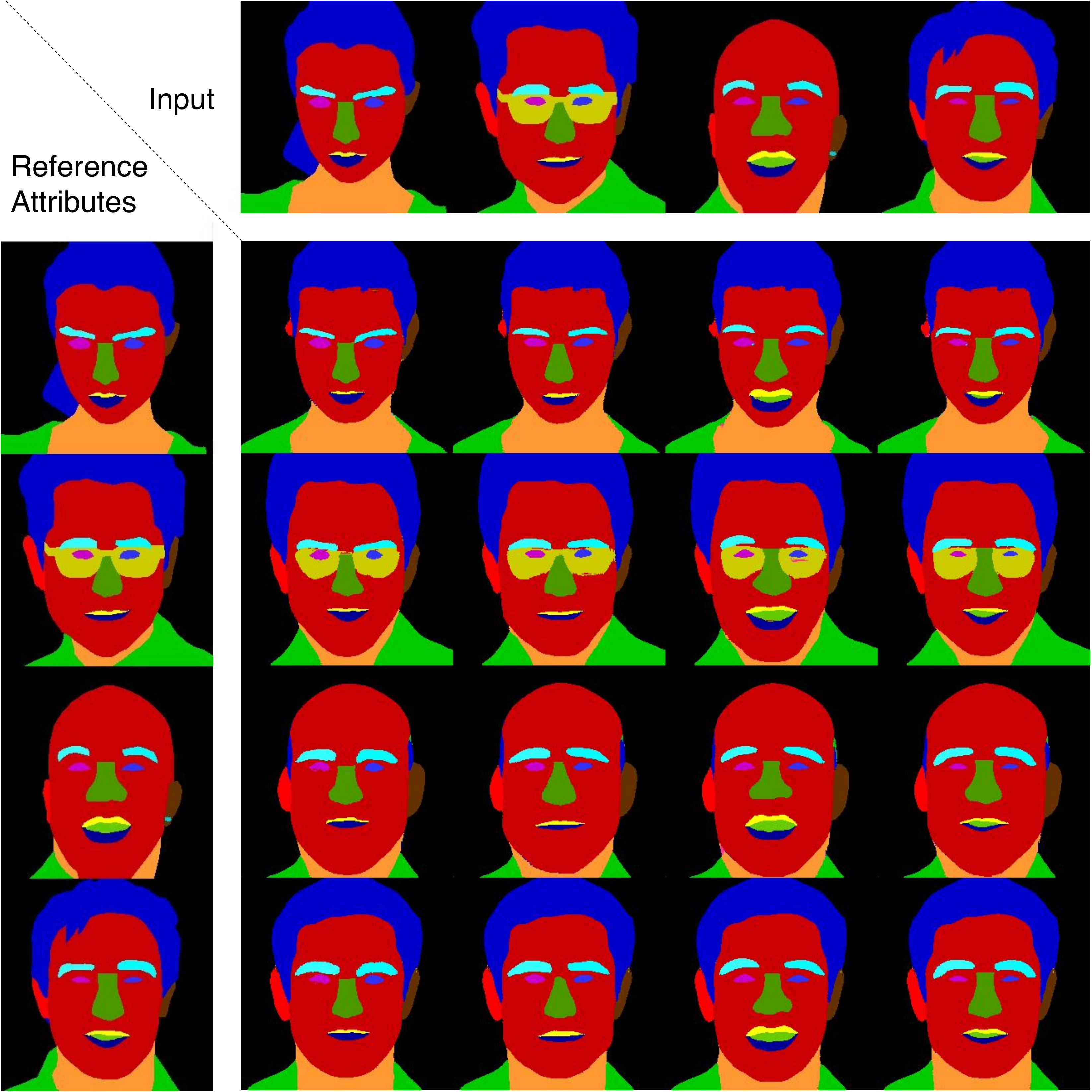} & 
    \includegraphics[width=0.2\textwidth]{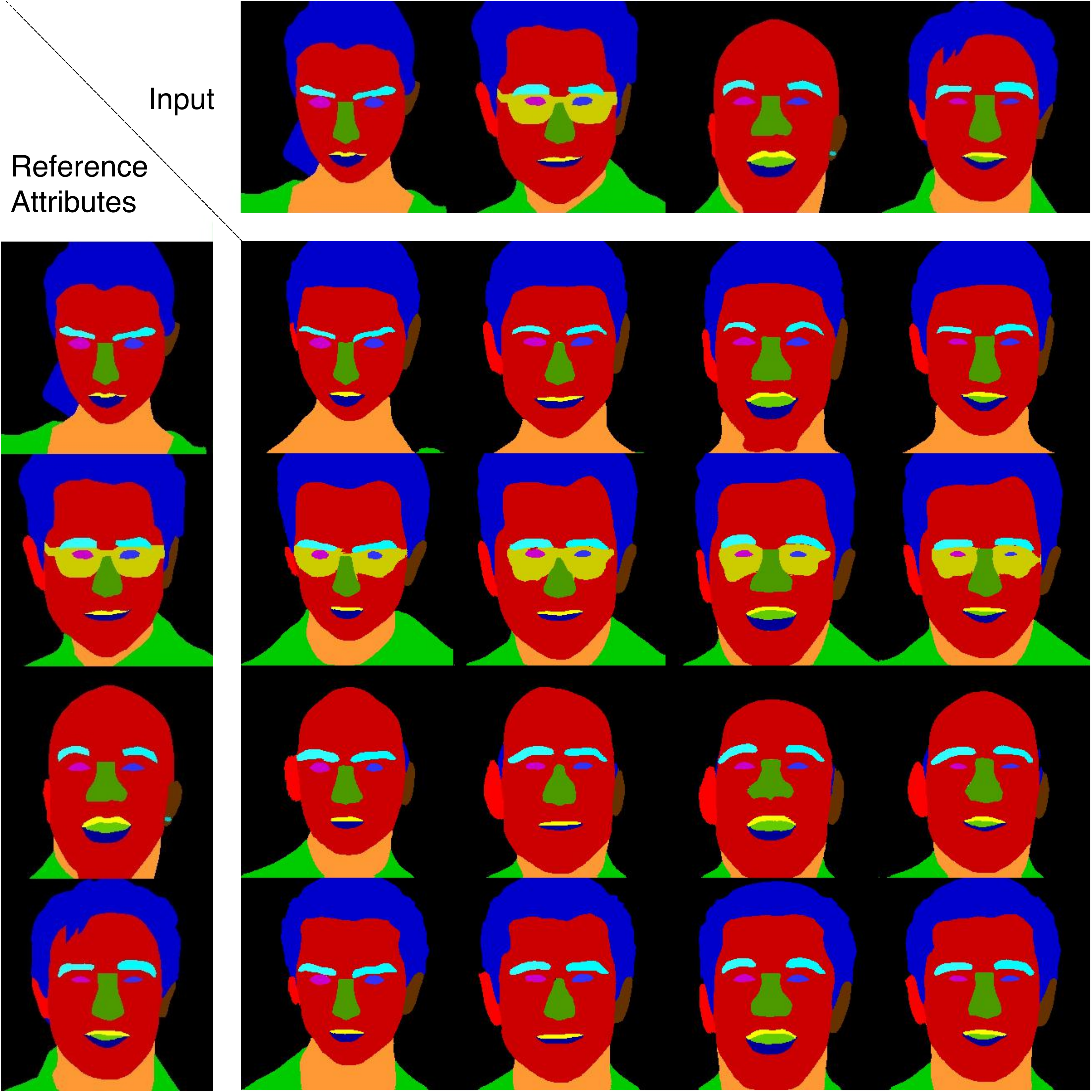} &
    \includegraphics[width=0.2\textwidth]{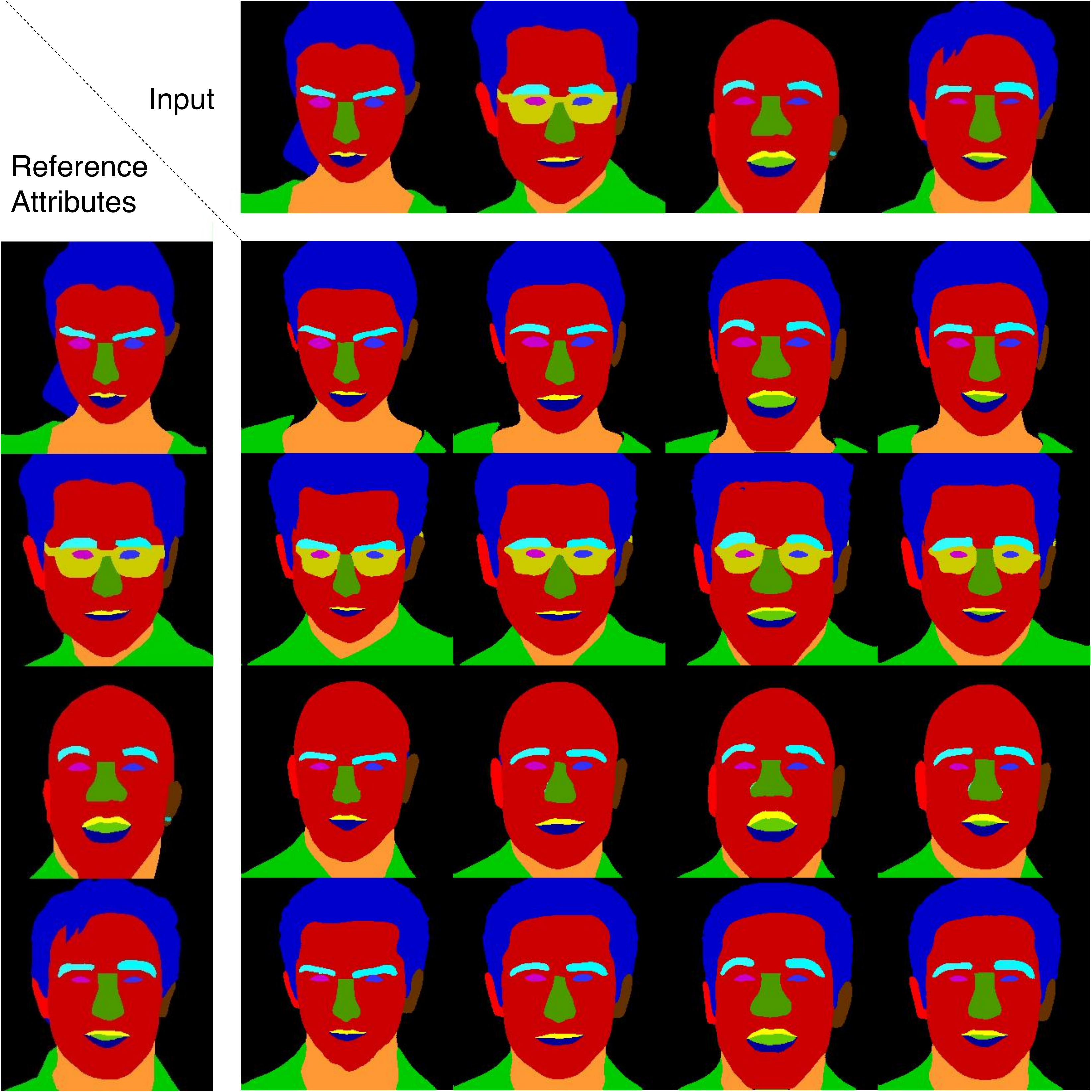}
    \\
    Baseline & Ablation \textit{A} & Ablation \textit{B} & Ablation \textit{C} & Ablation \textit{D}
\end{tabular}
}
\caption{Qualitative results of the ablation experiments for semantic manipulation. We use reference images to perform attribute transformation, \ie, for this visualization we transfer all the reference attributes to the input. Please zoom in for detailed assessment.}
\label{figure:ablation_sem}
\end{figure*}

\section{Proposed Approach}
Our main objective is to perform specific or global style-guided transformations using only the domain information as supervision. We argue that multi-domain exemplar style imposition (\eg, wearing someone else's sunglasses or hair replacement) is a very challenging problem, which normally is simplified by assuming mutually-exclusive domains~\cite{choi2019starganv2}, or one model per domain~\cite{benaim2019domain}. To this end, without simplifying this problem and inspired by recent developments in image synthesis~\cite{karras2020analyzing}, we develop our strategy in two stages: Semantic Manipulation and Improved Region-wise Semantic Synthesis.

\subsection{Semantic Manipulation}
First, we build a simple yet powerful multi-domain I2I translation model using the semantic map. We rely on the semantic information as it is simpler and rich enough to spot and transform noticeable facial attributes like eyeglasses, hats, earrings, hair, bangs and identity.

We build on top of StarGANv2~\cite{choi2019starganv2}, and as depicted in \fref{figure:overview} (left), our system is composed of several networks. In order to perform unsupervised fine-grained and more global translations, we rely on several key assumptions further developed in \sref{sec:model}.

\subsubsection{Model}
\label{sec:model}

Let $\mathcal{X}^{r}\in\mathbb{R}^{H\times{W}\times{M}}$ be the real image with $M$ semantic channels, for instance a mask with a parsing of the face where each channel represents different regions. $\mathcal{X}^{r}$ has associated $N$ attributes $y^{r}\in \mathbb{N}^{\{0,1\}}: \{y_{0}^{r}, \cdots y_{i}^{r}, \cdots y_{N-1}^{r}\}$. Importantly, we assume that for each possible attribute $y^{r}$ in $\mathcal{X}^{r}$, there is one style associated $s^{r}\in\mathbb{R}^{S}: \{s_{0}^{r}, \cdots s_{i}^{r}, \cdots s_{N-1}^{r}\}$. In detail, if for any given image $x^{i}$ that has 4 labels $y^{i} = (1,1,0,1)$, then each attribute can have different shape and color, namely style, so there are 4 style vectors ($\langle{s_{i}}\rangle$): $s^{i} = (\langle{s_{0}}\rangle,\langle{s_{1}}\rangle,\langle{s_{2}}\rangle,\langle{s_{3}}\rangle)$, one latent representation for each label. Note that we assume that the absence of an attribute is also associated with a style distribution, and not as a deterministic zero vector. 

Our purpose is to use the domain information as guidance for the style imposition. Particularly, for each domain we assume there is a style distribution associated to presence and a different style distribution associated to absence. Consequently, we can perform transformations ($\hat{X}$) in both directions: first using an image as reference by extracting the style from the style encoder ($\mathbb{S}$), and second, sampling from the style distribution and processing it through the mapping network ($\mathbb{F}$). Formally, we define these two transformations in \eref{equation:guidance} and \ref{equation:sampling}, respectively.

\begin{gather}
    \hat{X}_{guided} = \mathbb{G}(X^r, \thinspace \mathbb{S}(X^{ref})_{\hat{y}})
    \label{equation:guidance}
    \\
    \hat{X}_{random} = \mathbb{G}(X^r, \thinspace \mathbb{F}(\mathcal{N}(0,I))_{\hat{y}}),
    \label{equation:sampling}
\end{gather}
Where $\mathcal{N}(0,I)$ is a random vector sampled from the normal distribution. Note that these transformations require the selection of the presence or absence of domains ($\hat{y}$) in each style mapping $\mathbb{S}$ and $\mathbb{F}$. 

\subsubsection{Training Framework}
In this section we explain in detail our method to work with either inclusive or exclusive domains, and also fine-grained or coarsed transformations. 

First, as each domain has two style distributions, we use the domain information in form of multi-task learning to inject the desired style representation into the generator. The resultant style is a weighted concatenation of all the attributes. Second, we replace the AdaIN and convolution layers with modulated convolutions~\cite{karras2020analyzing}, and we discuss this architectural change in \sref{sec:results}. Third, we propose a novel training scheme critical for the success of the training stability. 

During the forward pass, we first sample a noise vector ($\mathcal{N}(0,I)$) and shuffle the real domain labels ($\hat{y}$) in order to generate a mapping latent vector ($\hat{s}$) that is fed to the generator. The random style is defined in \eref{equation:mapping}.
\begin{equation}
    \hat{s} = \mathbb{F}( \thinspace \mathcal{N}(0,I) \thinspace )_{\hat{y}} 
    \label{equation:mapping}
\end{equation}
For the Discriminator, Mapping Network, and Style encoder, we use multi-task learning on the active domains, and ignore the optimization for the zero-domain vectors.

Fake images are produced as $\hat{x} = \mathbb{G}(x, \hat{s})$. In contrast to StarGANv2, we only require one reconstruction step. We define the style reconstruction loss in \eref{equation:style_rec}.

\begin{equation}
    \mathcal{L}_{sty} = \min_{\mathbb{G},\mathbb{S}} \left[ {\|\hat{s} - \mathbb{S}(\mathbb{G}(x, \hat{s}))_{\hat{y}}\|}_{1} \right]
    \label{equation:style_rec}
\end{equation}

To further encourage the diversity across the transformations, we follow the same pixel-wise style diversification as in StarGAN2. See \eref{equation:style} for the style diversification loss.

\begin{equation}
    \mathcal{L}_{sd} = \max_{\mathbb{G}} \left[ {\|\mathbb{G}(x, \hat{s})) - \mathbb{G}(x, \hat{s}'))\|}_{1} \right]
    \label{equation:style}
\end{equation}

They key ingredient to stabilize our system relies on the reconstruction loss. 
As we are only learning $\mathbb{S}$ parameters using \eref{equation:style_rec}, and we need to align the style encoder for both real and fake images, for the reconstruction loss we simply detach the weights from the graph. With this strategy we force the two distributions $\mathbb{S}$ and $\mathbb{F}$ to be aligned. We found this trick to be crucial in the overall training framework. Therefore, the real style has the form of $\tilde{s}=\text{detach}(\mathbb{S}(x)_{y^r})$, and we define this loss in \eref{equation:reconstruction}.

\begin{equation}
    \mathcal{L}_{rec} = \min_{\mathbb{G}}\left[{\|x - \mathbb{G} (\mathbb{G}(x, \check{s}), \tilde{s})\|}_{1}\right]
    \label{equation:reconstruction}
\end{equation}

As usual, we use the adversarial loss (${L}_{adv}$) to produce photo-realistic images. We follow the same adversarial loss and regularizer as in StarGANv2.

Our full loss function is defined in \eref{equation:total_loss}.

\begin{equation}
  \begin{split}
    \mathcal{L} = & \thinspace \mathcal{L}_{adv} + \lambda_{rec}\mathcal{L}_{rec} + \lambda_{sty}\mathcal{L}_{sty}
    + \lambda_{sd}\mathcal{L}_{sd}
  \end{split}
  \label{equation:total_loss}
\end{equation}  
where each $\lambda$ represents the relative importance within the system.

\subsubsection{Experimental Setup}
We build our system for $256\times256$ image size. However, with enough computational time, we found that it can be easily extended to higher resolutions by keeping the same number of parameters. 

For all our experiments we set $\lambda_{rec}=1.0$, $\lambda_{sty}=1.0$, and $\lambda_{ds}=20.0$. We train our system during 200,000 iterations using a single GPU Titan Xp with a batch size of 6, and Adam Optimizer~\cite{kingma2014adam}. 

\aref{\ref{appendix:architectures}} for more details about the networks.

\begin{table*}[t]
\begin{center}
\resizebox{\linewidth}{!}{
\begin{tabular}{|l||c|c|c||c|c||c|||c|||c|}
\hline
& \multicolumn{8}{c|}{CelebA-HQ~\cite{karras2017progressive} | Latent Synthesis} \\
\cline{2-9}
& \multicolumn{3}{c||}{Pose$\downarrow$} & \multicolumn{2}{c||}{Attributes$\uparrow$} & \multicolumn{1}{c|||}{Reconstruction$\uparrow$} & \multicolumn{2}{c|}{Perceptual} \\
\cline{2-9}
& Roll & Pitch & Yaw & AP & F1 & mIoU & FID$\downarrow$ & Diversity$\uparrow$ \\
\hline
Baseline~\cite{choi2019starganv2} - RGB & 2.952 $\pm$ 0.856 & 16.900 $\pm$ 6.264 & 29.331 $\pm$ 8.134 & 0.795 $\pm$ 0.092 & 0.797 $\pm$ 0.079 & 0.964 $\pm$ 0.012 & 81.945 $\pm$ 24.276 & 0.018 $\pm$ 0.008 \\
(A): Baseline~\cite{choi2019starganv2} - SEM & 2.359 $\pm$ 0.678 & 13.520 $\pm$ 4.476 & 15.424 $\pm$ 6.432 & 0.889 $\pm$ 0.062 & 0.884 $\pm$ 0.051 & 0.994 $\pm$ 0.001 & 61.015 $\pm$ 22.235 & 0.382 $\pm$ 0.014 \\
(B): + Detaching Style                       & 2.732 $\pm$ 0.681 & 18.172 $\pm$ 4.500 & 17.626 $\pm$ 7.250 & 0.940 $\pm$ 0.039 & 0.928 $\pm$ 0.038 & 0.987 $\pm$ 0.003 & 46.797 $\pm$ 14.204 & 0.395 $\pm$ 0.013 \\
(C): + Modulated Conv                        & 2.683 $\pm$ 0.792 & 18.628 $\pm$ 6.243 & 10.553 $\pm$ 2.560 & 0.965 $\pm$ 0.028 & 0.953 $\pm$ 0.027 & 0.986 $\pm$ 0.002 & 48.123 $\pm$ 14.759 & 0.390 $\pm$ 0.013 \\
(D): + Weighting Classes                     & 2.589 $\pm$ 0.684 & 15.082 $\pm$ 4.097 & 11.286 $\pm$ 1.983 & 0.960 $\pm$ 0.031 & 0.946 $\pm$ 0.032 & 0.989 $\pm$ 0.002 & 43.151 $\pm$ 15.527 & 0.399 $\pm$ 0.020 \\
\hline
\end{tabular}
}
\caption{Quantitative contribution of each component of our system for Latent Synthesis manipulation. Each row depicts cumulative addition of each stage with respect to the baseline method.
$\downarrow$ and $\uparrow$ mean that lower is better and higher is better, respectively. Note that Diversity computes the LPIPS perceptual dissimilarity across different styles for a single input, therefore higher is better.}
\label{table:ablation_sem}
\end{center}
\end{table*}

\begin{table*}[t]
\begin{center}
\resizebox{\linewidth}{!}{
\begin{tabular}{|l||c|c|c||c|c||c|||c|||c|}
\hline
& \multicolumn{8}{c|}{CelebA-HQ~\cite{karras2017progressive} | Reference Synthesis} \\
\cline{2-9}
& \multicolumn{3}{c||}{Pose$\downarrow$} & \multicolumn{2}{c||}{Attributes$\uparrow$} & \multicolumn{1}{c|||}{Reconstruction$\uparrow$} & \multicolumn{2}{c|}{Perceptual} \\
\cline{2-9}
& Roll & Pitch & Yaw & AP & F1 & mIoU & FID$\downarrow$ & Diversity$\uparrow$ \\
\hline
Baseline~\cite{choi2019starganv2} - RGB & 2.472 $\pm$ 0.726 & 14.691 $\pm$ 3.987 & 31.071 $\pm$ 15.769 & 0.811 $\pm$ 0.086 & 0.806 $\pm$ 0.077 & 0.971 $\pm$ 0.012 & 72.910 $\pm$ 18.961 & 0.214 $\pm$ 0.051 \\
(A) Baseline~\cite{choi2019starganv2} - SEM & 2.011 $\pm$ 0.698 & 10.811 $\pm$ 4.247 & 13.765 $\pm$ 7.567 & 0.899 $\pm$ 0.063 & 0.887 $\pm$ 0.060 & 0.994 $\pm$ 0.001 & 65.863 $\pm$ 26.084 & 0.136 $\pm$ 0.058 \\
(B) + Detaching Style                       & 2.277 $\pm$ 0.595 & 16.362 $\pm$ 4.304 & 14.952 $\pm$ 5.364 & 0.919 $\pm$ 0.047 & 0.909 $\pm$ 0.043 & 0.987 $\pm$ 0.003 & 53.298 $\pm$ 23.361 & 0.132 $\pm$ 0.057 \\
(C) + Modulated Conv                        & 2.182 $\pm$ 0.652 & 17.142 $\pm$ 6.113 & 9.117 $\pm$ 1.280 & 0.943 $\pm$ 0.031 & 0.930 $\pm$ 0.029 & 0.986 $\pm$ 0.002 & 52.327 $\pm$ 23.352 & 0.111 $\pm$ 0.064 \\
(D) + Weighting Classes                     & 1.948 $\pm$ 0.450 & 13.225 $\pm$ 3.428 & 9.439 $\pm$ 1.826 & 0.942 $\pm$ 0.030 & 0.928 $\pm$ 0.031 & 0.989 $\pm$ 0.002 & 50.257 $\pm$ 24.735 & 0.129 $\pm$ 0.083 \\
\hline
\end{tabular}
}
\caption{Quantitative contribution of each component of our system for Exemplar Image manipulation. Each row depicts cumulative addition of each stage with respect to the baseline method. 
$\downarrow$ and $\uparrow$ mean that lower is better and higher is better, respectively. Note that Diversity computes the LPIPS perceptual dissimilarity across different styles for a single input, therefore higher is better.}
\label{table:ablation_sem2}
\end{center}
\end{table*}

\subsection{Improved Semantic Image Synthesis}
In order to map from semantic regions to an RGB image, we use our semantically guided image synthesis method to perform the corresponding generation either by using an exemplar image or exploring the latent space. 

Current methods~\cite{viazovetskyi2020stylegan2,shen2020interpreting,abdal2019image2stylegan} that use StyleGAN~\cite{karras2019style} for image manipulation, latent disentanglement or image projection have to go through several steps: (i) train StyleGAN until convergence, (ii) study the latent space to produce meaningful yet visible disentangled representation which usually involves more training stages, and (iii) optimize the latent space for a reference image. 
We propose a method that only requires training until convergence, and during inference both the latent space manipulation and image reconstruction can be efficiently and effectively achieved. 

\subsubsection{Model}
Recently, SEAN~\cite{zhu2020sean} and MaskGAN~\cite{lee2020maskgan} haven been proposed as strong alternatives for the generation of images using layout references by disentangling each style to each semantic region. However, the generation suffer from being tied to an exemplar image. In a similar direction, StyleGAN2~\cite{karras2020analyzing} is the current state-of-the-art for image generation. Inspired by Hong~\etal~\cite{hong2020low} and SEAN, we replace StyleGAN2 modulated convolutions (ModConv) with improved semantically region-wise adaptive convolutions (SACs). Let $w$ be the kernel weight, $h$ the input features of the convolution, $s$ the condition information, and $\sigma_E$ the standard deviation also known as the demodulating factor, we define SACs in~\eref{equation:sacs}.

\begin{gather}
  \text{ModConv}_w(\mathbf{h}, s) = \frac{w * \left(s \mathbf{h}\right)}{\sigma_E(w, s)} \Leftrightarrow \thinspace s \in \mathbb{R}^{1 \times C \times 1 \times 1} \nonumber 
  \\
  \text{SAC}_w(\mathbf{h}, \mathbf{s}) = \frac{w * \left(\mathbf{s} \odot \mathbf{h}\right)}{\sigma_E(w,  \mathbf{s})} \Leftrightarrow \thinspace \textbf{s} \in \mathbb{R}^{1 \times C \times H \times W} \label{equation:sacs},
\end{gather}
where,
\begin{equation*}
    \mathbf{s} = \alpha_w SM + (1-\alpha_w)M,
\end{equation*}

where $SM$ is the per-region style matrix, which can be either extracted from an image or sampled from a gaussian distribution, $M$ is the required semantic mask, and $\alpha_w$ is a learned parameter that weights for the relative importance of each element at each layer of the network. This equation can also be seen as the SEAN~\cite{zhu2020sean} gamma factor. Please see \cite{hong2020low} for further details on the mathematical development of~\eref{equation:sacs}.

\subsubsection{Training Framework}

To couple this proposed scheme with the StyleGAN2 training framework, we propose an alternate scheme training. First, we update the generator ($\mathbbm{G}$) and discriminator ($\mathbbm{D}$) for random generation as in StyleGAN2. Second, in addition to the generator and discriminator we also update an style encoder network ($\mathbbm{S}$) for exemplar-guided synthesis. For simplicity, we show the loss function for the generator during the reference synthesis in \eref{equation:stylegan2_reference}. To this end, let $x$ and $m$ be the real image and its corresponding semantic map, respectively. 

\begin{gather}
    \mathcal{L}_{feat} = \min_{\mathbbm{G},\mathbbm{S}} \sum_{i=1}^{T-1} \frac{1}{N_i} \left[ {\| \mathbbm{D}_k^{(i)}(x) - \mathbbm{D}_k^{(i)}(\mathbbm{G}(m, \mathbbm{S}(x, m) \|}_{1} \right] \nonumber \\
    \mathcal{L}_{\text{reference}} = \mathcal{L}_{adv} + \lambda_{feat}\mathcal{L}_{feat} 
    \label{equation:stylegan2_reference}
\end{gather}
where T is the total number of layers in the discriminator, $N$ is the number of elements in each layer, and $\lambda_{feat}$ and $\lambda_{pt}$ represents the importance of the feature matching loss~\cite{wang2018pix2pixHD}, and it is set to $10$.

Note that the feature matching loss is only required for the reference update.

\subsubsection{Experimental Setup}
The generator uses the semantic maps as starting point for the image synthesis. Instead of starting from a constant representation as in StyleGAN, and as the semantic segmentation information represents the high-level information of the data, we do not have to start the generator network from as low as $4 \time 4$ feature maps. Instead, we empirically found that starting from $8 \times 8$ yields a better performance. \aref{\ref{appendix:additional_syn_mask}} for this experiment.

Given current computational limitations to fully train StyleGAN2, we train our system during 300,000 iterations (roughly 3 weeks) using a single GPU Titan Xp with a batch size of 4 and image size of 256. 

\aref{\ref{appendix:architectures}} for more details about the networks.
\begin{figure*}[t]
\begin{center}
\resizebox{\linewidth}{!}{
\begin{tabular}{c||cccc}
    \rotatebox{90}{Random Sampling} & \includegraphics[width=0.25\textwidth]{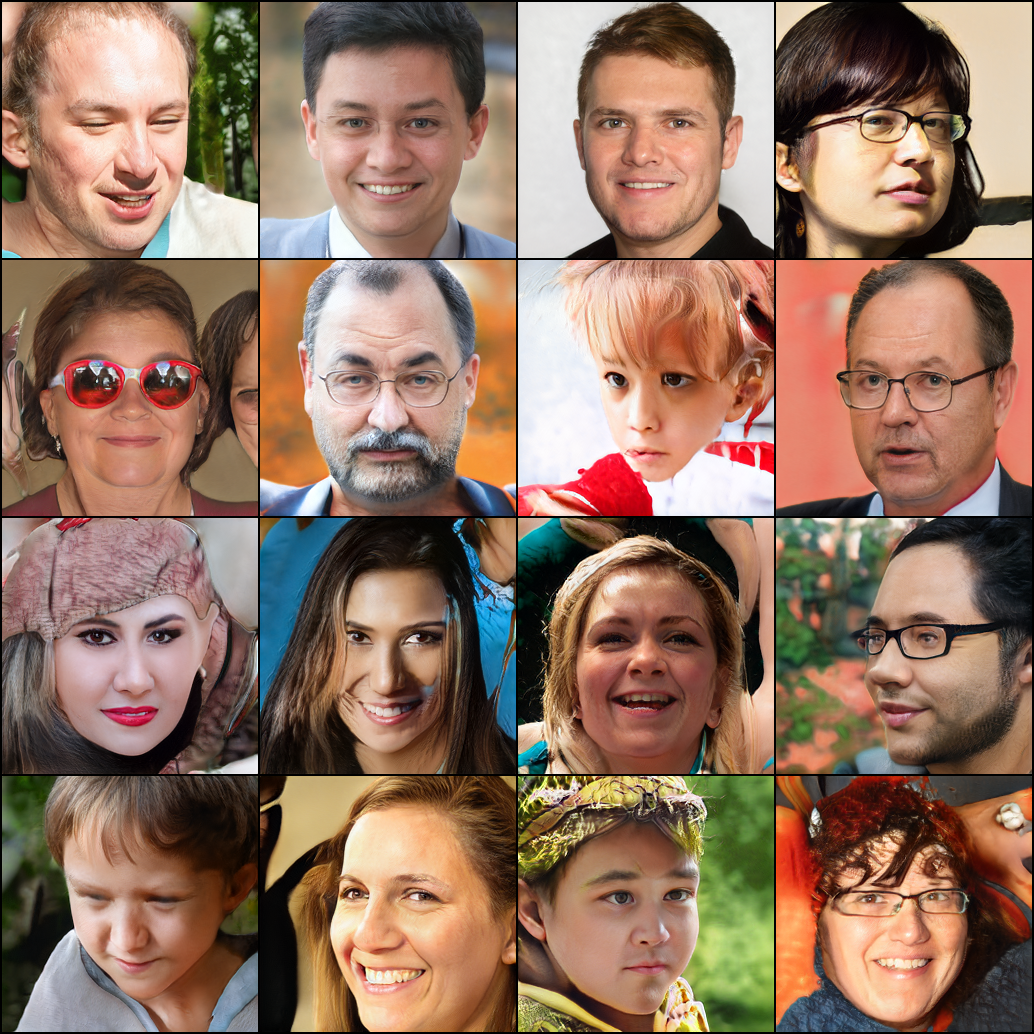} &
    \includegraphics[width=0.25\textwidth]{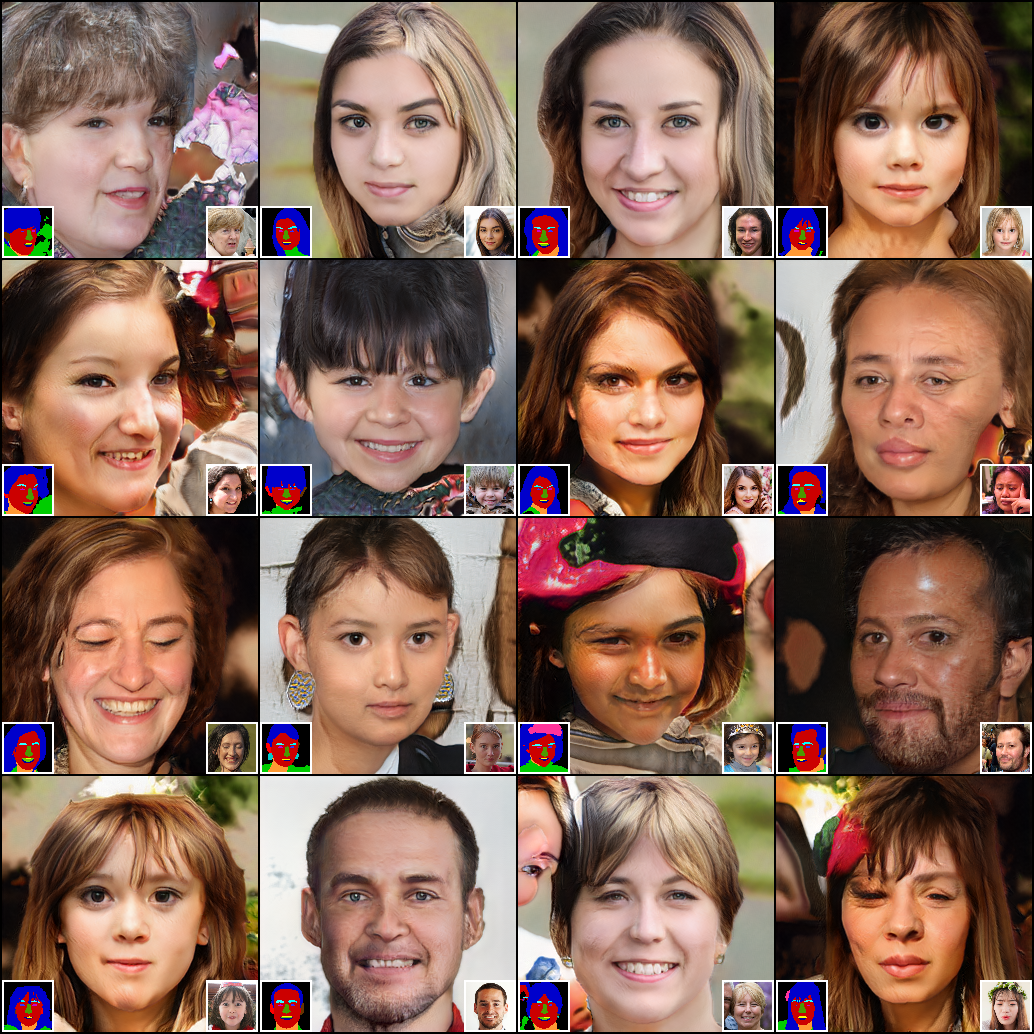} & 
    \includegraphics[width=0.25\textwidth]{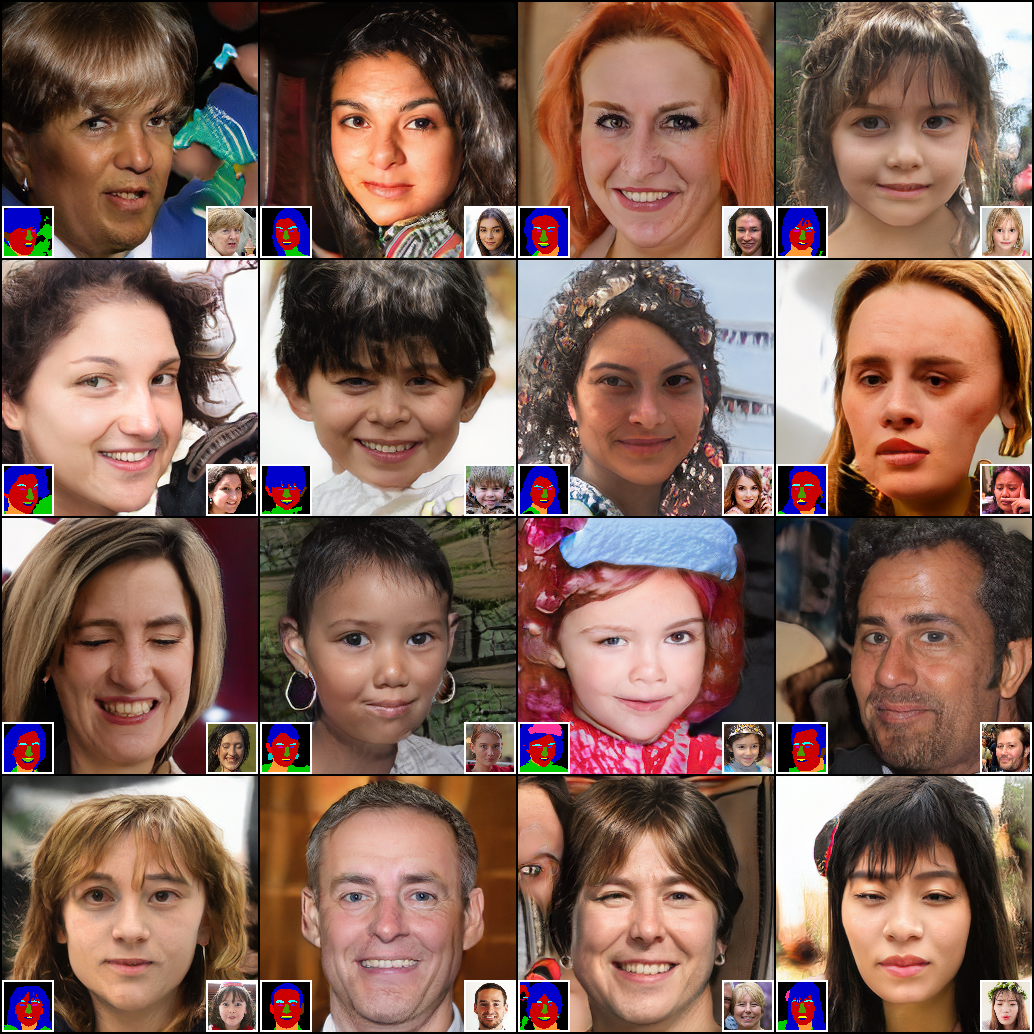} & 
    \includegraphics[width=0.25\textwidth]{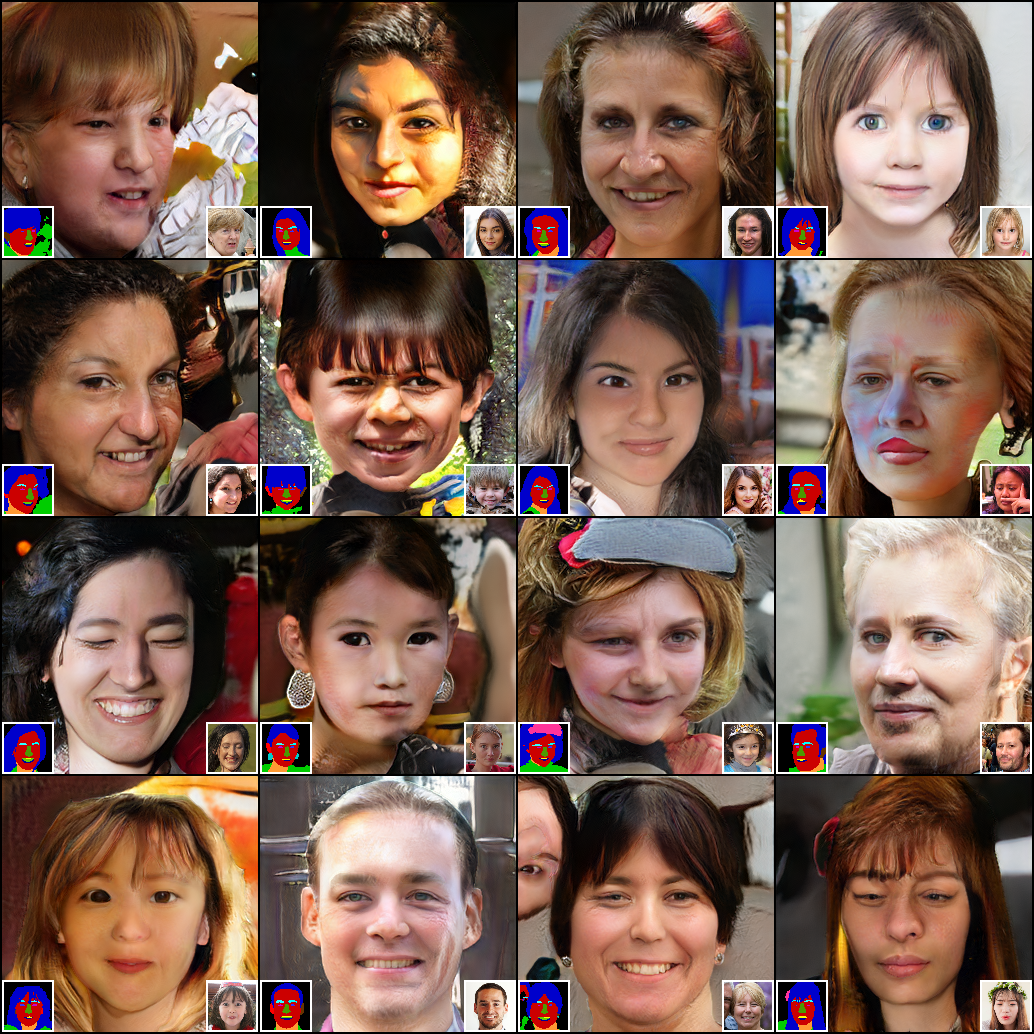} \\
    \midrule
    \rotatebox{90}{Projection Sampling} & \includegraphics[width=0.25\textwidth]{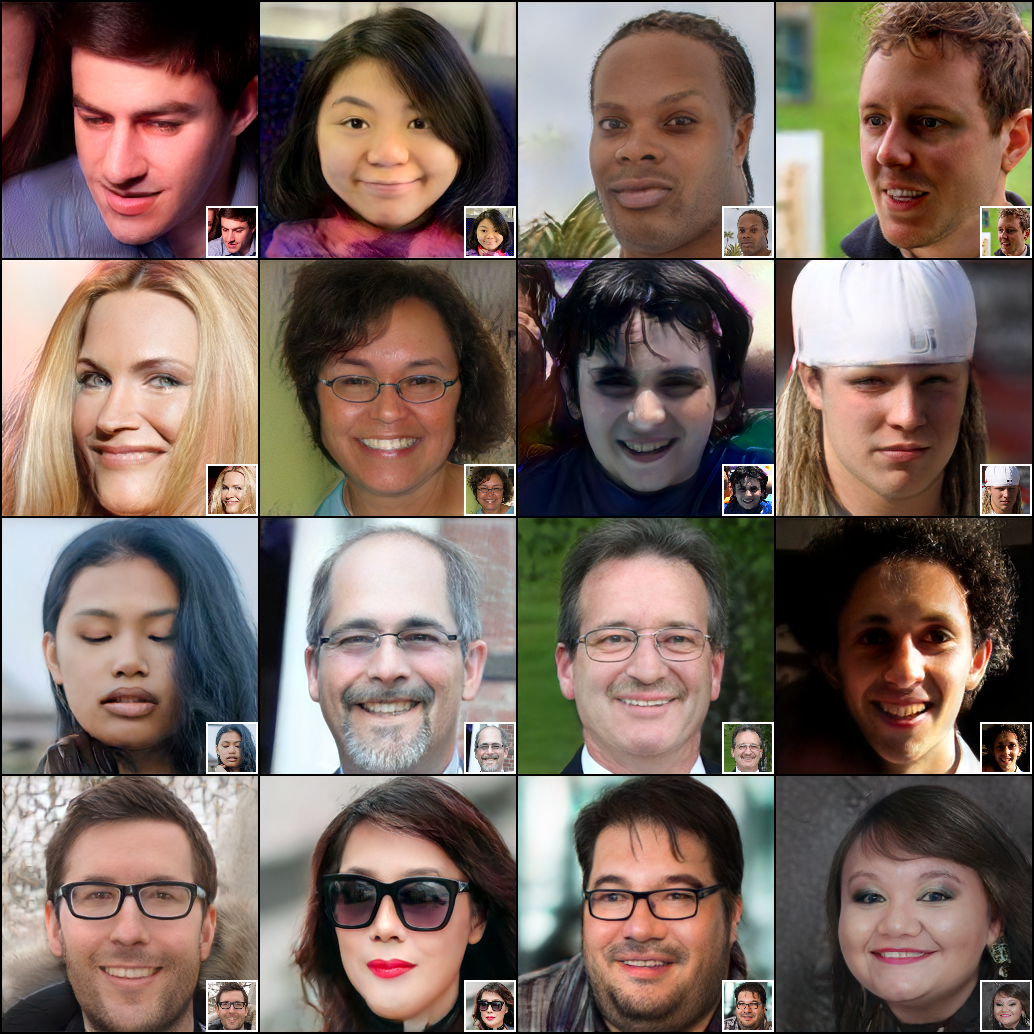} &
    \includegraphics[width=0.25\textwidth]{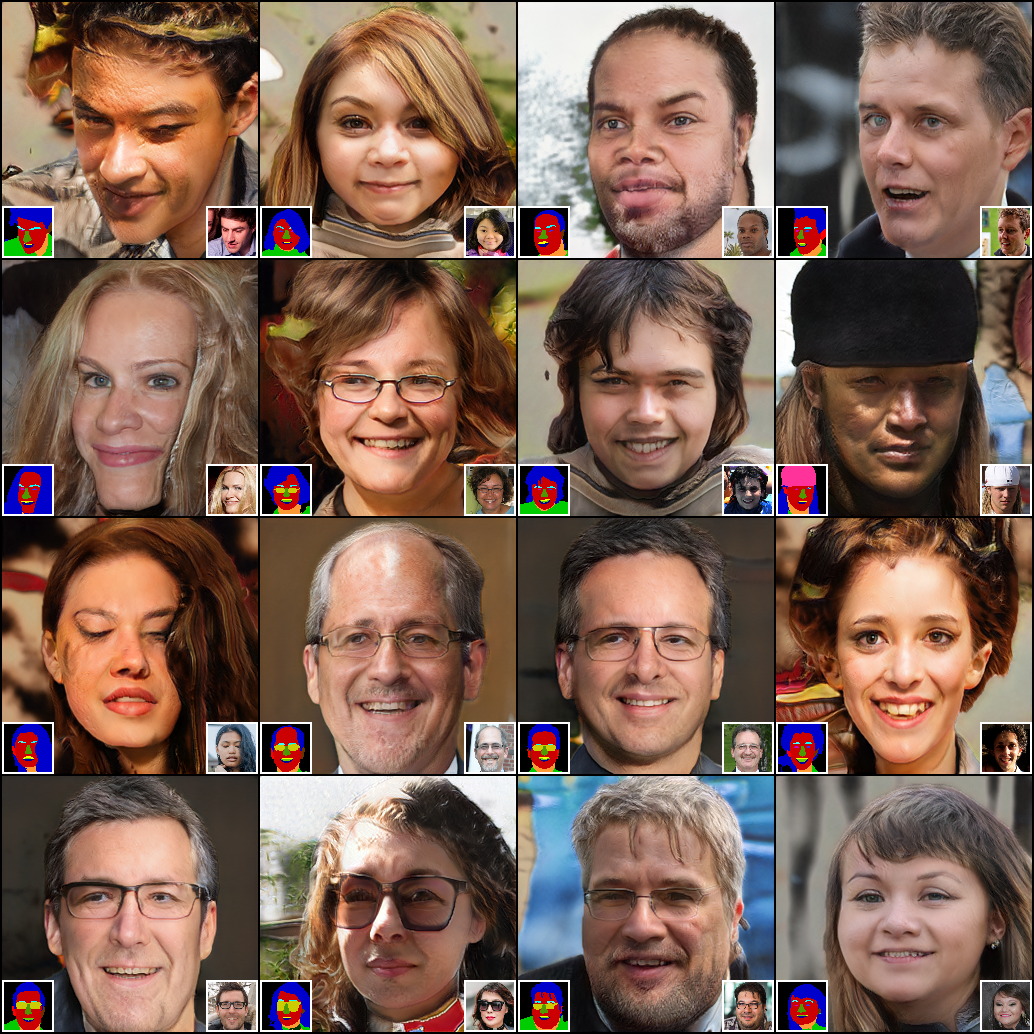} & 
    \includegraphics[width=0.25\textwidth]{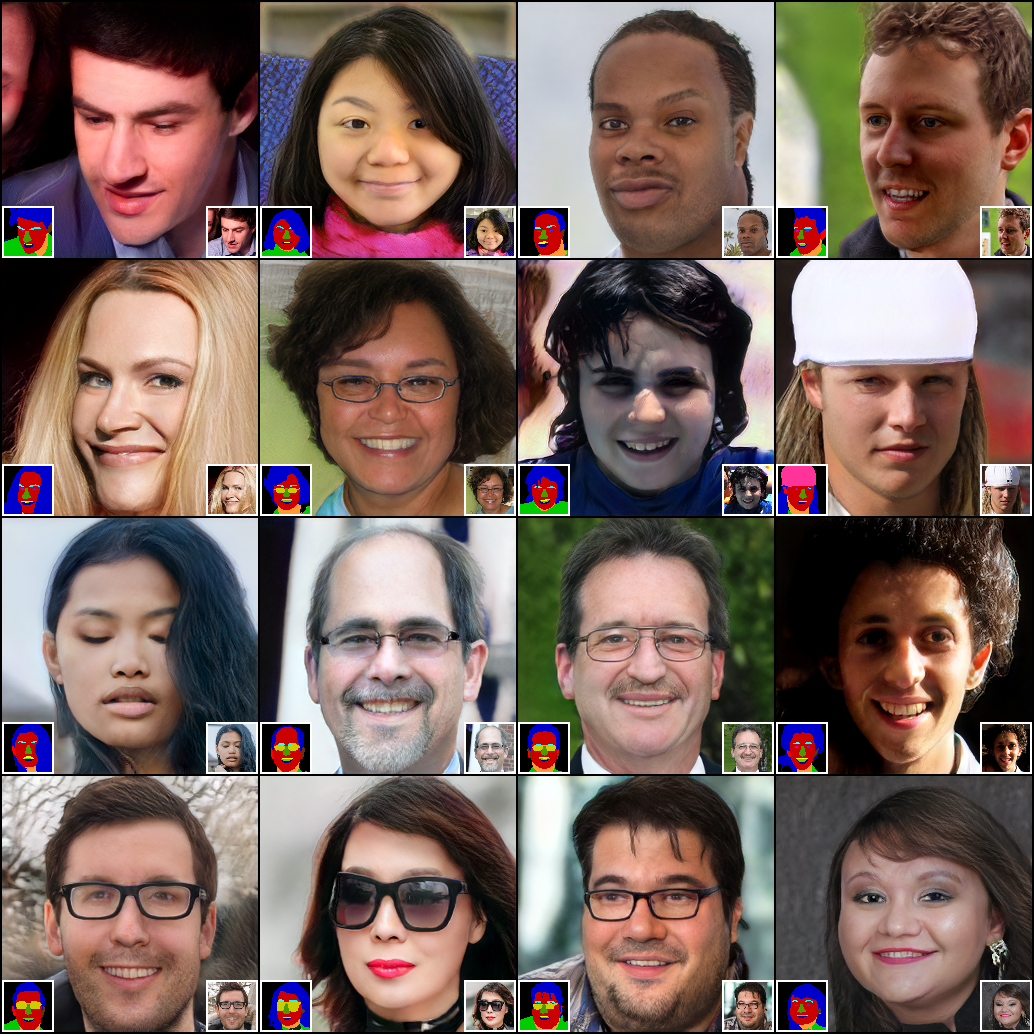} & 
    \includegraphics[width=0.25\textwidth]{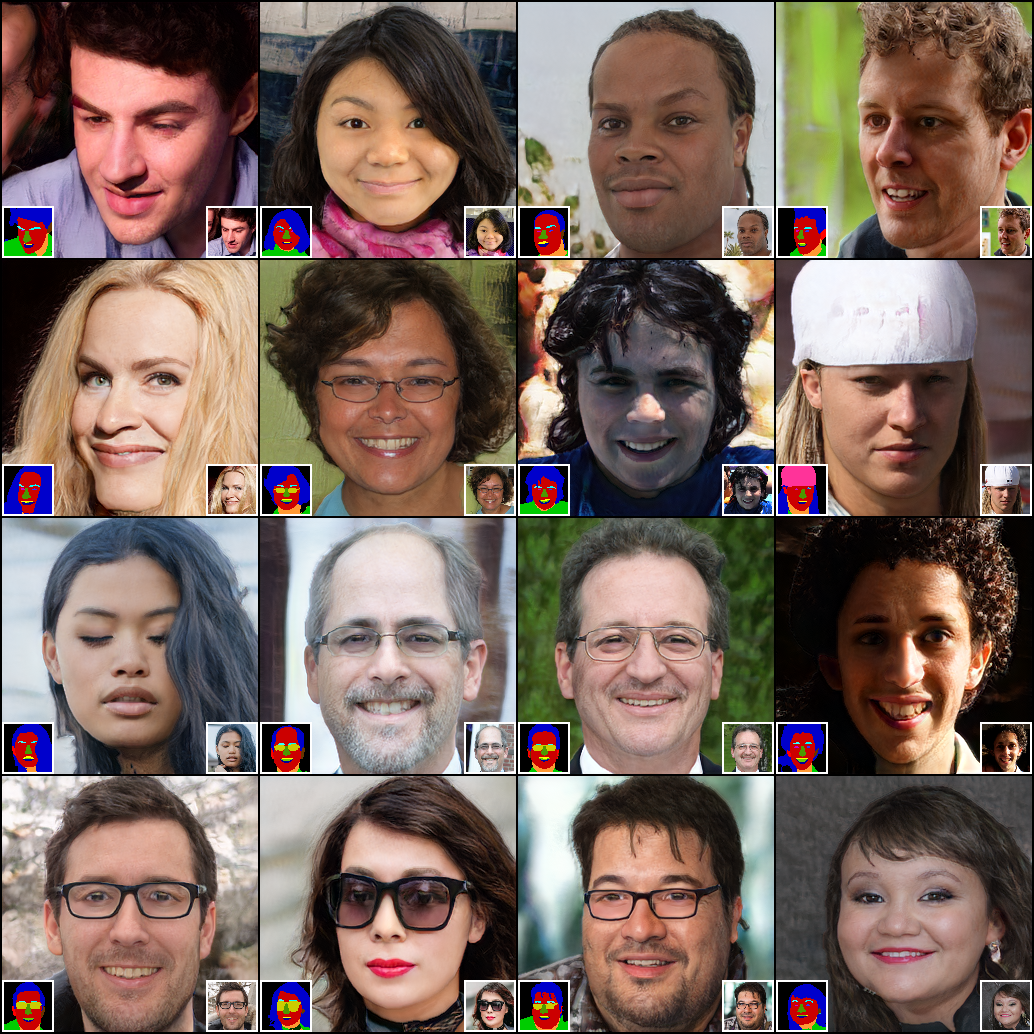} \\
\end{tabular}
}
\begin{tabular}{@{}*{4}{P{.25\textwidth}@{}}}
    \hspace{1.2cm} Baseline & \hspace{0.7cm} Ablation A {\small(+ mask)} & \hspace{0.4cm} Ablation B {\small(+ style matrix)} & Ablation C {\small(+ encoder)}
\end{tabular}
\end{center}
\caption{Ablation experiments for image synthesis. We progressively add ablation experiments to the baseline~\cite{karras2020analyzing} in order to present our system in the last column. Upper and bottom parts of the figure shows qualitative results for random sampling and projection reconstruction, respectively. We show the input semantics and reference images in the bottom corners of each image. Zoom in for better details.}
\label{figure:ablation_syn}
\end{figure*}

\subsection{Datasets}
\paragraph{Semantic Manipulation}
We validate our semantic manipulation method in CelebA-HQ~\cite{karras2017progressive} that consists of multiple facial attribute labels. Since we are tackling semantic manipulation, we selected 6 visible attributes that were not related to facial texture: eyeglasses, hat, amount of hair, bangs, earrings and identity\footnote{As binary gender might be a sensitive topic, we refer to identity as the Male/Female label in the CelebA-HQ dataset.}. For the semantic segmentation labels, we use the ones provided by CelebA-Mask~\cite{lee2020maskgan}.

\paragraph{Semantically Image Synthesis}
Since the semantically image synthesis does not require having access to labels, we validate this part of the system using FFHQ~\cite{karras2019style}. 

\subsection{Evaluation Framework}
For our entire system we study independent performances under two circumstances: generation by latent space and generation by exemplar images. Since our proposed solution lies in a unexplored area, the evaluation of semantic segmentation manipulations is challenging and it does not have a standard benchmark in the literature. Therefore, we first evaluate the Semantically Image Synthesis approach, and use this model to generate RGB from the Semantic Manipulation approach and thus evaluate it using standard metrics.  

\paragraph{Semantically Image Synthesis}
As it is common for image synthesis, we report the Fr\'echet Inception Distance~\cite{salimans2016improved} (FID), and the Perceptual Similarity Score (LPIPS)~\cite{zhang2018lpisp} as a measure of dissimilarity across transformations (Diversity). We strictly follow the same evaluation framework proposed in StyleGAN2 for FID. Since we have to use real semantic annotations for the evaluation protocol, we use 10.000 samples (the entire test set of FFHQ) to compute the FID score. For Diversity we generate 10 different samples from a single semantic input, and compute the LPIPS score across each pair, for all possible pairs. In our case, LPIPS score is associated to diversity rather than similarity.

\paragraph{Semantic Manipulation}
There are two main aspects we consider for the proper evaluation of our system: the transformation mapping must resemble the pose of the person and the image must contain the target attributes. To this end, we use an off-the-shelf pose estimator~\cite{ruiz2018fine} (HopeNet) and use the training set of CelebA-HQ to train an attribute classifier using MobilenetV2~\cite{sandler2018mobilenetv2}. For the entire CelebA-HQ test set, we manipulate each image using an specific attribute and keeping the others unaltered (for instance only male $\leftrightarrow$ female), and we perform 10 transformations per image. We then extract the average Yaw, Pitch and Roll using HopeNet, and the average attribute scores using MobilenetV2, and compute the Root-Mean Squared Error (RMSE), Average Precision (AP) and F1 score between the test set and generated images per attribute, for the entire set attributes. Furthermore, under the same protocol, we also report the FID between the training set and generated images for each attribute. In addition to FID, we compute the perceptual Diversity metric across each image in the test set and 10 different transformations. For both perceptual metrics, we follow the same validation protocol as in StarGANv2. Furthermore, we also report the mean Intersection over Union (mIOU) over the input image and the reconstructed cycle image. (\eref{equation:reconstruction}).

\begin{table*}[t]
\begin{center}
\resizebox{\linewidth}{!}{
\begin{tabular}{|l||c|c|c||c|c|c|c|c||c|}
\hline
\multicolumn{10}{|c|}{FFHQ~\cite{karras2019style}} \\
\cline{1-10}
\multicolumn{1}{|c||}{\multirow{2}{*}{Experiment}} &
\multicolumn{3}{c||}{Latent Synthesis} & 
\multicolumn{5}{c||}{Reference Synthesis} &
\multicolumn{1}{c|}{\multirow{2}{*}{Training Time [days]}}\\
\cline{2-9}
 & FID$\downarrow$  & Diversity$\uparrow$ & Runtime [s/img] & LPIPS$\downarrow$ & PSNR$\uparrow$ & SSIM$\uparrow$ & RMSE$\downarrow$ & Runtime [s/img] & \\
\hline
StyleGAN2~\cite{karras2020analyzing}  & 15.15\footnotemark & - & 0.03 & - & - & - & - & 120 & 2.5 \\
(A) + masks                           & 24.12 & 0.08 $\pm$ 0.03 & 0.06  & - & - & - & - & 180 & 3.8 \\
(B) + style matrix                    & 13.08 & 0.42 $\pm$ 0.04 & 0.13 & - & - & - & - & 210 & 9.4 \\
(C) + encoder                         & 16.99 & 0.43 $\pm$ 0.03 & 0.13 & 0.22 $\pm$ 0.06 & 18.19 $\pm$ 2.84 & 0.19 $\pm$ 0.04 & 0.13 $\pm$ 0.03 & 0.13 $\pm$ 0.04 & 17.5 \\
\hline
\end{tabular}
}
\caption{Quantitative evaluation of our system in image synthesis under different configurations, and in comparison with the baseline.}
\label{table:ablation_syn}
\end{center}
\end{table*}
\begin{figure*}[t]
\begin{center}
\centering
\resizebox{\linewidth}{!}{
\begin{tabular}{c||c}
    \rotatebox{90}{Video Guided Sampling} & \includegraphics[width=.8\textwidth]{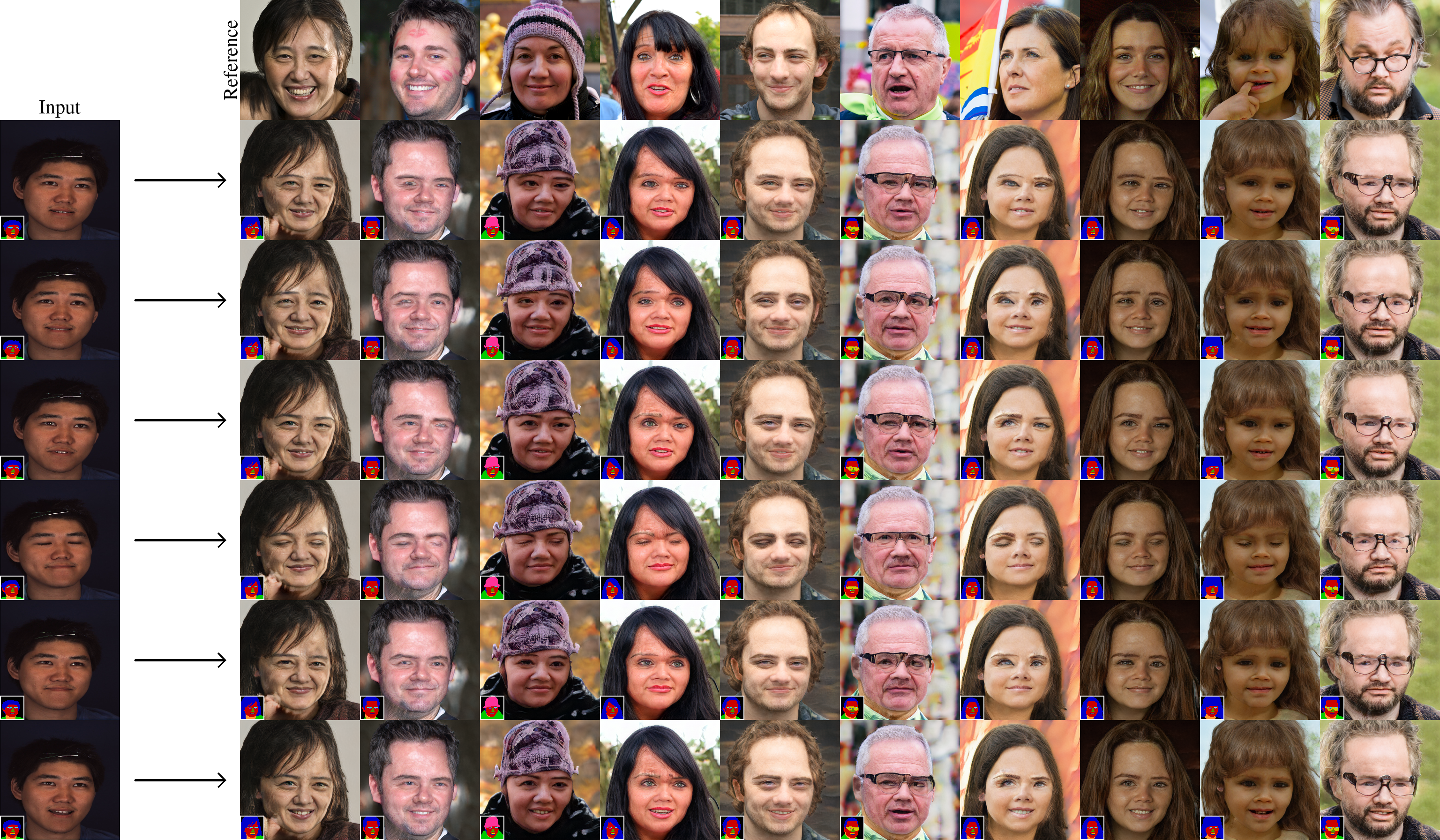} \\
    \midrule
    \rotatebox{90}{Video Random Sampling} & \includegraphics[width=.8\textwidth]{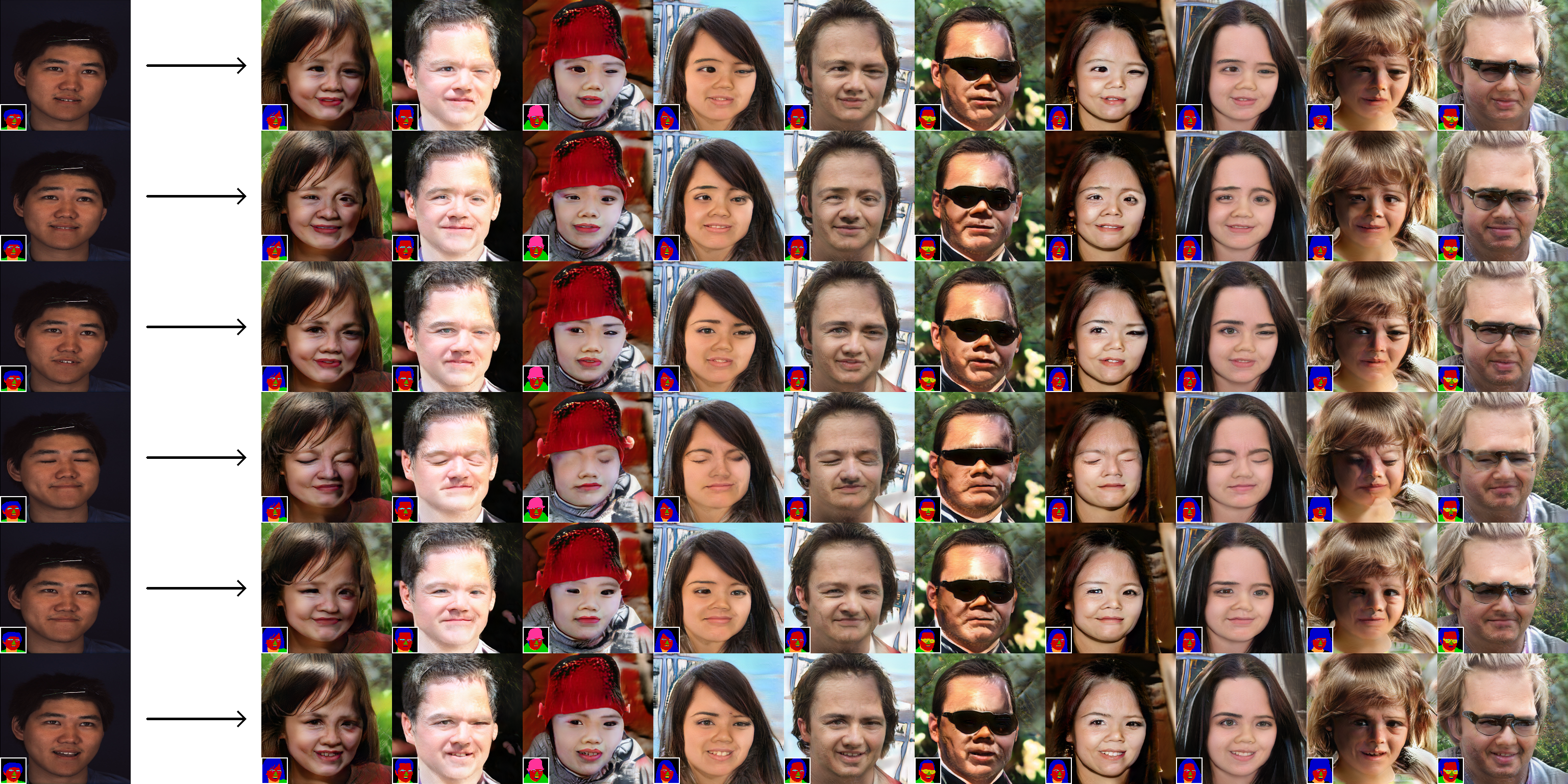} \\
\end{tabular}
}
\end{center}
   \caption{Qualitative visualizations for \SHORTTITLE{}. These images highlights the full transformation framework. First, we manipulate the semantic input with respect to the style reference. Second, for both reference and random sampling we keep the same style matrix and the video generation is driven by the semantic feed. Note that upper and lower figures share the same semantics (left bottom corner of each image). Remarkably, we do not train our system using video information.}
\label{figure:qualitative_image2}
\end{figure*}

\section{Discussion}
\label{sec:results}
In this section we discuss in detail the aspects that strengthen our method. We depict in~\fref{figure:ablation_sem} and \ref{figure:ablation_syn}, and quantitatively evidence in~\tref{table:ablation_sem}, \ref{table:ablation_sem2} and \ref{table:ablation_syn} each part of our system. Each change is made in a cumulative way starting from the baseline. Note that the numbers reported in \tref{table:ablation_sem} and \ref{table:ablation_sem2} are the average scores for the 8 different attribute manipulations. To this end, we report the standard deviation for each score. \aref{} for a detailed table for each attribute manipulation.  

\subsection{Semantic Manipulation}
By using the state-of-the-art method~\cite{choi2019starganv2} in multi-domain image manipulation as baseline, we first extend it to deal with multiple mutually-inclusive domains. For this, we apply the concatenation of all the target styles as an input in addition to the RGB image. As~\fref{figure:ablation_sem} shows and~\tref{table:ablation_sem} indicate, the baseline using RGB completely fails at this task, \ie, it does not generalize well to different domains. Interestingly, we empirically found that StarGANv2 struggles when trained in a one domain framework different than Male/Female. We hypothesize that it is due to the fact that the style encoder extracts general characteristics of the entire image and because of the lack of supervision it cannot focus on fine-grained styles. In order to circumvent this problem, we instead use the semantic information as input, and better perform manipulations in this space (A). This change leads to cleaner and sharper transformations that better approximates the desired domain. 

Moreover, we found that (B) by disabling the gradients of the style encoder during the reconstruction pass, it is sufficient for the overall training framework and reduces the training time by half. Our rationale is as follows: as we are injecting a random style through the mapping function, and the style encoder learns to reconstruct it, then we can assume that after enough iterations the style encoder extracts the corresponding style from the real image. 

\footnotetext{We report the StyleGAN2 FID for a model trained with batch size 4 and 300,000 iterations. It is possible that this result does not match with the one reported in the original paper.}

Simplifying our system for the semantic space brings the problem of losing the texture information. We found that using Adaptive Instance Normalization layers (AdaIN~\cite{huang2017arbitraryadain}) it deforms the input's image pose, in particular Yaw, during the transformation, and yet it still minimizes the proposed formulation in~\eref{equation:total_loss}. Therefore, we noticed that (C) conditionally modulating the weights of the convolutions~\cite{karras2020analyzing} alleviates this issue and the output resembles the pose of the input.

Furthermore, As we assumed equal dimensionality contribution per domain, transformations produced in stage (C) are not diverse enough across domains. By closely inspecting the resultant images we found that the identity domain is the least diverse in spite of being the most abstract and the one that models the biggest part of the image (for instance facial structure, clothes, hair style, etc). To this end, we weighted the identity domain to have more representation in the final latent vector with respect to the others (D), so this domain has a bigger impact than other domains (\eg, eyeglasses, bangs, etc) in the reconstruction style loss, and it can consequently produce more diverse transformations. This change of dimensionality for the identity domain is inspired in an observation from the semantic space. Most of the selected domains have one specific corresponding channel in the semantic space that facilitates the style encoding, yet it is not the case for the identity.

Additionally, to further assess a quantitative disentangled level of the manipulations, we studied how each independent transformation affects the unaltered attributes. We accomplished this feat by computing Precision and Recall curves over each manipulation. \aref{\ref{appendix:additional_sem_pr}} for the generated curves.

\subsection{Semantically Image Synthesis}
StyleGAN2~\cite{karras2020analyzing} is the state-of-the-art method in image synthesis. Using this method for disentangled representations or modifying an existing image is challenging and normally involves different post-processing techniques. In order to modify StyleGAN2 backbone to be able to perform both disentangled representations in the semantic space and modify existing images, we progressively introduce different subtle but critical changes to the architecture. See~\fref{figure:ablation_syn} and~\tref{table:ablation_syn} for qualitative and quantitative ablative comparison, respectively. 

To overcome the semantic injection into the StyleGAN2 pipeline, we replace all the Modulated Convolutions by SAC layers (\eref{equation:sacs}). For the first experiment, (A) we only generate random images using the semantic map in a SPADE~\cite{park2019SPADE} fashion, which implies that there is not enough diversity in the generation. Next, inspired in SEAN~\cite{zhu2020sean}, (B) we use the full style per-region matrix in conjunction with the segmentation mask to generate diverse images controlled by random noise in each region. Finally, in order to take full advantage of the StyleGAN2 training framework, and incorporate a new style encoder, (C) we train our system in an alternate way producing thus random and original images, respectively. As a result of the alignment between the mapping random style and style encoder, we encountered a trade-off in the performance. This trade-off is expected due to the random nature of the latent sampling, as the mapping network can receive very different styles for skin, neck, and ears for example, which is not the case for the reference synthesis.
Our proposed system compares favorably with StyleGAN2 yet does not require a post-optimization process for image projection. 
\section{Other applications}
One important drawback of face-swapping or deep fakes videos is that the target face must match the entire face of the source, which produce an unpleasant feeling by fitting the content of the target into the source. Instead of face-swapping, with \SHORTTITLE, We found that the style matrix extracted from a single image is a strong basis for head swapping (modifying the source content with respect to the target) and also for face reenactment (see
~\fref{figure:qualitative_image2}). \aref{\ref{appendix:reenactment}} for more qualitative results.

\section{Conclusions}
We introduced \SHORTTITLE, a method for multi-attribute image-to-image translation using random sampling or image guiding reference. We extensively show that our method outperforms previous state-of-the-art baselines StarGANv2~\cite{choi2019starganv2} and StyleGAN2~\cite{karras2019style} for both image manipulation and image synthesis.

\paragraph{Acknowledgements} 
This work was partly supported by ETH Zurich Fund (OK), Huawei, Amazon AWS and Nvidia GPU grants. 

{\small
\bibliographystyle{ieee_fullname}
\bibliography{egbib}

\begin{thebibliography}{10}\itemsep=-1pt

\bibitem{abdal2019image2stylegan}
Rameen Abdal, Yipeng Qin, and Peter Wonka.
\newblock Image2stylegan: How to embed images into the stylegan latent space?
\newblock In {\em CVPR}, pages 4432--4441, 2019.

\bibitem{almahairi2018augmentedcyclegan}
Amjad Almahairi, Sai Rajeswar, Alessandro Sordoni, Philip Bachman, and Aaron
  Courville.
\newblock Augmented cyclegan: Learning many-to-many mappings from unpaired
  data.
\newblock In {\em ICML}, pages 195--204, 2018.

\bibitem{bau2020rewriting}
David Bau, Steven Liu, Tongzhou Wang, Jun-Yan Zhu, and Antonio Torralba.
\newblock Rewriting a deep generative model.
\newblock In {\em ECCV}, 2020.

\bibitem{benaim2019domain}
Sagie Benaim, Michael Khaitov, Tomer Galanti, and Lior Wolf.
\newblock Domain intersection and domain difference.
\newblock In {\em ICCV}, pages 3445--3453, 2019.

\bibitem{buhler2020deepsee}
Marcel~Christoph B{\"u}hler, Andr{\'e}s Romero, and Radu Timofte.
\newblock {DeepSEE:} deep disentangled semantic explorative extreme
  super-resolution.
\newblock In {\em ACCV}, 2020.

\bibitem{chang2018pairedcyclegan}
Huiwen Chang, Jingwan Lu, Fisher Yu, and Adam Finkelstein.
\newblock Pairedcyclegan: Asymmetric style transfer for applying and removing
  makeup.
\newblock In {\em CVPR}, pages 40--48, 2018.

\bibitem{cheng2020controllable}
Yen-Chi Cheng, Hsin-Ying Lee, Min Sun, and Ming-Hsuan Yang.
\newblock Controllable image synthesis via segvae.
\newblock In {\em ECCV}, 2020.

\bibitem{choi2017stargan}
Yunjey Choi, Minje Choi, Munyoung Kim, Jung-Woo Ha, Sunghun Kim, and Jaegul
  Choo.
\newblock Stargan: Unified generative adversarial networks for multi-domain
  image-to-image translation.
\newblock In {\em CVPR}, pages 8789--8797, 2018.

\bibitem{choi2019starganv2}
Yunjey Choi, Youngjung Uh, Jaejun Yoo, and Jung-Woo Ha.
\newblock {StarGAN v2:} diverse image synthesis for multiple domains.
\newblock {\em CVPR}, 2020.

\bibitem{goodfellow2014generative}
Ian Goodfellow, Jean Pouget-Abadie, Mehdi Mirza, Bing Xu, David Warde-Farley,
  Sherjil Ozair, Aaron Courville, and Yoshua Bengio.
\newblock Generative adversarial nets.
\newblock In {\em NeurIPS}, pages 2672--2680, 2014.

\bibitem{guo2019mulgan}
Jingtao Guo, Zhenzhen Qian, Zuowei Zhou, and Yi Liu.
\newblock Mulgan: Facial attribute editing by exemplar.
\newblock {\em arXiv preprint arXiv:1912.12396}, 2019.

\bibitem{he2019attgan}
Zhenliang He, Wangmeng Zuo, Meina Kan, Shiguang Shan, and Xilin Chen.
\newblock Attgan: Facial attribute editing by only changing what you want.
\newblock {\em IEEE Transactions on Image Processing}, 28(11):5464--5478, 2019.

\bibitem{hong2020low}
Sarah~Jane Hong, Martin Arjovsky, Ian Thompson, and Darryl Barnhardt.
\newblock Low distortion block-resampling with spatially stochastic networks.
\newblock {\em arXiv preprint arXiv:2006.05394}, 2020.

\bibitem{huang2017arbitraryadain}
Xun Huang and Serge~J Belongie.
\newblock Arbitrary style transfer in real-time with adaptive instance
  normalization.
\newblock In {\em ICCV}, pages 1501--1510, 2017.

\bibitem{huang2018munit}
Xun Huang, Ming-Yu Liu, Serge Belongie, and Jan Kautz.
\newblock Multimodal unsupervised image-to-image translation.
\newblock In {\em ECCV}, pages 172--189, 2018.

\bibitem{karras2017progressive}
Tero Karras, Timo Aila, Samuli Laine, and Jaakko Lehtinen.
\newblock Progressive growing of gans for improved quality, stability, and
  variation.
\newblock In {\em ICLR}, 2018.

\bibitem{karras2019style}
Tero Karras, Samuli Laine, and Timo Aila.
\newblock A style-based generator architecture for generative adversarial
  networks.
\newblock In {\em CVPR}, pages 4401--4410, 2019.

\bibitem{karras2020analyzing}
Tero Karras, Samuli Laine, Miika Aittala, Janne Hellsten, Jaakko Lehtinen, and
  Timo Aila.
\newblock Analyzing and improving the image quality of stylegan.
\newblock In {\em CVPR}, pages 8110--8119, 2020.

\bibitem{kingma2014adam}
Diederik~P Kingma and Jimmy Ba.
\newblock Adam: A method for stochastic optimization.
\newblock {\em arXiv preprint arXiv:1412.6980}, 2014.

\bibitem{lample2017fader}
Guillaume Lample, Neil Zeghidour, Nicolas Usunier, Antoine Bordes, Ludovic
  Denoyer, et~al.
\newblock Fader networks: Manipulating images by sliding attributes.
\newblock In {\em NeurIPS}, pages 5967--5976, 2017.

\bibitem{larsen2015vaegan}
Anders Boesen~Lindbo Larsen, S{\o}ren~Kaae S{\o}nderby, Hugo Larochelle, and
  Ole Winther.
\newblock Autoencoding beyond pixels using a learned similarity metric.
\newblock In {\em ICML}, pages 1558--1566, 2016.

\bibitem{lee2020maskgan}
Cheng-Han Lee, Ziwei Liu, Lingyun Wu, and Ping Luo.
\newblock Maskgan: Towards diverse and interactive facial image manipulation.
\newblock In {\em CVPR}, pages 5549--5558, 2020.

\bibitem{DRIT++}
Hsin-Ying Lee, Hung-Yu Tseng, Qi Mao, Jia-Bin Huang, Yu-Ding Lu, Maneesh~Kumar
  Singh, and Ming-Hsuan Yang.
\newblock Drit++: Diverse image-to-image translation via disentangled
  representations.
\newblock {\em International Journal of Computer Vision}, pages 1--16, 2020.

\bibitem{lin2020distribution}
Minxuan Lin, Fan Tang, Weiming Dong, Xiao Li, Chongyang Ma, and Changsheng Xu.
\newblock Distribution aligned multimodal and multi-domain image stylization.
\newblock {\em arXiv preprint arXiv:2006.01431}, 2020.

\bibitem{liu2019stgan}
Ming Liu, Yukang Ding, Min Xia, Xiao Liu, Errui Ding, Wangmeng Zuo, and Shilei
  Wen.
\newblock Stgan: A unified selective transfer network for arbitrary image
  attribute editing.
\newblock In {\em CVPR}, pages 3673--3682, 2019.

\bibitem{ma2019exemplar}
Liqian Ma, Xu Jia, Stamatios Georgoulis, Tinne Tuytelaars, and Luc Van~Gool.
\newblock Exemplar guided unsupervised image-to-image translation.
\newblock In {\em ICLR}, 2019.

\bibitem{menon2020pulse}
Sachit Menon, Alexandru Damian, Shijia Hu, Nikhil Ravi, and Cynthia Rudin.
\newblock Pulse: Self-supervised photo upsampling via latent space exploration
  of generative models.
\newblock In {\em CVPR}, pages 2437--2445, 2020.

\bibitem{mokady2019mask}
Ron Mokady, Sagie Benaim, Lior Wolf, and Amit Bermano.
\newblock Mask based unsupervised content transfer.
\newblock In {\em ICLR}, 2020.

\bibitem{ntavelis2020sesame}
Evangelos Ntavelis, Andr{\'e}s Romero, Iason Kastanis, Luc Van~Gool, and Radu
  Timofte.
\newblock {SESAME:} semantic editing of scenes by adding, manipulating or
  erasing objects.
\newblock In {\em ECCV}, 2020.

\bibitem{park2019SPADE}
Taesung Park, Ming-Yu Liu, Ting-Chun Wang, and Jun-Yan Zhu.
\newblock Semantic image synthesis with spatially-adaptive normalization.
\newblock In {\em CVPR}, 2019.

\bibitem{park2020swapping}
Taesung Park, Jun-Yan Zhu, Oliver Wang, Jingwan Lu, Eli Shechtman, Alexei~A
  Efros, and Richard Zhang.
\newblock Swapping autoencoder for deep image manipulation.
\newblock {\em arXiv preprint arXiv:2007.00653}, 2020.

\bibitem{pumarola2018ganimation}
Albert Pumarola, Antonio Agudo, Aleix~M Martinez, Alberto Sanfeliu, and
  Francesc Moreno-Noguer.
\newblock Ganimation: Anatomically-aware facial animation from a single image.
\newblock In {\em ECCV}, pages 818--833, 2018.

\bibitem{romero2019smit}
Andr{\'e}s Romero, Pablo Arbel{\'a}ez, Luc Van~Gool, and Radu Timofte.
\newblock {SMIT:} stochastic multi-label image-to-image translation.
\newblock In {\em ICCV Workshops}, pages 0--0, 2019.

\bibitem{ruiz2018fine}
Nataniel Ruiz, Eunji Chong, and James~M Rehg.
\newblock Fine-grained head pose estimation without keypoints.
\newblock In {\em CVPR Workshops}, pages 2074--2083, 2018.

\bibitem{salimans2016improved}
Tim Salimans, Ian Goodfellow, Wojciech Zaremba, Vicki Cheung, Alec Radford, and
  Xi Chen.
\newblock Improved techniques for training gans.
\newblock In {\em NeurIPS}, pages 2234--2242, 2016.

\bibitem{sandler2018mobilenetv2}
Mark Sandler, Andrew Howard, Menglong Zhu, Andrey Zhmoginov, and Liang-Chieh
  Chen.
\newblock Mobilenetv2: Inverted residuals and linear bottlenecks.
\newblock In {\em CVPR}, pages 4510--4520, 2018.

\bibitem{shen2020interpreting}
Yujun Shen, Jinjin Gu, Xiaoou Tang, and Bolei Zhou.
\newblock Interpreting the latent space of gans for semantic face editing.
\newblock In {\em CVPR}, 2020.

\bibitem{viazovetskyi2020stylegan2}
Yuri Viazovetskyi, Vladimir Ivashkin, and Evgeny Kashin.
\newblock Stylegan2 distillation for feed-forward image manipulation.
\newblock {\em arXiv preprint arXiv:2003.03581}, 2020.

\bibitem{wang2018pix2pixHD}
Ting-Chun Wang, Ming-Yu Liu, Jun-Yan Zhu, Andrew Tao, Jan Kautz, and Bryan
  Catanzaro.
\newblock High-resolution image synthesis and semantic manipulation with
  conditional gans.
\newblock In {\em CVPR}, pages 8798--8807, 2018.

\bibitem{wang2019controlling}
Yaxing Wang, Abel Gonzalez-Garcia, Joost van~de Weijer, and Luis Herranz.
\newblock Controlling biases and diversity in diverse image-to-image
  translation.
\newblock {\em arXiv preprint arXiv:1907.09754}, 2019.

\bibitem{wu2019relgan}
Po-Wei Wu, Yu-Jing Lin, Che-Han Chang, Edward~Y Chang, and Shih-Wei Liao.
\newblock Relgan: Multi-domain image-to-image translation via relative
  attributes.
\newblock In {\em ICCV}, pages 5914--5922, 2019.

\bibitem{xiao2018dna}
Taihong Xiao, Jiapeng Hong, and Jinwen Ma.
\newblock Dna-gan: Learning disentangled representations from multi-attribute
  images.
\newblock In {\em ICLR, Workshop}, 2018.

\bibitem{xiao2018elegant}
Taihong Xiao, Jiapeng Hong, and Jinwen Ma.
\newblock Elegant: Exchanging latent encodings with gan for transferring
  multiple face attributes.
\newblock In {\em ECCV}, pages 172--187, September 2018.

\bibitem{yu2019free}
Jiahui Yu, Zhe Lin, Jimei Yang, Xiaohui Shen, Xin Lu, and Thomas~S Huang.
\newblock Free-form image inpainting with gated convolution.
\newblock In {\em CVPR}, pages 4471--4480, 2019.

\bibitem{yu2019dmit}
Xiaoming Yu, Yuanqi Chen, Shan Liu, Thomas Li, and Ge Li.
\newblock Multi-mapping image-to-image translation via learning
  disentanglement.
\newblock In {\em NeurIPS}, pages 2994--3004, 2019.

\bibitem{zhang2018lpisp}
Richard Zhang, Phillip Isola, Alexei~A Efros, Eli Shechtman, and Oliver Wang.
\newblock The unreasonable effectiveness of deep features as a perceptual
  metric.
\newblock In {\em CVPR}, pages 586--595, 2018.

\bibitem{zhou2017genegan}
Shuchang Zhou, Taihong Xiao, Yi Yang, Dieqiao Feng, Qinyao He, and Weiran He.
\newblock Genegan: Learning object transfiguration and attribute subspace from
  unpaired data.
\newblock In {\em BMVC}, 2017.

\bibitem{zhu2017cyclegan}
Jun-Yan Zhu, Taesung Park, Phillip Isola, and Alexei~A Efros.
\newblock Unpaired image-to-image translation using cycle-consistent
  adversarial networks.
\newblock In {\em ICCV}, pages 2223--2232, 2017.

\bibitem{zhu2017bicycle}
Jun-Yan Zhu, Richard Zhang, Deepak Pathak, Trevor Darrell, Alexei~A Efros,
  Oliver Wang, and Eli Shechtman.
\newblock Toward multimodal image-to-image translation.
\newblock In {\em NeurIPS}, pages 465--476, 2017.

\bibitem{zhu2020sean}
Peihao Zhu, Rameen Abdal, Yipeng Qin, and Peter Wonka.
\newblock Sean: Image synthesis with semantic region-adaptive normalization.
\newblock In {\em CVPR}, pages 5104--5113, 2020.

\end{thebibliography}
}

\clearpage
\appendix
\begin{appendix}

\section{Network Architectures}
\label{appendix:architectures}
In ~\tref{table:sem_network_generator}, \ref{table:sem_network_discriminator} and \ref{table:sem_network_mapping}, we describe our networks for the semantic manipulation stage. Similarly, \tref{table:syn_network_generator}, \ref{table:syn_network_encoder}, and \ref{table:syn_network_mapping} shows the description for the image synthesis stage. For the latter, we bypass the description of the discriminator as it is a copy of the StyleGAN2 one. For both stages, and in order to maintain both environments as unaltered as possible, we use the same training hyperparameters (Adam Beta optimizers and learning rates) as in the corresponding baselines StarGANv2 and StyleGAN2, respectively.

\begin{table*}[t]
\setlength{\tabcolsep}{7pt}
\renewcommand{\arraystretch}{1.5}
\begin{center}
\begin{tabular}{c  c  c}
Part & Input $\rightarrow$ Output Shape & Layer Information \\
\hline \hline
\multirow{8}{*}{Down-sampling}
& $(256, 256, \mathbb{N}_{mask}) \rightarrow (256, 256, 32)$ & Conv2d(dim\_out=32, kernel=3, stride=1, padding=1) \\
& $(256, 256, 32) \rightarrow (128, 128, 64)$ & Residual Block: IN, LReLU, Conv2d(64, 4, 2, 1), AvgPool2D \\
& $(128, 128, 64) \rightarrow (64, 64, 128)$ & Residual Block: IN, LReLU, Conv2d(128, 4, 2, 1), AvgPool2D \\
& $(64, 64, 128) \rightarrow (32, 32, 256)$ & Residual Block: IN, LReLU, Conv2d(256, 4, 2, 1), AvgPool2D \\
& $(32, 32, 256) \rightarrow (16, 16, 512)$ & Residual Block: IN, LReLU, Conv2d(256, 4, 2, 1), AvgPool2D \\
& $(16, 16, 512) \rightarrow (8, 8, 512)$ & Residual Block: IN, ReLU, Conv2d(256, 3, 1, 1), AvgPool2D \\
& $(8, 8, 512) \rightarrow (8, 8, 512)$ & Residual Block: IN, LReLU, Conv2d(256, 3, 1, 1) \\
& $(8, 8, 512) \rightarrow (8, 8, 512)$ & Residual Block: IN, LReLU, Conv2d(256, 3, 1, 1)  \\
\midrule
\multirow{8}{*}{Up-sampling} 
 & $(8, 8, 512) \rightarrow (8, 8, 512)$ &  LReLU, ModulatedConv2d(256, 3, 1, 1 \\
 & $(8, 8, 512) \rightarrow (8, 8, 512)$ &  LReLU, ModulatedConv2d(256, 3, 1, 1) \\
 & $(8, 8, 512) \rightarrow (16, 16, 512)$  & LReLU, NearestUp, ModulatedConv2d(128, 3, 1, 1) \\
 & $(16, 16, 256) \rightarrow (32, 32, 256)$  & LReLU, NearestUp, ModulatedConv2d(128, 3, 1, 1) \\
 & $(32, 32, 256) \rightarrow (64, 64, 128)$  & LReLU, NearestUp, ModulatedConv2d(128, 3, 1, 1) \\
 & $(64, 64, 128) \rightarrow (128, 128, 64)$ & LReLU, NearestUp, ModulatedConv2d(64, 3, 1, 1) \\
 & $(128, 128, 64) \rightarrow (256, 256, 32)$ & LReLU, NearestUp, ModulatedConv2d(32, 3, 1, 1) \\ 
 & $(256, 256, 32) \rightarrow (256, 256, \mathbb{N}_{mask})$ & ReLU, ModulatedConv2d($\mathbb{N}_{mask}$, 3, 1, 1) \\
\hline
\hline
\end{tabular}
\end{center}
\caption{\textbf{\SHORTTITLE{} Semantic Manipulation Generator network architecture.} As we use the CelebA-Mask dataset~\cite{lee2020maskgan}, we set $\mathbb{N}_{mask}$ to 19.}
\label{table:sem_network_generator}
\end{table*}

\begin{table*}[t]
\setlength{\tabcolsep}{7pt}
\renewcommand{\arraystretch}{1.7}
\begin{center}
\begin{tabular}{c c c}
Layer & Input $\rightarrow$ Output Shape & Layer Information \\
\hline \hline
Input Layer & $(256, 256, \mathbb{N}_{mask}) \rightarrow (256, 256, 64) $ & Conv2d(dim\_out=64, kernel=3, stride=1, padding=1), LReLU \\
\Xhline{1.0pt}
Hidden Layer & $(256, 256, 64) \rightarrow (128, 128, 128) $ & Residual Block: LReLU, Conv2d(128, 3, 1, 1), AvgPool2D \\
Hidden Layer & $(128, 128, 128) \rightarrow (64, 64, 256) $ & Residual Block: LReLU, Conv2d(128, 3, 1, 1), AvgPool2D \\
Hidden Layer & $(64, 64, 256) \rightarrow (32, 32, 512) $ & Residual Block: LReLU, Conv2d(128, 3, 1, 1), AvgPool2D \\
Hidden Layer & $(32, 32, 512) \rightarrow (16, 16, 512) $ & Residual Block: LReLU, Conv2d(256, 3, 1, 1), AvgPool2D \\
Hidden Layer & $(16, 16, 256) \rightarrow (8, 8, 512) $ & Residual Block: LReLU, Conv2d(512, 3, 1, 1), AvgPool2D \\
Hidden Layer & $(8, 8, 512) \rightarrow (4, 4, 512) $ & Residual Block: LReLU, Conv2d(512, 3, 1, 1), AvgPool2D \\
Hidden Layer & $(4, 4, 512) \rightarrow (1, 1, 512) $ & LReLU, Conv2d(512, 4, 1, 0) \\
\Xhline{1.0pt}
Output Layer & $(1, 1, 512) \rightarrow (1, 1, d \times N \times 2) $ & LReLU, DS\_Layer(512, d$\times$N$\times$2, 1, 0) \\
\hline \hline
\end{tabular}
\end{center}
\caption{\textbf{\SHORTTITLE{} Semantic Manipulation Discriminator and Style Encoder network architecture.} The difference between the two networks lie in the last layer, where \textit{DS\_Layer} is a Convolution and Linear layer and \textit{d} is the number of dimensions equal to 1 and 64 for Discriminator and Style Encoder, respectively. As we consider each domain as the presence or absence of each attribute, the number of outputs is scaled by 2. Note that as part of our method we weight the style dimensions for Male/Female, so all the remaining are in a 4:1 ratio. }
\label{table:sem_network_discriminator}
\end{table*}

\begin{table*}[t]
\setlength{\tabcolsep}{7pt}
\renewcommand{\arraystretch}{1.7}
\begin{center}
\begin{tabular}{c c c}
Layer & Input $\rightarrow$ Output Shape & Layer Information \\
\hline \hline
\multirow{4}{*}{Shared Layers}
& $(16) \rightarrow (512) $ & Linear(dim\_out=512), ReLU\\
& $(512) \rightarrow (512) $ & Linear(dim\_out=512), ReLU\\
& $(512) \rightarrow (512) $ & Linear(dim\_out=512), ReLU\\
& $(512) \rightarrow (512) $ & Linear(dim\_out=512), ReLU\\
\midrule
\multirow{4}{*}{Unshared Layers}
& $(512) \rightarrow (512) $ & Linear(dim\_out=512), ReLU\\
& $(512) \rightarrow (512) $ & Linear(dim\_out=512), ReLU\\
& $(512) \rightarrow (512) $ & Linear(dim\_out=512), ReLU\\
& $(512) \rightarrow (WS) $ & Linear(dim\_out=WS)\\
\hline \hline
\end{tabular}
\end{center}
\caption{\textbf{\SHORTTITLE{} Semantic Manipulation Mapping network architecture.} All attributes share the first part of the network, and each attribute has independent branch of unshared layers. \textit{WS} stands for the Weighted Style to each attribute. For Male/Female we set 64, and 16 for other attributes. Not that presence and absence of each attribute have two independent branches in the mapping network.}
\label{table:sem_network_mapping}
\end{table*}

\begin{table*}[t]
\setlength{\tabcolsep}{7pt}
\renewcommand{\arraystretch}{1.5}
\begin{center}
\begin{tabular}{c  c  c}
Part & Input $\rightarrow$ Output Shape & Layer Information \\
\hline \hline
\multirow{5}{*}{Mask Feature Extractor} 
 & $(64, 64, \mathbb{N}_{mask}) \rightarrow (64, 64, 256) $ & EqualConv2d(dim\_out=32, kernel=3, stride=1, padding=1) \\
 & $(64, 64, 256) \rightarrow (64, 64, 256) $ & FusedLeakyReLU \\
 & $(64, 64, 256) \rightarrow (32, 32, 256) $ & Blur, EqualConv2d(256, 3, 2, 0), FusedLeakyReLU \\
 & $(32, 32, 256) \rightarrow (16, 16, 256) $ & Blur, EqualConv2d(256, 3, 2, 0), FusedLeakyReLU \\
 & $(16, 16, 256) \rightarrow (8, 8, 512) $ & Blur, EqualConv2d(512, 3, 2, 0), FusedLeakyReLU \\ 
\Xhline{1.0pt}
\multirow{12}{*}{Up-sampling} 
 & $(8, 8, 512) \rightarrow (8, 8, 512)$ &  SAC(dim\_out=512, kernel=3, upsample=False) \\
 & $(8, 8, 512) \rightarrow (8, 8, 512)$ & Noise, FusedLeakyReLU \\
 & $(8, 8, 512) \rightarrow (16, 16, 512)$ &  SAC(512, 3, True), Noise, FusedLeakyReLU \\
 & $(16, 16, 512) \rightarrow (16, 16, 512)$ &  SAC(512, 3, False), Noise, FusedLeakyReLU \\
 & $(16, 16, 512) \rightarrow (32, 32, 512)$ &  SAC(512, 3, True), Noise, FusedLeakyReLU \\
 & $(32, 32, 512) \rightarrow (32, 32, 512)$ &  SAC(512, 3, False), Noise, FusedLeakyReLU \\
 & $(32, 32, 512) \rightarrow (64, 64, 256)$ &  SAC(256, 3, True), Noise, FusedLeakyReLU \\
 & $(64, 64, 256) \rightarrow (64, 64, 256)$ &  SAC(256, 3, False), Noise, FusedLeakyReLU \\
 & $(64, 64, 256) \rightarrow (128, 128, 256)$ &  SAC(256, 3, True), Noise, FusedLeakyReLU \\
 & $(128, 128, 128) \rightarrow (128, 128, 256)$ &  SAC(256, 3, False), Noise, FusedLeakyReLU \\
 & $(128, 128, 256) \rightarrow (256, 256, 128)$ &  SAC*(256, 3, True), Noise, FusedLeakyReLU \\
 & $(256, 256, 128) \rightarrow (256, 256, 128)$ &  SAC*(256, 3, False), Noise, FusedLeakyReLU \\
\Xhline{1.0pt}
Output Layer & $(256, 256, 128) \rightarrow (256, 256, 3)$ &  SAC*(256, 3, False) \\
\hline
\hline
\end{tabular}
\end{center}
\caption{\textbf{\SHORTTITLE{} Semantically-driven Image Synthesis Generator network architecture.} We leverage on the StyleGAN2~\cite{karras2020analyzing} architecture with minor yet significant modifications. We replace the modulated convolutions with our Semantically Adaptive Convolutions (SACs) introduced in \eref{equation:sacs}. Additionally, for the last three layers (SACs*) we only introduce the semantics in a SPADE fashion. EqualConv2d, Blur, Noise and FusedLeakyReLU are mirror layers from StyleGAN~\cite{karras2019style}. For the most part of the coding, we heavily borrow from \url{https://github.com/rosinality/stylegan2-pytorch}.}
\label{table:syn_network_generator}
\end{table*}

\begin{table*}[t]
\setlength{\tabcolsep}{7pt}
\renewcommand{\arraystretch}{1.7}
\begin{center}
\begin{tabular}{c c c}
Layer & Input $\rightarrow$ Output Shape & Layer Information \\
\hline \hline
Input Layer & $(256, 256, 3) \rightarrow (256, 256, 32) $ & EqualConv2d(dim\_out=32, kernel=3, stride=1, padding=1), FusedLeakyReLU \\
\Xhline{1.0pt}
Hidden Layer & $(256, 256, 32) \rightarrow (128, 128, 64) $ & Blur, EqualConv2d(64, 3, 2, 0), FusedLeakyReLU \\
Hidden Layer & $(128, 128, 64) \rightarrow (64, 64, 128) $ & Blur, EqualConv2d(128, 3, 2, 0), FusedLeakyReLU \\
Hidden Layer & $(64, 64, 128) \rightarrow (128, 128, 256) $ & EqualConv2d(256, 3, 2, 0, upsample=True), Blur, FusedLeakyReLU \\
Hidden Layer & $(128, 128, 256) \rightarrow (64, 64, \mathbb{N}_{s}^{syn}) $ & Blur, EqualConv2d(512, 3, 2, 0), FusedLeakyReLU \\
\Xhline{1.0pt}
Output Layer & $(64, 64, \mathbb{N}_{s}^{syn}) \rightarrow (\mathbb{N}_{mask}, \mathbb{N}_{s}^{syn}) $ & Mask Average Pooling \\
\hline \hline
\end{tabular}
\end{center}
\caption{\textbf{\SHORTTITLE{} Semantically-driven Image Synthesis Style Encoder network architecture.} Using the RGB and Semantic Mask ($\mathbb{N}_{mask}: 19$) as input, it outputs a per region style with dimentionality $\mathbb{N}_{s}^{syn}: 64$. EqualConv2D, FusedLeakyReLU and Blur are layers borrowed from StyleGAN~\cite{karras2019style}. Mask Average Pooling combines the semantic information with the style encoded features.}
\label{table:syn_network_encoder}
\end{table*}

\begin{table*}[t]
\setlength{\tabcolsep}{7pt}
\renewcommand{\arraystretch}{1.7}
\begin{center}
\begin{tabular}{c c c}
Layer & Input $\rightarrow$ Output Shape & Layer Information \\
\hline \hline
Input Layer & $(\mathbb{N}_{mask}\times \mathbb{N}_{s}^{syn}) \rightarrow (\mathbb{N}_{mask}\times \mathbb{N}_{s}^{syn}) $ & PixelNorm \\
\Xhline{1.0pt}
& $(\mathbb{N}_{mask}\times \mathbb{N}_{s}^{syn}) \rightarrow (\mathbb{N}_{mask}\times \mathbb{N}_{s}^{syn}) $ & EqualLinear(dim\_out=$(\mathbb{N}_{mask}\times \mathbb{N}_{s}^{syn})$) \\
& $(\mathbb{N}_{mask}\times \mathbb{N}_{s}^{syn}) \rightarrow (\mathbb{N}_{mask}\times \mathbb{N}_{s}^{syn}) $ & EqualLinear(dim\_out=$(\mathbb{N}_{mask}\times \mathbb{N}_{s}^{syn})$) \\
& $(\mathbb{N}_{mask}\times \mathbb{N}_{s}^{syn}) \rightarrow (\mathbb{N}_{mask}\times \mathbb{N}_{s}^{syn}) $ & EqualLinear(dim\_out=$(\mathbb{N}_{mask}\times \mathbb{N}_{s}^{syn})$) \\
& $(\mathbb{N}_{mask}\times \mathbb{N}_{s}^{syn}) \rightarrow (\mathbb{N}_{mask}\times \mathbb{N}_{s}^{syn}) $ & EqualLinear(dim\_out=$(\mathbb{N}_{mask}\times \mathbb{N}_{s}^{syn})$) \\& $(\mathbb{N}_{mask}\times \mathbb{N}_{s}^{syn}) \rightarrow (\mathbb{N}_{mask}\times \mathbb{N}_{s}^{syn}) $ & EqualLinear(dim\_out=$(\mathbb{N}_{mask}\times \mathbb{N}_{s}^{syn})$) \\& $(\mathbb{N}_{mask}\times \mathbb{N}_{s}^{syn}) \rightarrow (\mathbb{N}_{mask}\times \mathbb{N}_{s}^{syn}) $ & EqualLinear(dim\_out=$(\mathbb{N}_{mask}\times \mathbb{N}_{s}^{syn})$) \\
& $(\mathbb{N}_{mask}\times \mathbb{N}_{s}^{syn}) \rightarrow (\mathbb{N}_{mask}\times \mathbb{N}_{s}^{syn}) $ & EqualLinear(dim\_out=$(\mathbb{N}_{mask}\times \mathbb{N}_{s}^{syn})$) \\
\Xhline{1.0pt}
& $(\mathbb{N}_{mask}\times \mathbb{N}_{s}^{syn}) \rightarrow (\mathbb{N}_{mask}\times \mathbb{N}_{s}^{syn}) $ & EqualLinear(dim\_out=$(\mathbb{N}_{mask}\times \mathbb{N}_{s}^{syn})$) \\
\hline \hline
\end{tabular}
\end{center}
\caption{\textbf{\SHORTTITLE{} Semantically-driven Image Synthesis Mapping network architecture.} The input is sampled from a gaussian distribution, and the output is also known as the style distribution $\mathbb{W}$. PixelNorm and EqualLinear are pixel normalization and linearly normalized layers introduced in StyleGAN~\cite{karras2019style}, respectively. }
\label{table:syn_network_mapping}
\end{table*}

\section{Additional Results for Semantic Manipulation}
\label{appendix:additional_sem}
In this section we present additional results for the semantic manipulation. We report FID, LPIPS, F1, AP, and other metrics reported in the main paper for each attribute manipulation. 

In~\tref{table:additional_sem_ref} and~\ref{table:additional_sem_lat}, we show quantitative evaluation for each attribute independently. Not surprisingly, Male and Female attributes are easier manipulations than hat and earrings. We set a benchmark for the evaluation of this manipulation. 

\begin{figure*}[t]
    \centering%
    \begin{minipage}{.8\textwidth}
        \begin{minipage}[c]{0.475\textwidth}
            \includegraphics[width=\textwidth]{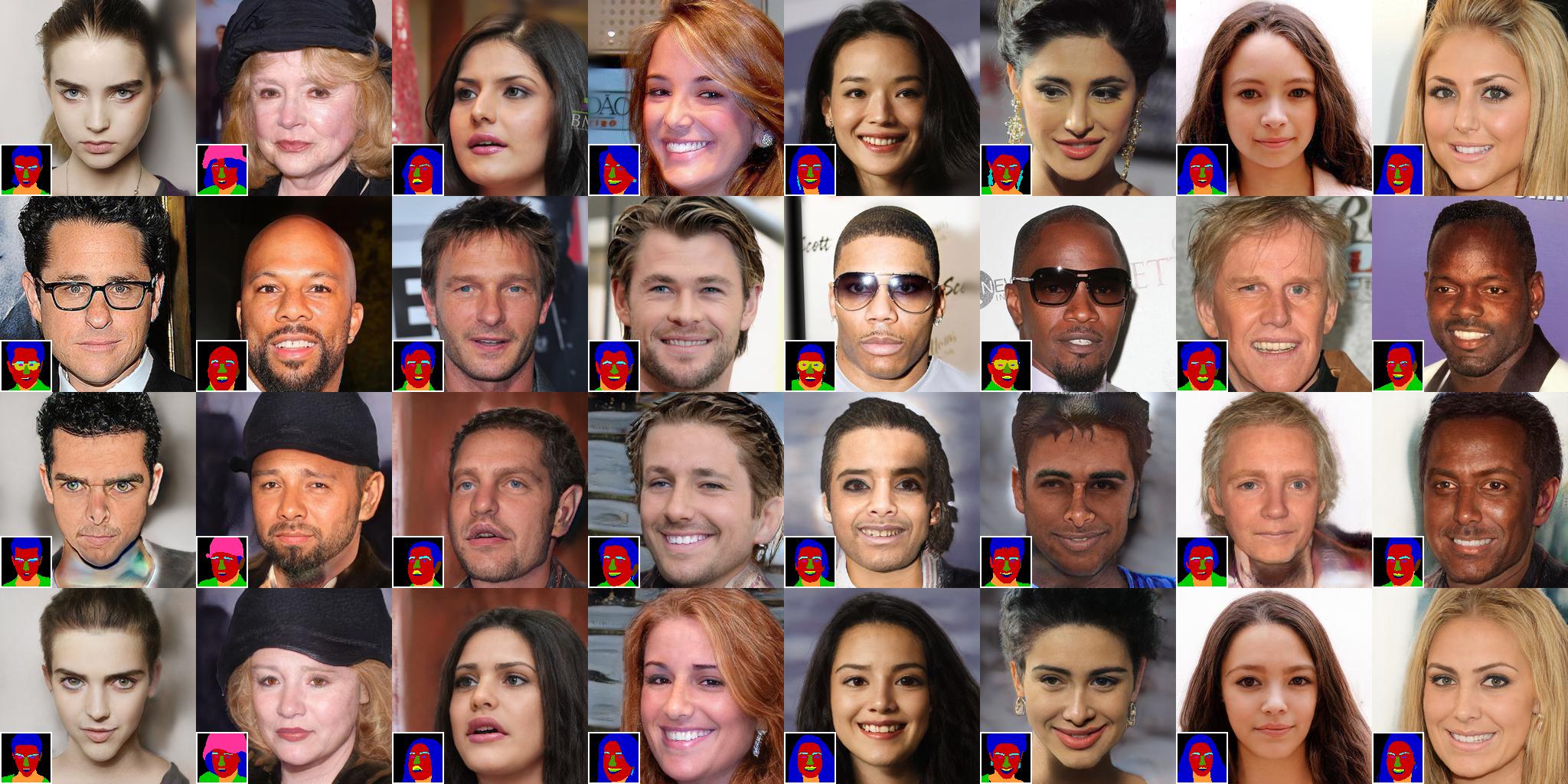} \subcaption{Reconstruction}
            
            \includegraphics[width=\textwidth]{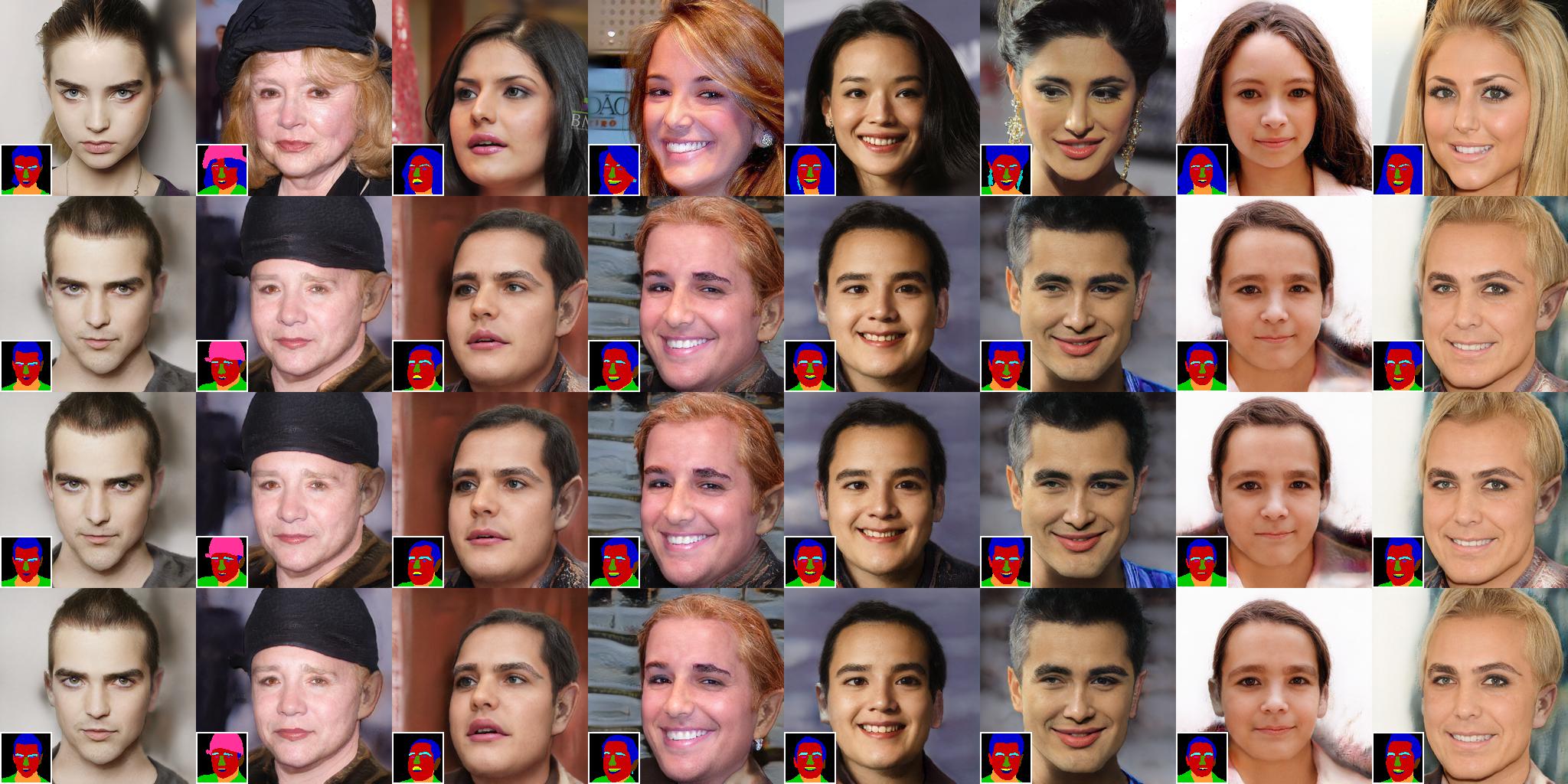}
            \subcaption{Latent Synthesis}        
        \end{minipage}    
        \hfill\vline\hfill
        \begin{minipage}[c]{0.515\textwidth}
            \includegraphics[width=\textwidth]{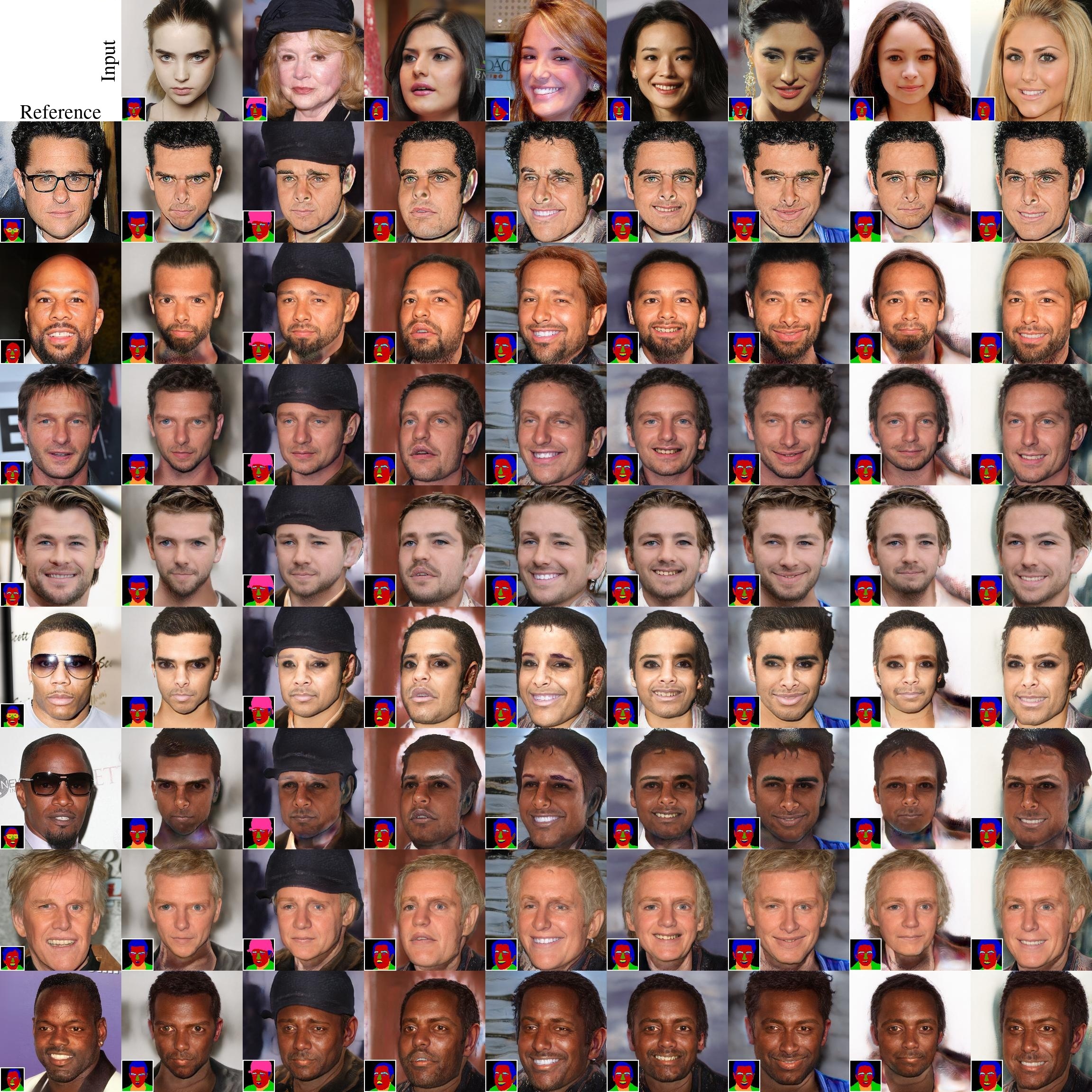}
            \subcaption{Reference Synthesis}
        \end{minipage}        
    \end{minipage}    
    \caption{\textbf{Qualitative Results for Male manipulation \underline{only}}. (a) Reconstruction results. First and second row are input and reference images, respectively. Third and fourth rows are forward and reconstruction outputs, respectively. (b) Latent synthesis generation for both semantic and rgb outputs. First row represents input images. (c) reference image synthesis for both semantic and rgb space.}
    \label{figure:qualitative_male}
\end{figure*}

\begin{figure*}[t]
    \centering%
    \begin{minipage}{.8\textwidth}
        \begin{minipage}[c]{0.475\textwidth}
            \includegraphics[width=\textwidth]{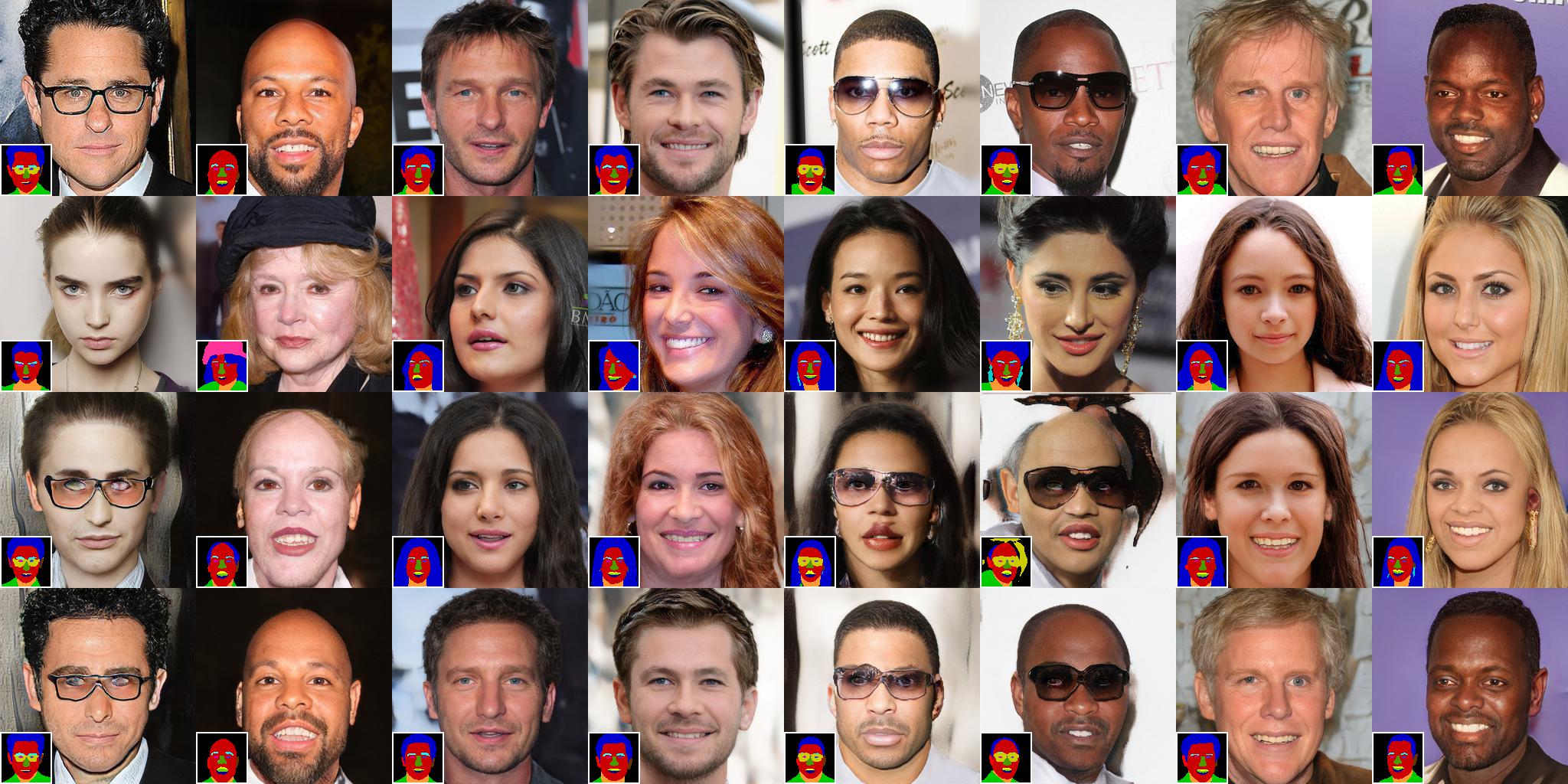} \subcaption{Reconstruction}
            
            \includegraphics[width=\textwidth]{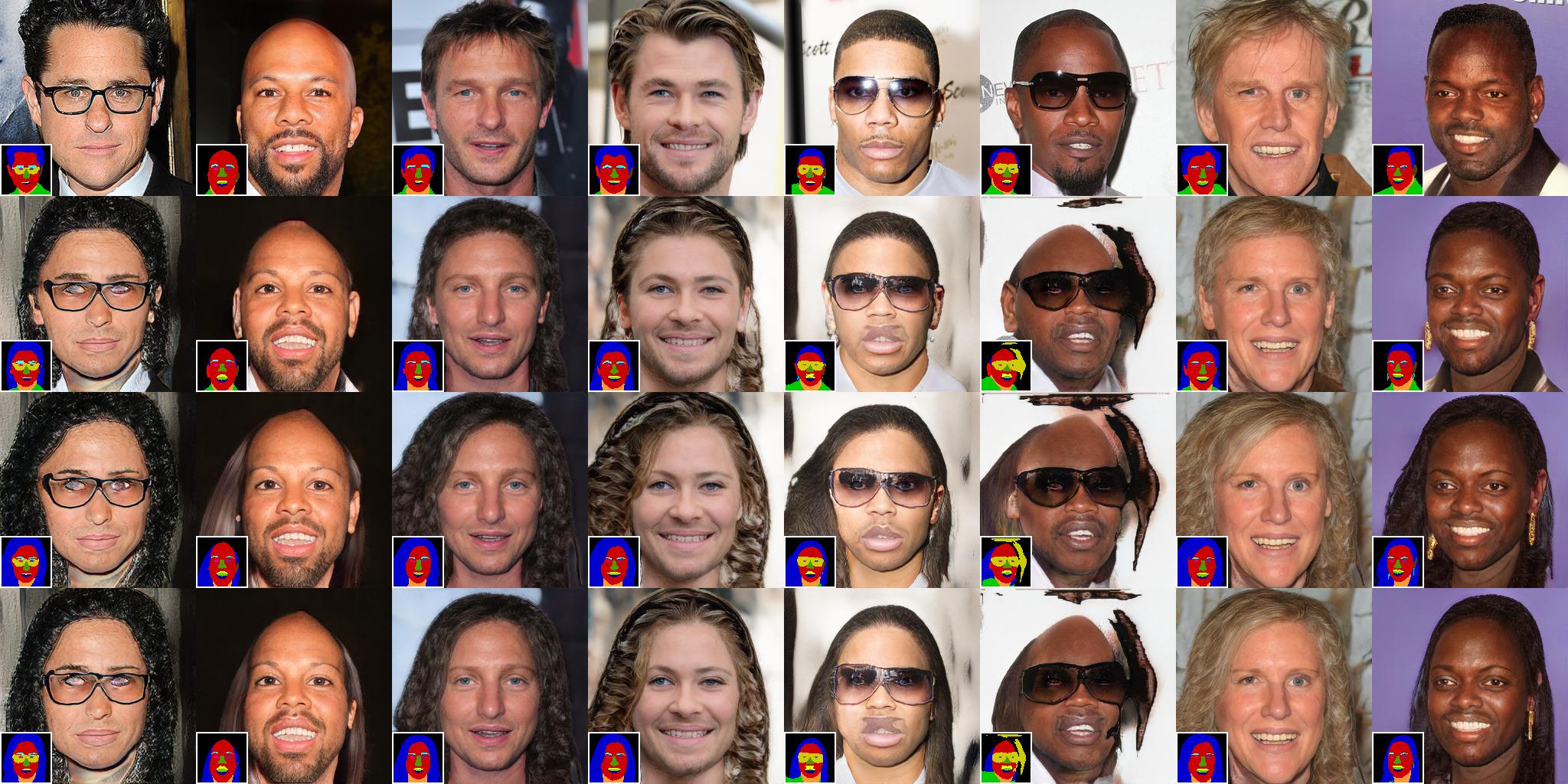}
            \subcaption{Latent Synthesis}        
        \end{minipage}    
        \hfill\vline\hfill
        \begin{minipage}[c]{0.515\textwidth}
            \includegraphics[width=\textwidth]{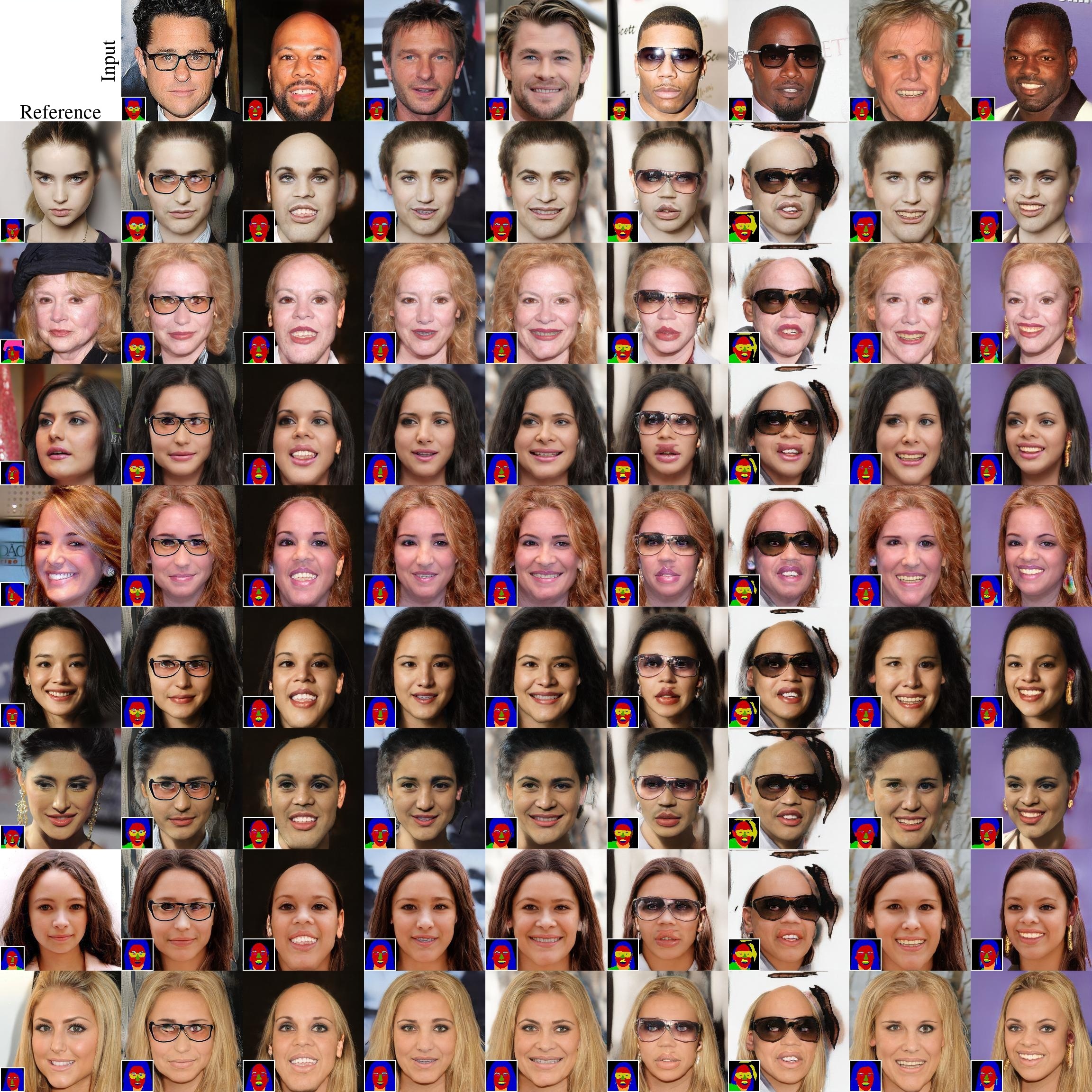}
            \subcaption{Reference Synthesis}
        \end{minipage}        
    \end{minipage}    
    \caption{\textbf{Qualitative Results for Female manipulation \underline{only}}. (a) Reconstruction results. First and second row are input and reference images, respectively. Third and fourth rows are forward and reconstruction outputs, respectively. (b) Latent synthesis generation for both semantic and rgb outputs. First row represents input images. (c) reference image synthesis for both semantic and rgb space.}
    \label{figure:qualitative_female}
\end{figure*}

\begin{figure*}[t]
    \centering%
    \begin{minipage}{.8\textwidth}
        \begin{minipage}[c]{0.475\textwidth}
            \includegraphics[width=\textwidth]{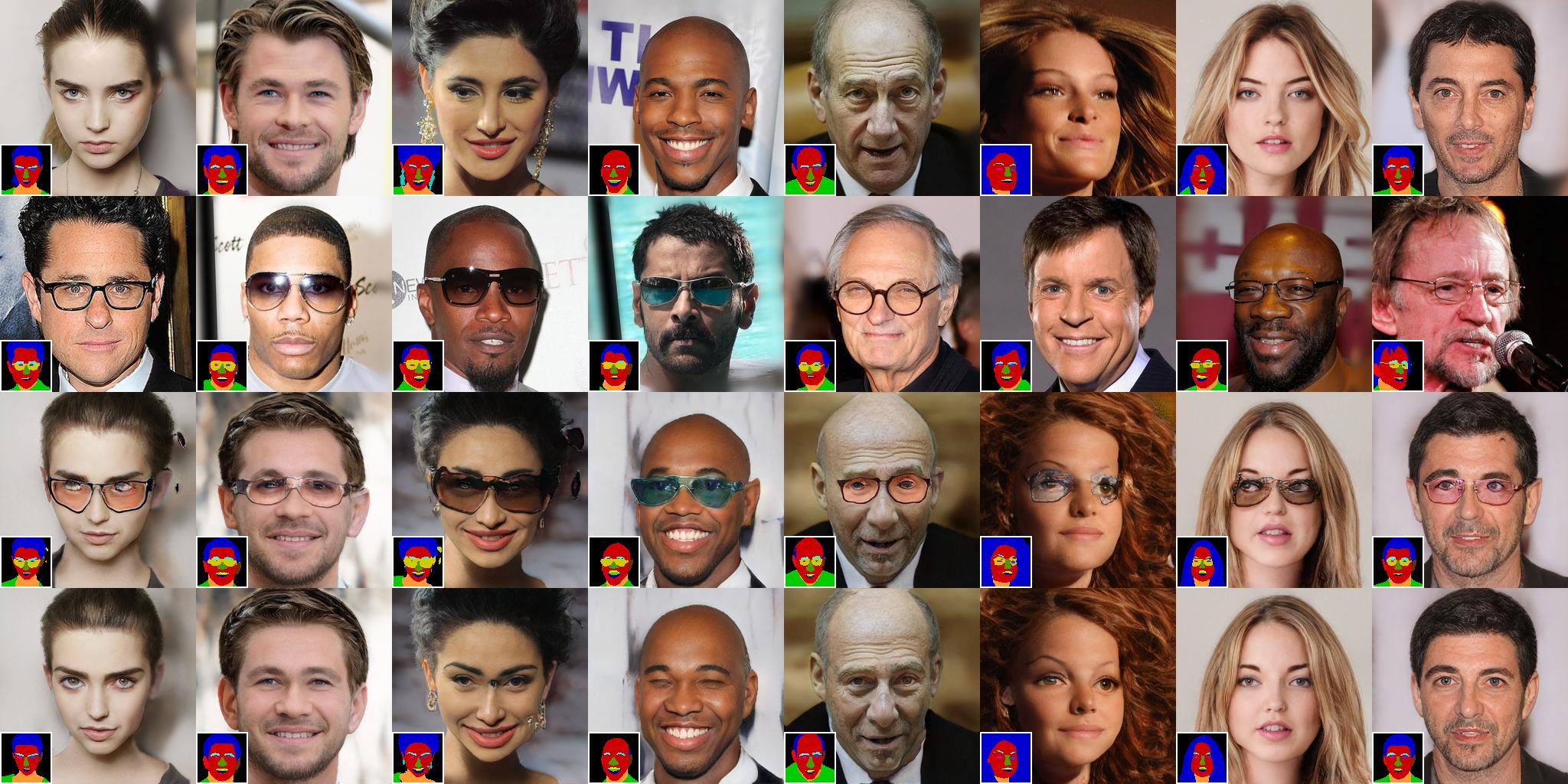} \subcaption{Reconstruction}
            
            \includegraphics[width=\textwidth]{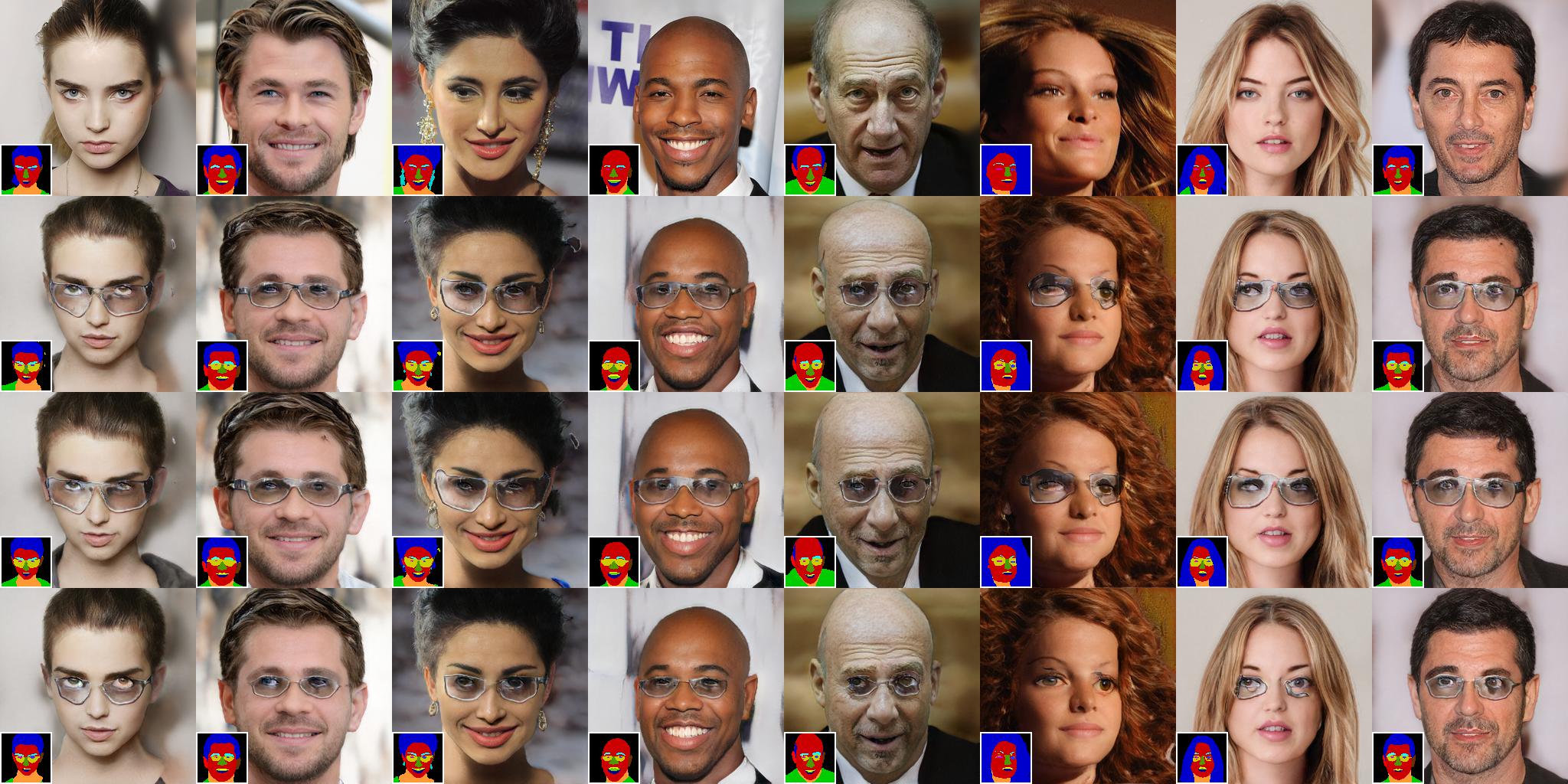}
            \subcaption{Latent Synthesis}        
        \end{minipage}    
        \hfill\vline\hfill
        \begin{minipage}[c]{0.515\textwidth}
            \includegraphics[width=\textwidth]{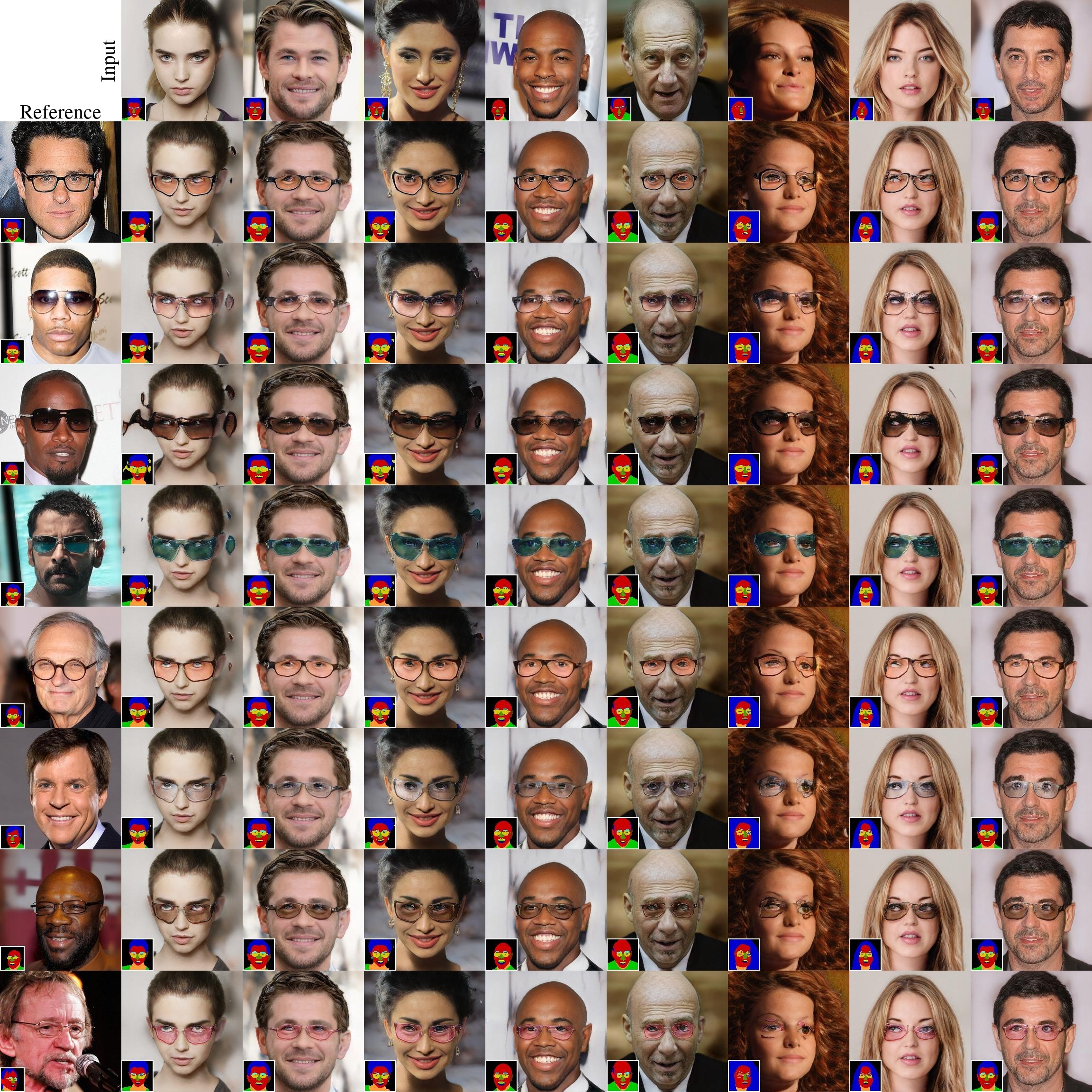}
            \subcaption{Reference Synthesis}
        \end{minipage}        
    \end{minipage}    
    \caption{\textbf{Qualitative Results for Eyeglasses manipulation \underline{only}}. (a) Reconstruction results. First and second row are input and reference images, respectively. Third and fourth rows are forward and reconstruction outputs, respectively. (b) Latent synthesis generation for both semantic and rgb outputs. First row represents input images. (c) reference image synthesis for both semantic and rgb space.}
    \label{figure:qualitative_eyeglasses}
\end{figure*}

\begin{figure*}[t]
    \centering%
    \begin{minipage}{.8\textwidth}
        \begin{minipage}[c]{0.475\textwidth}
            \includegraphics[width=\textwidth]{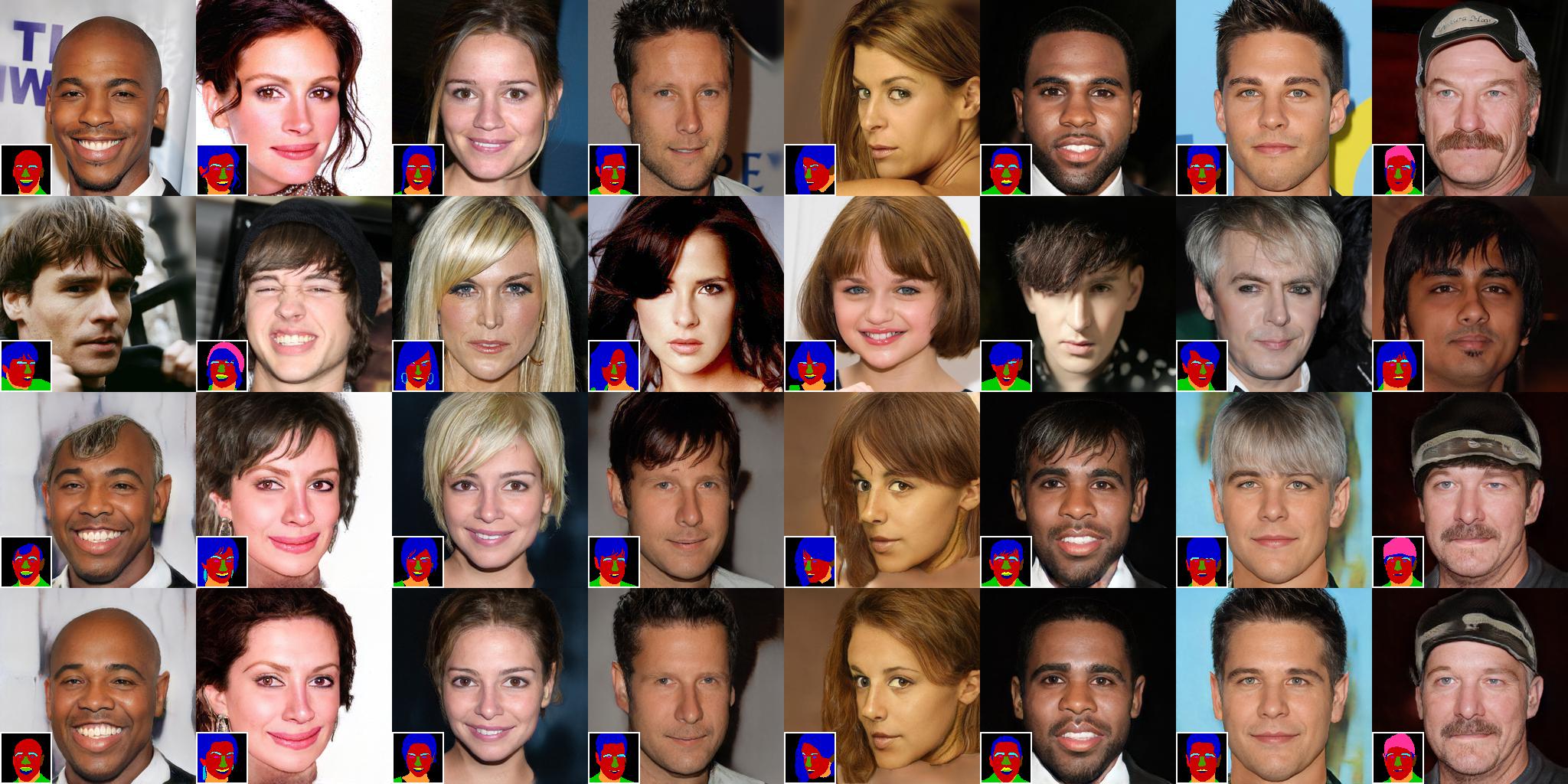} \subcaption{Reconstruction}
            
            \includegraphics[width=\textwidth]{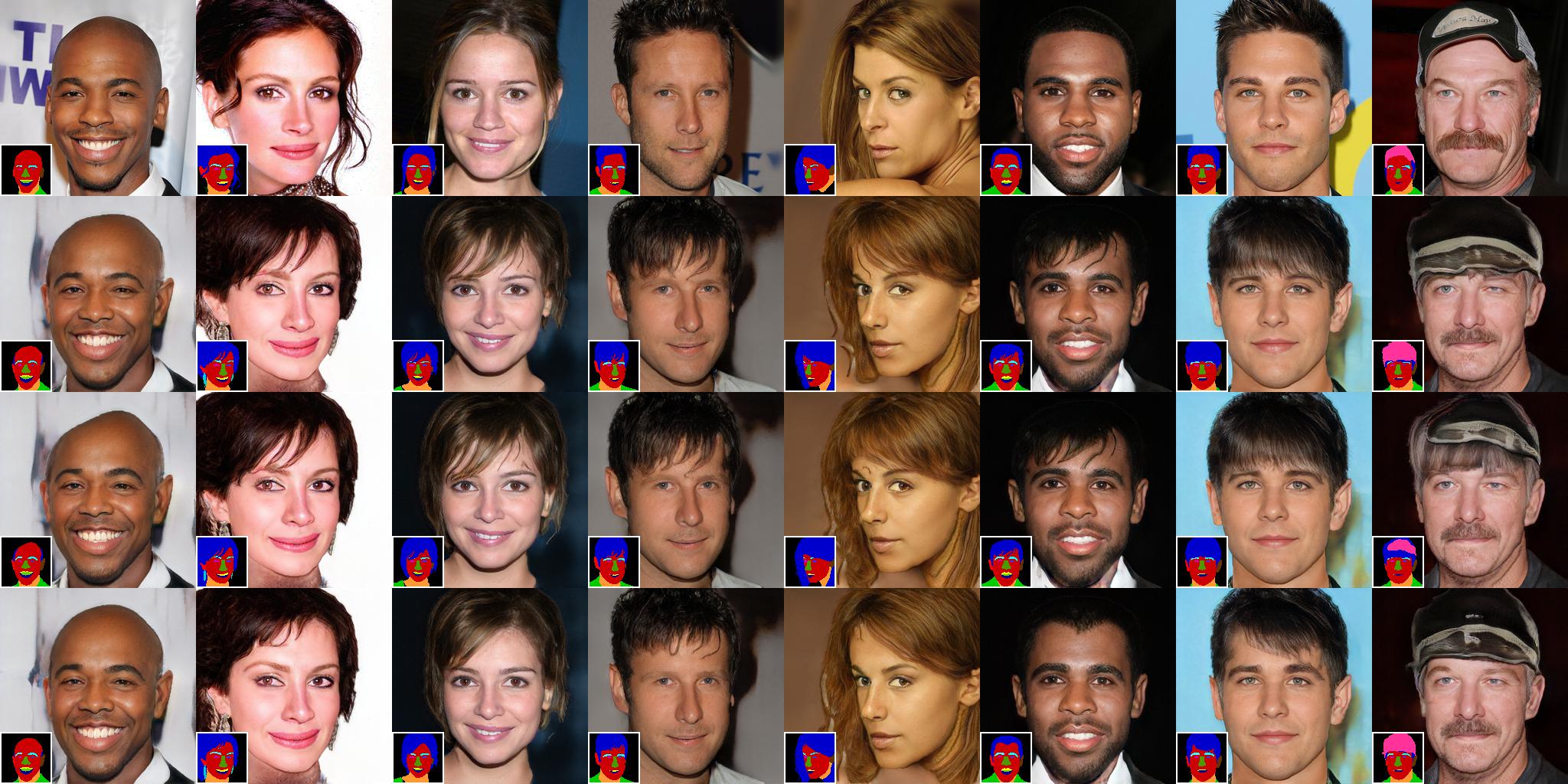}
            \subcaption{Latent Synthesis}        
        \end{minipage}    
        \hfill\vline\hfill
        \begin{minipage}[c]{0.515\textwidth}
            \includegraphics[width=\textwidth]{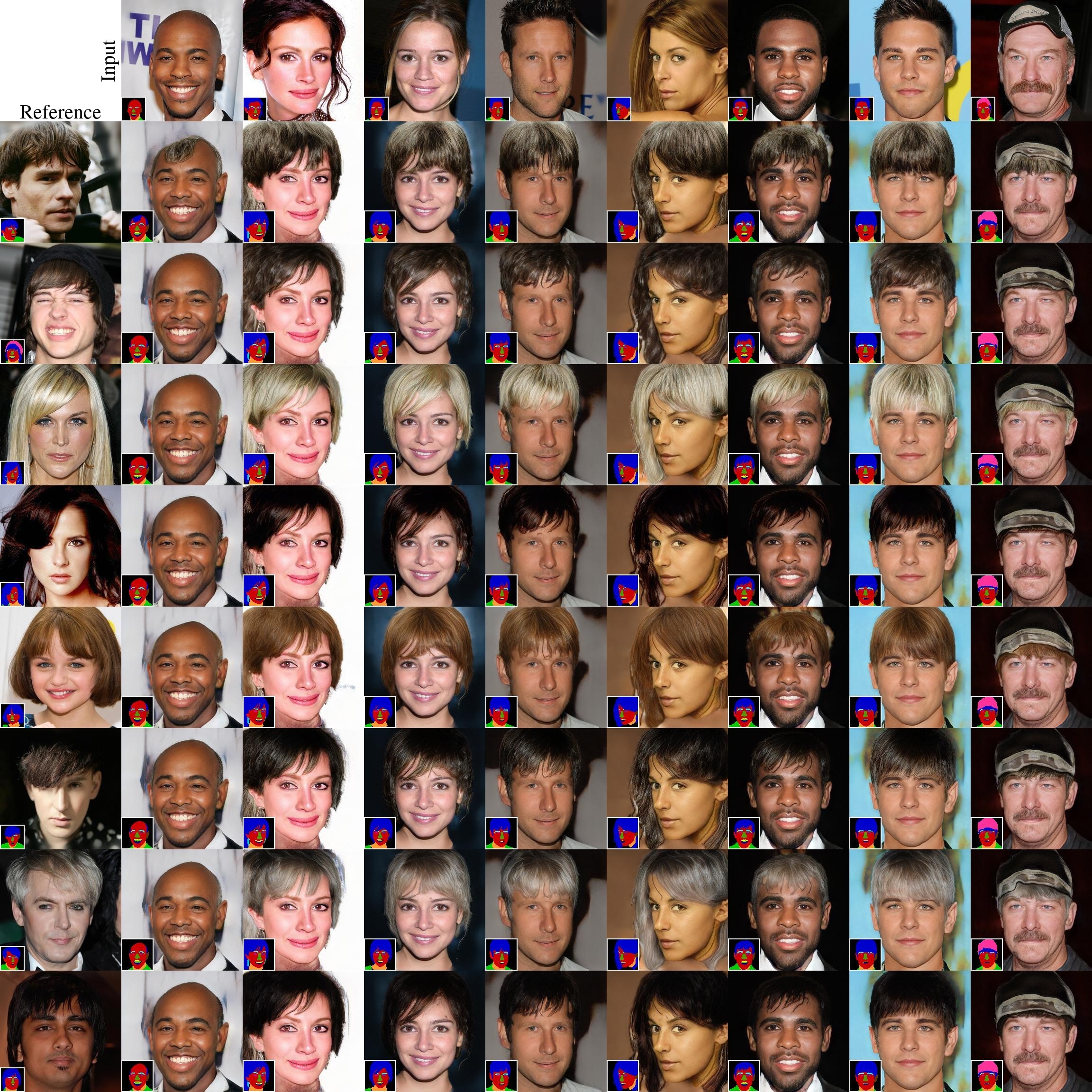}
            \subcaption{Reference Synthesis}
        \end{minipage}        
    \end{minipage}    
    \caption{\textbf{Qualitative Results for Bangs manipulation \underline{only}}. (a) Reconstruction results. First and second row are input and reference images, respectively. Third and fourth rows are forward and reconstruction outputs, respectively. (b) Latent synthesis generation for both semantic and rgb outputs. First row represents input images. (c) reference image synthesis for both semantic and rgb space.}
    \label{figure:qualitative_bangs}
\end{figure*}

\begin{figure*}[t]
    \centering%
    \begin{minipage}{.8\textwidth}
        \begin{minipage}[c]{0.475\textwidth}
            \includegraphics[width=\textwidth]{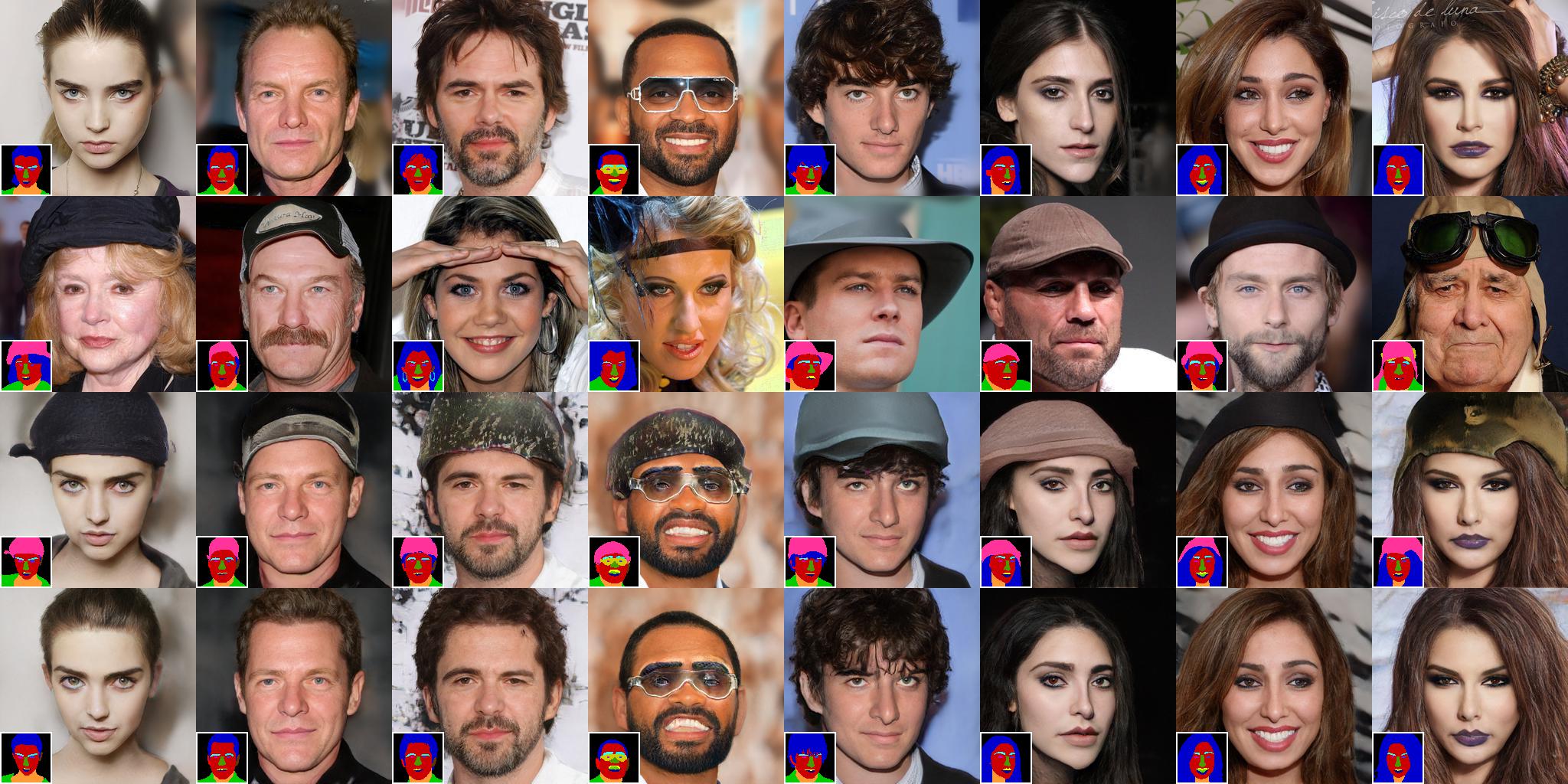} \subcaption{Reconstruction}
            
            \includegraphics[width=\textwidth]{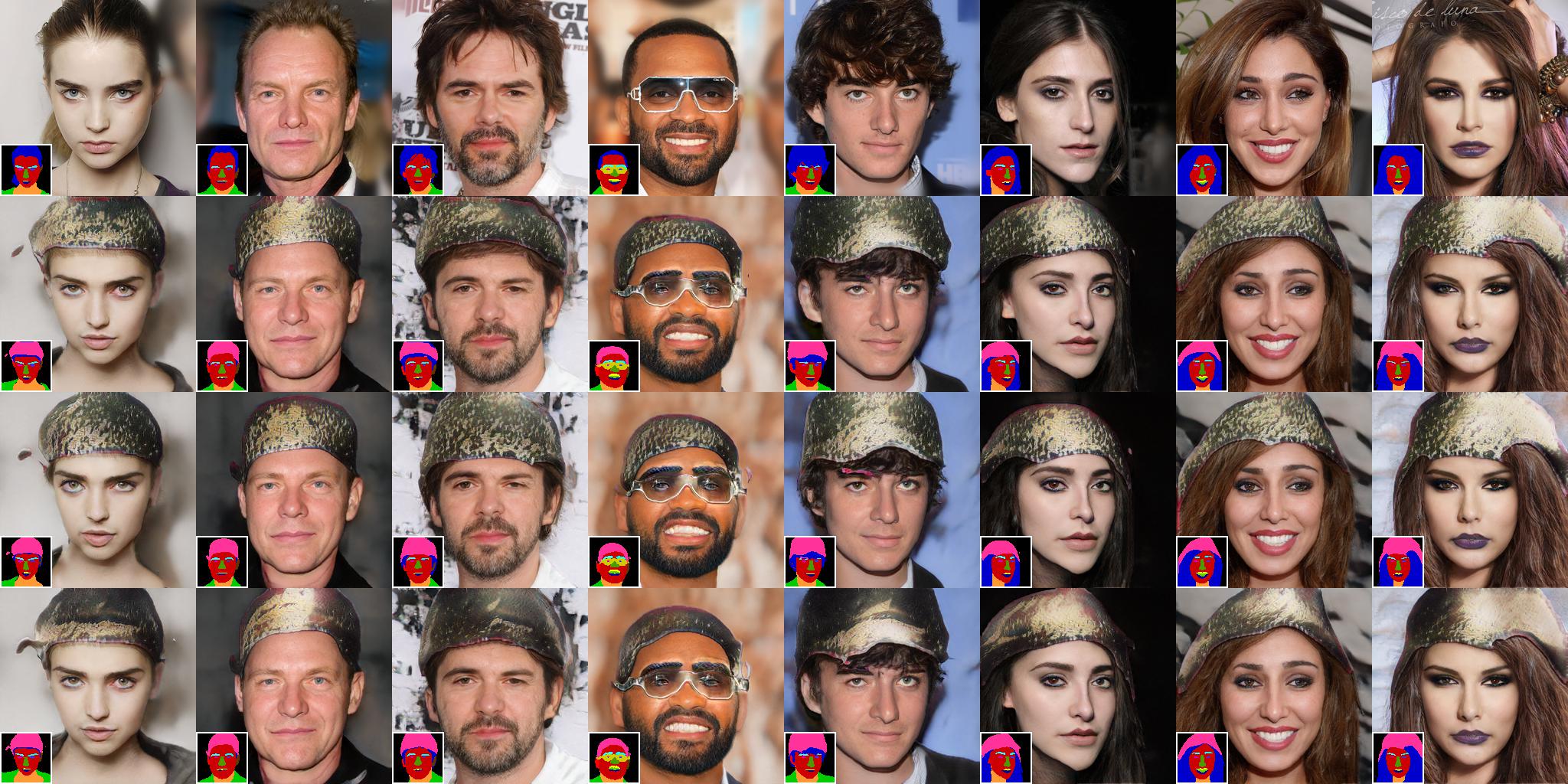}
            \subcaption{Latent Synthesis}        
        \end{minipage}    
        \hfill\vline\hfill
        \begin{minipage}[c]{0.515\textwidth}
            \includegraphics[width=\textwidth]{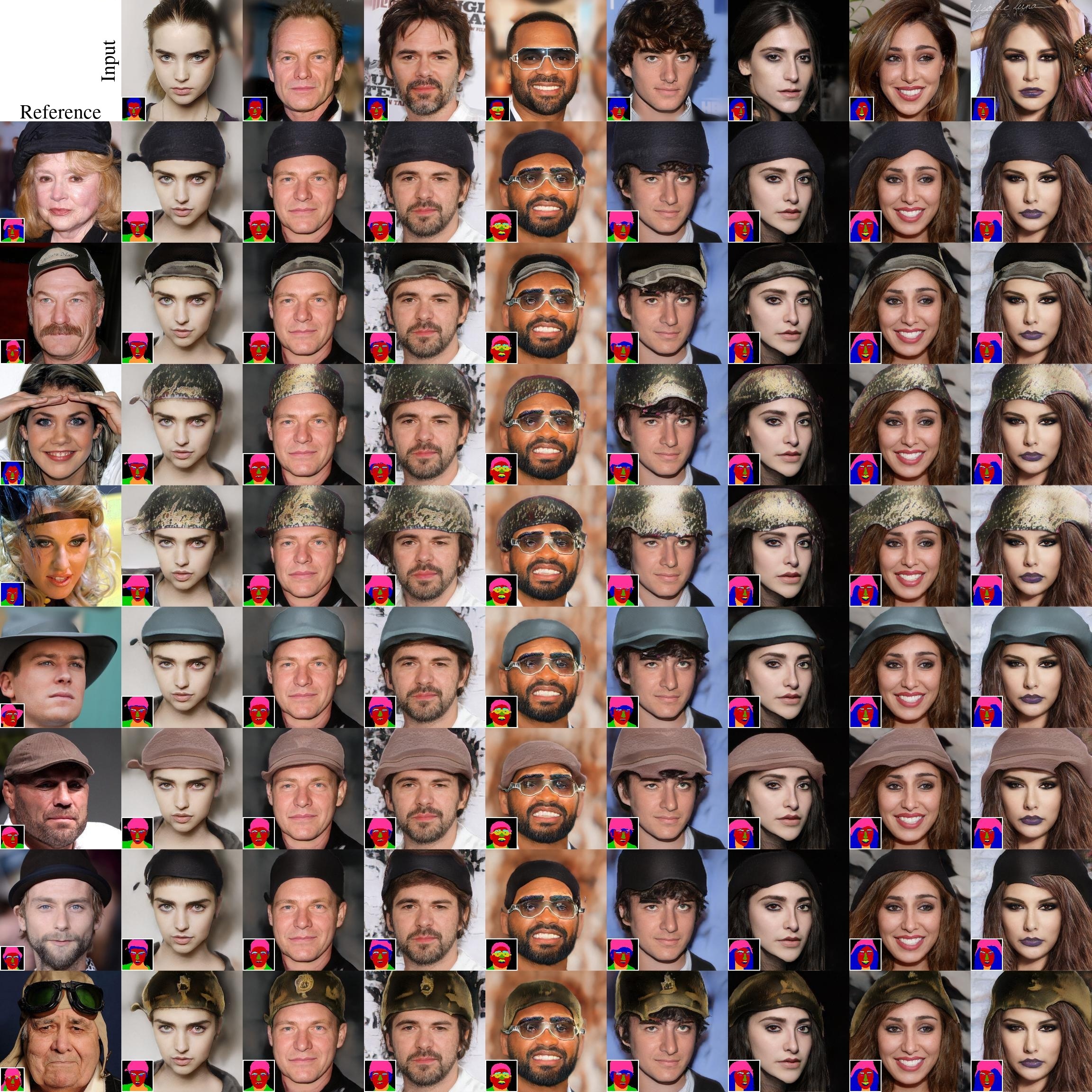}
            \subcaption{Reference Synthesis}
        \end{minipage}        
    \end{minipage}    
    \caption{\textbf{Qualitative Results for Hat manipulation \underline{only}}. (a) Reconstruction results. First and second row are input and reference images, respectively. Third and fourth rows are forward and reconstruction outputs, respectively. (b) Latent synthesis generation for both semantic and rgb outputs. First row represents input images. (c) reference image synthesis for both semantic and rgb space.}
    \label{figure:qualitative_hat}
\end{figure*}

\begin{figure*}[t]
    \centering%
    \begin{minipage}{.8\textwidth}
        \begin{minipage}[c]{0.475\textwidth}
            \includegraphics[width=\textwidth]{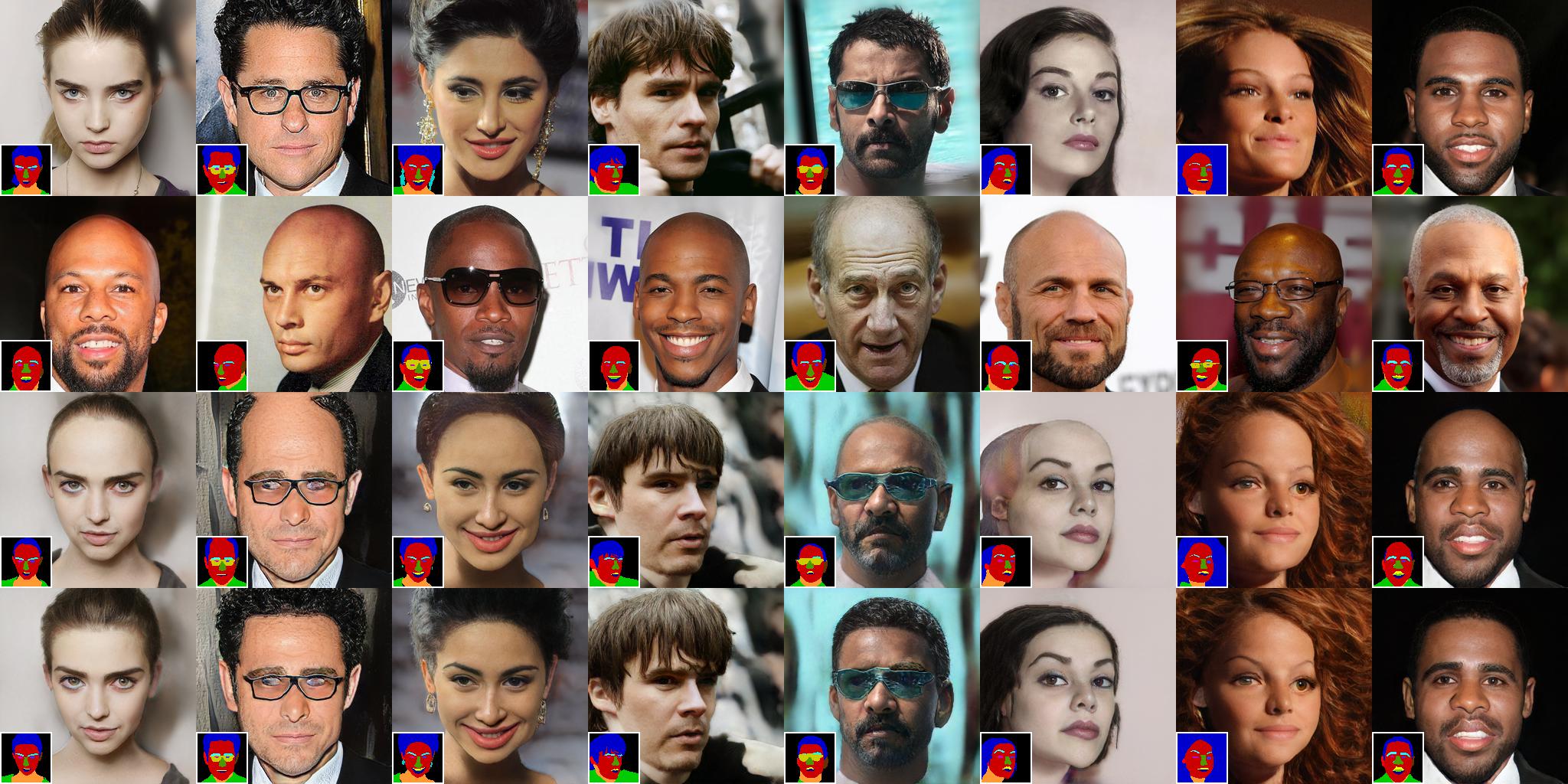} \subcaption{Reconstruction}
            
            \includegraphics[width=\textwidth]{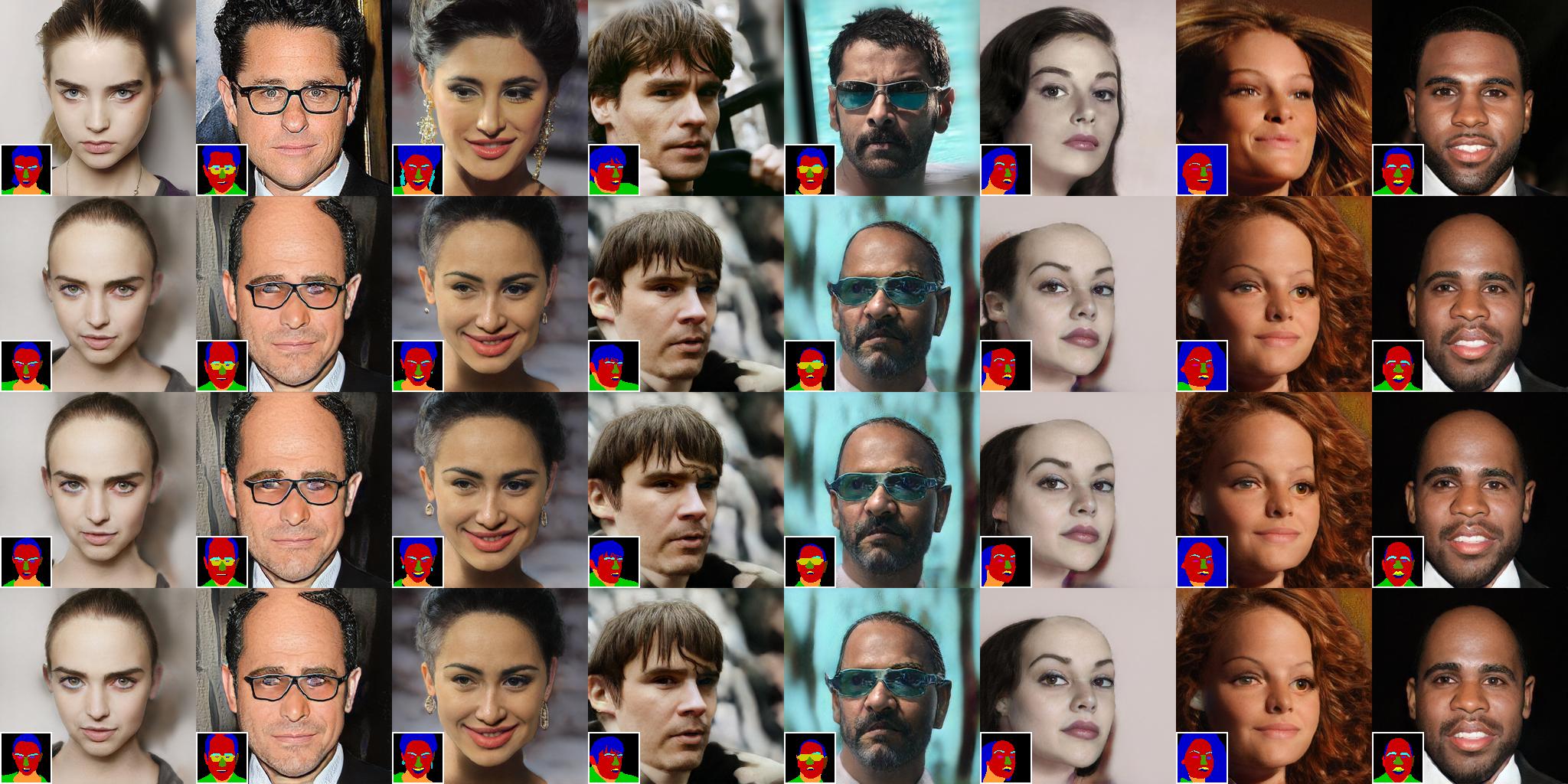}
            \subcaption{Latent Synthesis}        
        \end{minipage}    
        \hfill\vline\hfill
        \begin{minipage}[c]{0.515\textwidth}
            \includegraphics[width=\textwidth]{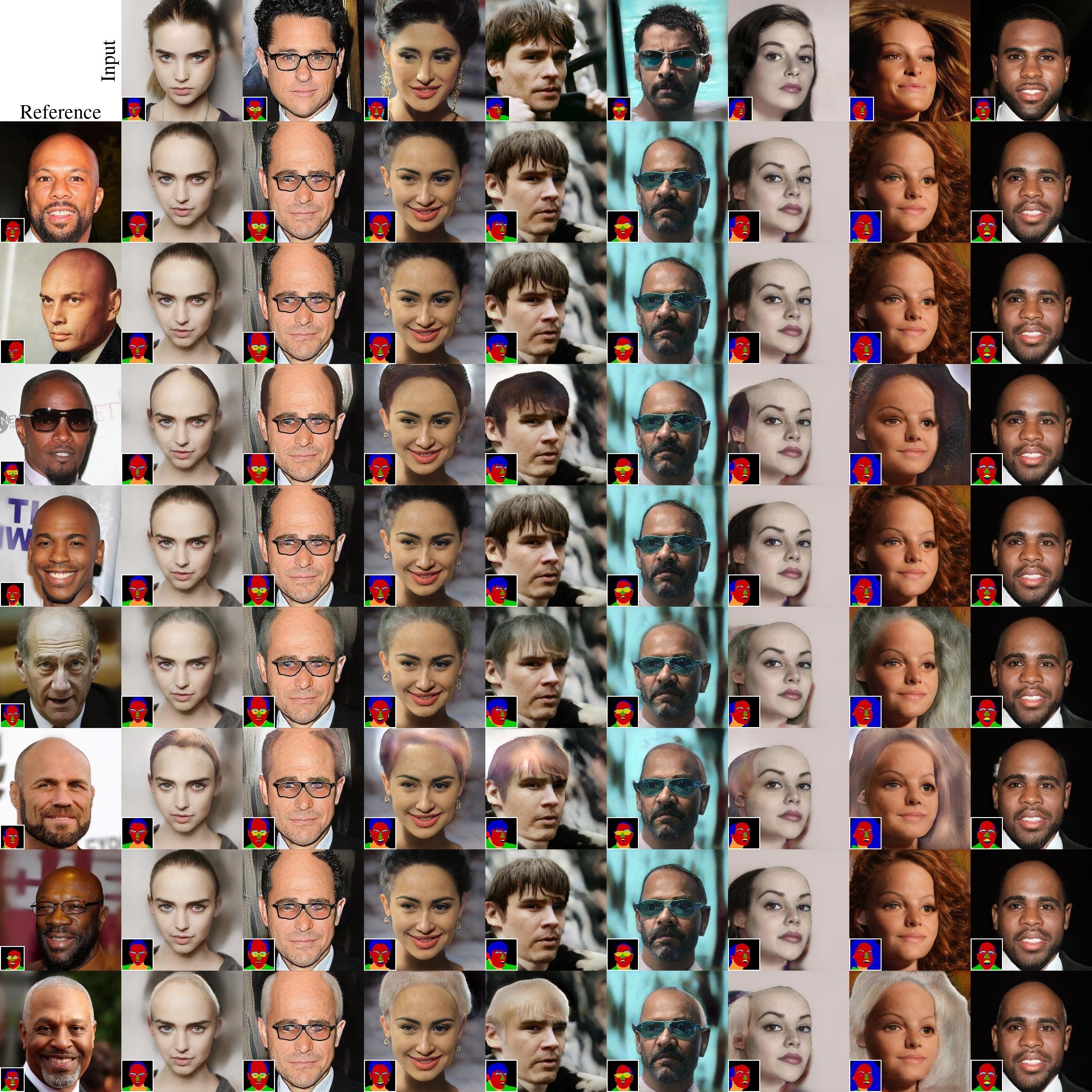}
            \subcaption{Reference Synthesis}
        \end{minipage}        
    \end{minipage}    
    \caption{\textbf{Qualitative Results for Hair manipulation \underline{only}}. (a) Reconstruction results. First and second row are input and reference images, respectively. Third and fourth rows are forward and reconstruction outputs, respectively. (b) Latent synthesis generation for both semantic and rgb outputs. First row represents input images. (c) reference image synthesis for both semantic and rgb space.}
    \label{figure:qualitative_hair}
\end{figure*}

\begin{figure*}[t]
    \centering%
    \begin{minipage}{.8\textwidth}
        \begin{minipage}[c]{0.475\textwidth}
            \includegraphics[width=\textwidth]{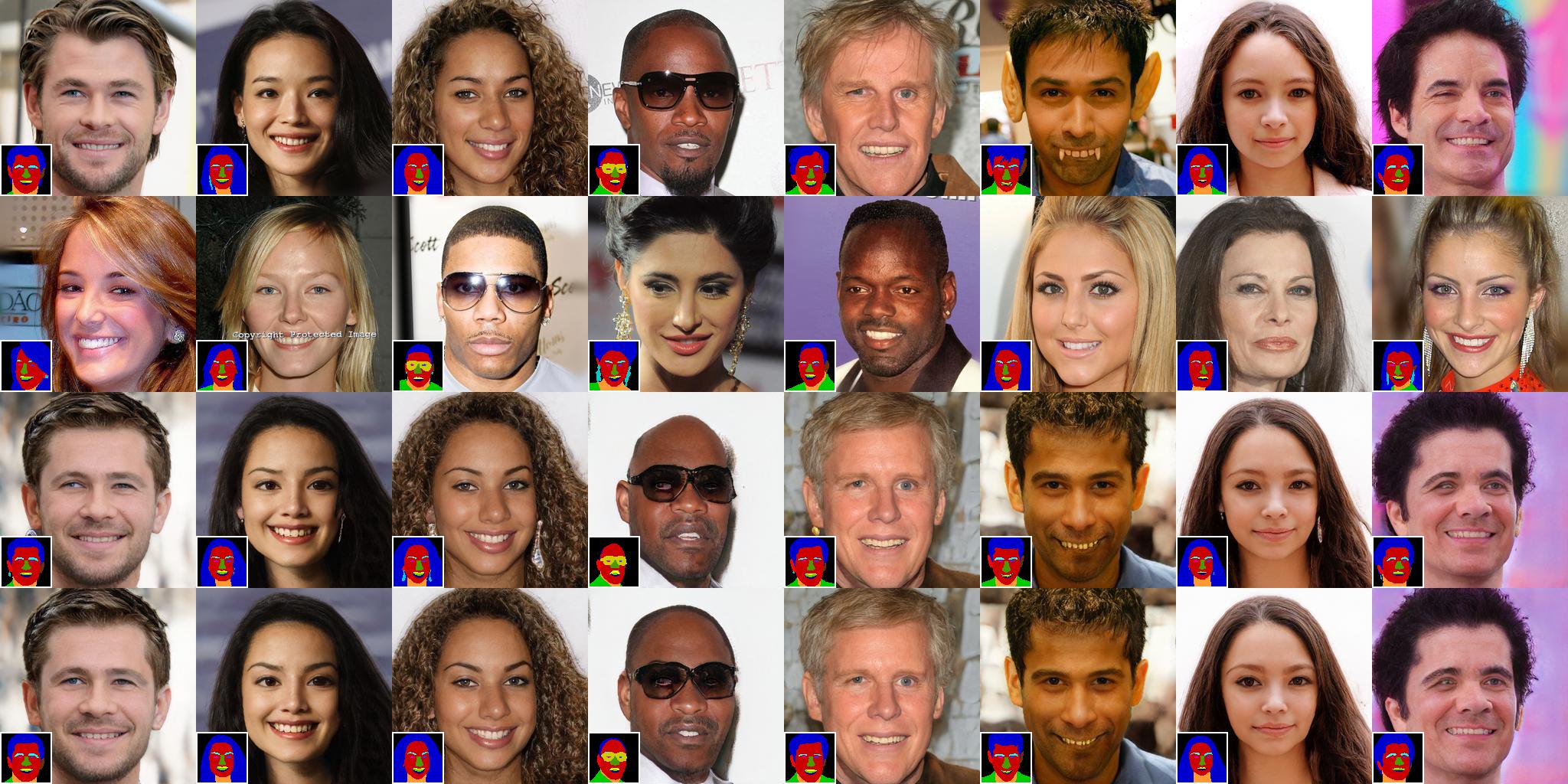} \subcaption{Reconstruction}
            
            \includegraphics[width=\textwidth]{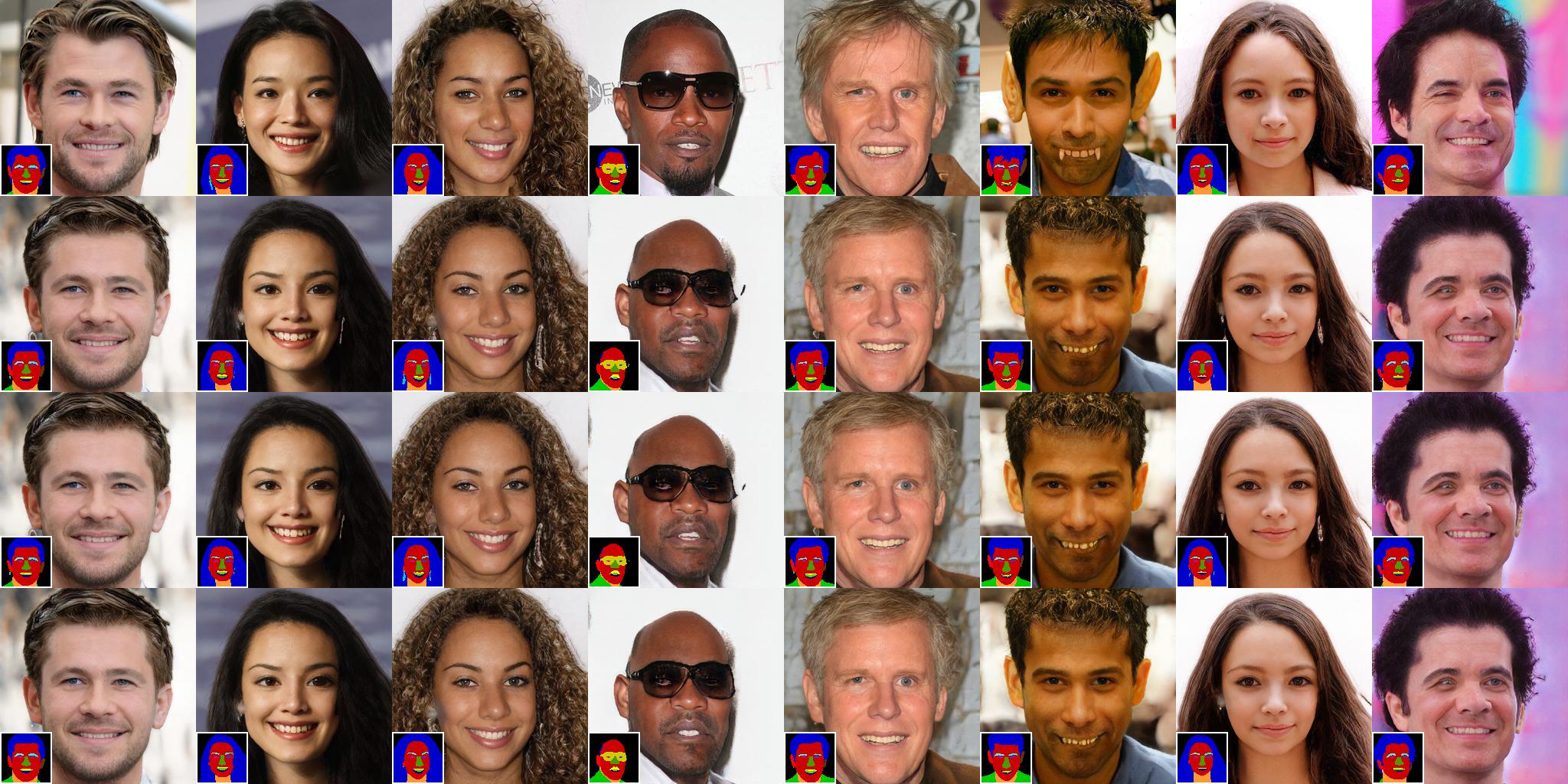}
            \subcaption{Latent Synthesis}        
        \end{minipage}    
        \hfill\vline\hfill
        \begin{minipage}[c]{0.515\textwidth}
            \includegraphics[width=\textwidth]{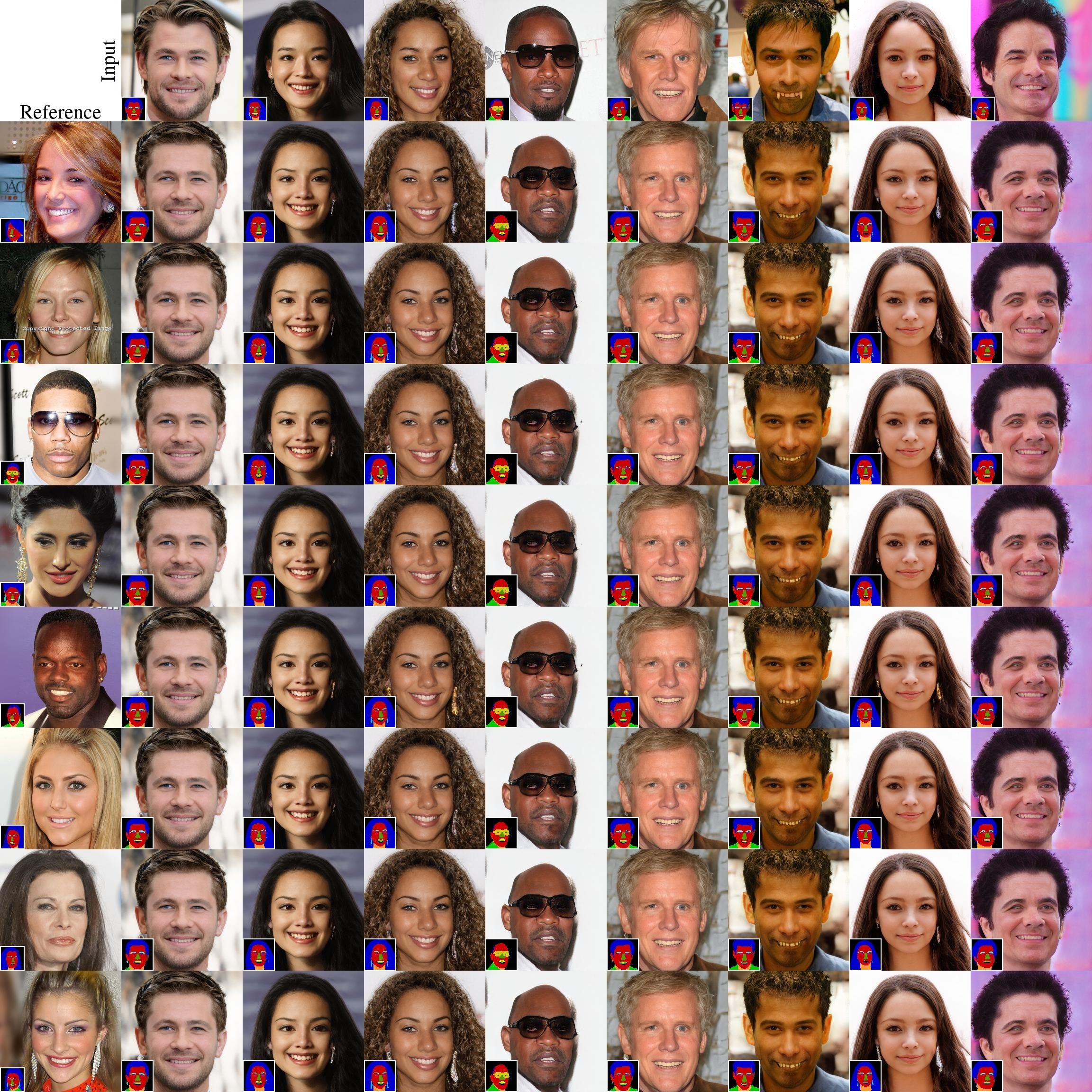}
            \subcaption{Reference Synthesis}
        \end{minipage}        
    \end{minipage}    
    \caption{\textbf{Qualitative Results for Earrings manipulation \underline{only}}. (a) Reconstruction results. First and second row are input and reference images, respectively. Third and fourth rows are forward and reconstruction outputs, respectively. (b) Latent synthesis generation for both semantic and rgb outputs. First row represents input images. (c) reference image synthesis for both semantic and rgb space.}
    \label{figure:qualitative_earrings}
\end{figure*}

~\fref{figure:qualitative_male},~\ref{figure:qualitative_female},~\ref{figure:qualitative_eyeglasses}~\ref{figure:qualitative_bangs}~\ref{figure:qualitative_hat}~\ref{figure:qualitative_hair} and~\ref{figure:qualitative_earrings} depict qualitative visualizations for each attribute independently.

\begin{table*}[t]
\resizebox{\linewidth}{!}{
\begin{tabular}{|l||c|c|c||c|c||c|||c|||c|}
\hline
\multirow{3}{*}{Attribute Manipulation} & \multicolumn{8}{c|}{CelebA-HQ~\cite{karras2017progressive} | Reference Synthesis} \\
\cline{2-9}
& \multicolumn{3}{c||}{Pose$\downarrow$} & \multicolumn{2}{c||}{Attributes$\uparrow$} & \multicolumn{1}{c|||}{Reconstruction$\uparrow$} & \multicolumn{2}{c|}{Perceptual} \\
\cline{2-9}
& Pitch & Roll & Yaw & AP & F1 & mIoU & FID$\downarrow$ & LPIPS$\uparrow$ \\
\hline
\textbf{Male}                & 14.812 $\pm$ 22.972 & 1.735 $\pm$ 6.619 & 12.462 $\pm$ 24.730 & 0.916 $\pm$ 0.127 & 0.914 $\pm$ 0.143 & 0.990 $\pm$ 0.005 & 33.197 $\pm$ 0.000  & 0.275 $\pm$ 0.036 \\
\textbf{Female}              & 12.600 $\pm$ 32.221 & 2.250 $\pm$ 8.427 & 12.983 $\pm$ 24.233 & 0.942 $\pm$ 0.103 & 0.939 $\pm$ 0.095 & 0.990 $\pm$ 0.004 & 16.408 $\pm$ 0.000  & 0.281 $\pm$ 0.030 \\
\textbf{Removing Eyeglasses} & 15.426 $\pm$ 30.238 & 2.036 $\pm$ 3.528 & 9.981 $\pm$ 23.557  & 0.908 $\pm$ 0.173 & 0.892 $\pm$ 0.179 & 0.987 $\pm$ 0.004 & 69.677 $\pm$ 0.000  & 0.059 $\pm$ 0.021 \\
\textbf{Adding Eyeglasses}   & 17.774 $\pm$ 25.477 & 2.388 $\pm$ 7.971 & 9.269 $\pm$ 21.661  & 0.951 $\pm$ 0.074 & 0.922 $\pm$ 0.095 & 0.990 $\pm$ 0.004 & 37.946 $\pm$ 0.000  & 0.102 $\pm$ 0.025 \\
\textbf{Removing Hair}       & 17.561 $\pm$ 36.659 & 2.167 $\pm$ 6.065 & 9.580 $\pm$ 22.614  & 0.927 $\pm$ 0.091 & 0.905 $\pm$ 0.090 & 0.989 $\pm$ 0.004 & 68.845 $\pm$ 0.000  & 0.095 $\pm$ 0.052 \\
\textbf{Adding Hair}         & 12.212 $\pm$ 28.076 & 1.704 $\pm$ 2.898 & 9.933 $\pm$ 15.999  & 0.994 $\pm$ 0.009 & 0.990 $\pm$ 0.014 & 0.990 $\pm$ 0.003 & 110.764 $\pm$ 0.000 & 0.071 $\pm$ 0.018 \\
\textbf{Removing Bangs}      & 8.608 $\pm$ 14.954  & 2.605 $\pm$ 9.193 & 10.471 $\pm$ 27.656 & 0.894 $\pm$ 0.173 & 0.892 $\pm$ 0.137 & 0.988 $\pm$ 0.005 & 33.784 $\pm$ 0.000  & 0.164 $\pm$ 0.047 \\
\textbf{Adding Bangs}        & 10.000 $\pm$ 23.522 & 1.680 $\pm$ 5.321 & 7.913 $\pm$ 17.688  & 0.958 $\pm$ 0.065 & 0.941 $\pm$ 0.072 & 0.990 $\pm$ 0.004 & 25.799 $\pm$ 0.000  & 0.158 $\pm$ 0.045 \\
\textbf{Removing Earrings}   & 6.767 $\pm$ 13.369  & 0.959 $\pm$ 1.799 & 6.655 $\pm$ 15.190  & 0.955 $\pm$ 0.073 & 0.940 $\pm$ 0.076 & 0.989 $\pm$ 0.004 & 52.971 $\pm$ 0.000  & 0.025 $\pm$ 0.011 \\
\textbf{Adding Earrings}     & 12.059 $\pm$ 27.609 & 1.674 $\pm$ 5.348 & 7.656 $\pm$ 19.257  & 0.983 $\pm$ 0.028 & 0.965 $\pm$ 0.036 & 0.990 $\pm$ 0.004 & 35.760 $\pm$ 0.000  & 0.036 $\pm$ 0.018 \\
\textbf{Removing Hat}        & 17.459 $\pm$ 30.457 & 2.531 $\pm$ 3.671 & 8.516 $\pm$ 15.985  & 0.962 $\pm$ 0.086 & 0.952 $\pm$ 0.107 & 0.983 $\pm$ 0.008 & 67.065 $\pm$ 0.000  & 0.087 $\pm$ 0.030 \\
\textbf{Adding Hat}          & 13.427 $\pm$ 27.384 & 1.646 $\pm$ 6.487 & 7.848 $\pm$ 17.526  & 0.910 $\pm$ 0.147 & 0.881 $\pm$ 0.166 & 0.990 $\pm$ 0.004 & 50.865 $\pm$ 0.000  & 0.194 $\pm$ 0.042 \\
\hline
\end{tabular}
}
\caption{\textbf{Additional quantitative results for semantic manipulation using exemplar images.}}
\label{table:additional_sem_ref}
\end{table*}

\begin{table*}[t]
\resizebox{\linewidth}{!}{
\begin{tabular}{|l||c|c|c||c|c||c|||c|||c|}
\hline
\multirow{3}{*}{Attribute Manipulation} & \multicolumn{8}{c|}{CelebA-HQ~\cite{karras2017progressive} | Latent Synthesis} \\
\cline{2-9}
& \multicolumn{3}{c||}{Pose$\downarrow$} & \multicolumn{2}{c||}{Attributes$\uparrow$} & \multicolumn{1}{c|||}{Reconstruction$\uparrow$} & \multicolumn{2}{c|}{Perceptual} \\
\cline{2-9}
& Pitch & Roll & Yaw & AP & F1 & mIoU & FID$\downarrow$ & LPIPS$\uparrow$ \\
\hline
\textbf{Male}                & 13.590 $\pm$ 22.054  & 2.072 $\pm$ 9.391   & 13.673 $\pm$ 30.232 & 0.958 $\pm$ 0.094 & 0.942 $\pm$ 0.130 & 0.990 $\pm$ 0.005 & 40.799 $\pm$ 0.000 & 0.418 $\pm$ 0.026 \\
\textbf{Female}              & 11.955 $\pm$ 29.208  & 2.350 $\pm$ 9.625   & 10.281 $\pm$ 31.014 & 0.946 $\pm$ 0.116 & 0.947 $\pm$ 0.105 & 0.990 $\pm$ 0.004 & 22.844 $\pm$ 0.000 & 0.424 $\pm$ 0.030 \\
\textbf{Removing Eyeglasses} & 22.192 $\pm$ 45.327  & 3.591 $\pm$ 8.847   & 11.065 $\pm$ 26.139 & 0.913 $\pm$ 0.167 & 0.924 $\pm$ 0.140 & 0.987 $\pm$ 0.004 & 45.112 $\pm$ 0.000 & 0.393 $\pm$ 0.031 \\
\textbf{Adding Eyeglasses}   & 21.041 $\pm$ 29.497  & 3.287 $\pm$ 9.143   & 11.461 $\pm$ 26.072 & 0.969 $\pm$ 0.055 & 0.939 $\pm$ 0.078 & 0.990 $\pm$ 0.004 & 42.884 $\pm$ 0.000 & 0.410 $\pm$ 0.030 \\
\textbf{Removing Hair}       & 19.733 $\pm$ 33.123  & 3.052 $\pm$ 8.357   & 11.608 $\pm$ 28.834 & 0.959 $\pm$ 0.043 & 0.929 $\pm$ 0.057 & 0.989 $\pm$ 0.004 & 74.794 $\pm$ 0.000 & 0.388 $\pm$ 0.032 \\
\textbf{Adding Hair}         & 17.214 $\pm$ 29.187  & 2.467 $\pm$ 5.099   & 13.014 $\pm$ 46.367 & 1.000 $\pm$ 0.000 & 1.000 $\pm$ 0.000 & 0.989 $\pm$ 0.003 & 72.738 $\pm$ 0.000 & 0.392 $\pm$ 0.026 \\
\textbf{Removing Bangs}      & 11.708 $\pm$ 27.843  & 2.995 $\pm$ 11.400  & 12.537 $\pm$ 33.663 & 0.985 $\pm$ 0.029 & 0.965 $\pm$ 0.054 & 0.988 $\pm$ 0.005 & 29.660 $\pm$ 0.000 & 0.381 $\pm$ 0.033 \\
\textbf{Adding Bangs}        & 10.996 $\pm$ 21.612  & 2.121 $\pm$ 5.462   & 9.060 $\pm$ 22.742  & 0.981 $\pm$ 0.030 & 0.960 $\pm$ 0.046 & 0.990 $\pm$ 0.004 & 27.725 $\pm$ 0.000 & 0.398 $\pm$ 0.031 \\
\textbf{Removing Earrings}   & 8.376 $\pm$ 14.259   & 1.336 $\pm$ 2.549   & 8.695 $\pm$ 28.302  & 0.989 $\pm$ 0.026 & 0.976 $\pm$ 0.050 & 0.989 $\pm$ 0.004 & 39.078 $\pm$ 0.000 & 0.371 $\pm$ 0.033 \\
\textbf{Adding Earrings}     & 13.158 $\pm$ 25.004  & 2.221 $\pm$ 6.837   & 9.153 $\pm$ 23.328  & 0.997 $\pm$ 0.003 & 0.987 $\pm$ 0.010 & 0.990 $\pm$ 0.004 & 32.287 $\pm$ 0.000 & 0.376 $\pm$ 0.032 \\
\textbf{Removing Hat}        & 15.321 $\pm$ 22.200  & 3.623 $\pm$ 5.605   & 15.344 $\pm$ 29.127 & 0.904 $\pm$ 0.188 & 0.891 $\pm$ 0.182 & 0.983 $\pm$ 0.008 & 49.037 $\pm$ 0.000 & 0.394 $\pm$ 0.028 \\
\textbf{Adding Hat}          & 15.697 $\pm$ 25.855  & 1.949 $\pm$ 5.856   & 9.540 $\pm$ 21.746  & 0.926 $\pm$ 0.125 & 0.896 $\pm$ 0.146 & 0.990 $\pm$ 0.004 & 40.860 $\pm$ 0.000 & 0.442 $\pm$ 0.031 \\
\hline
\end{tabular}
}
\caption{\textbf{Additional quantitative results for semantic manipulation using latent synthesis.}}
\label{table:additional_sem_lat}
\end{table*}

\subsection{Semantic Manipulation Disentanglement}
\label{appendix:additional_sem_pr}
\begin{figure*}[t]
\begin{center}
\centering
\resizebox{\linewidth}{!}{
\begin{tabular}{|c|c|||c|c|}
  \hline
  \includegraphics[width=\textwidth]{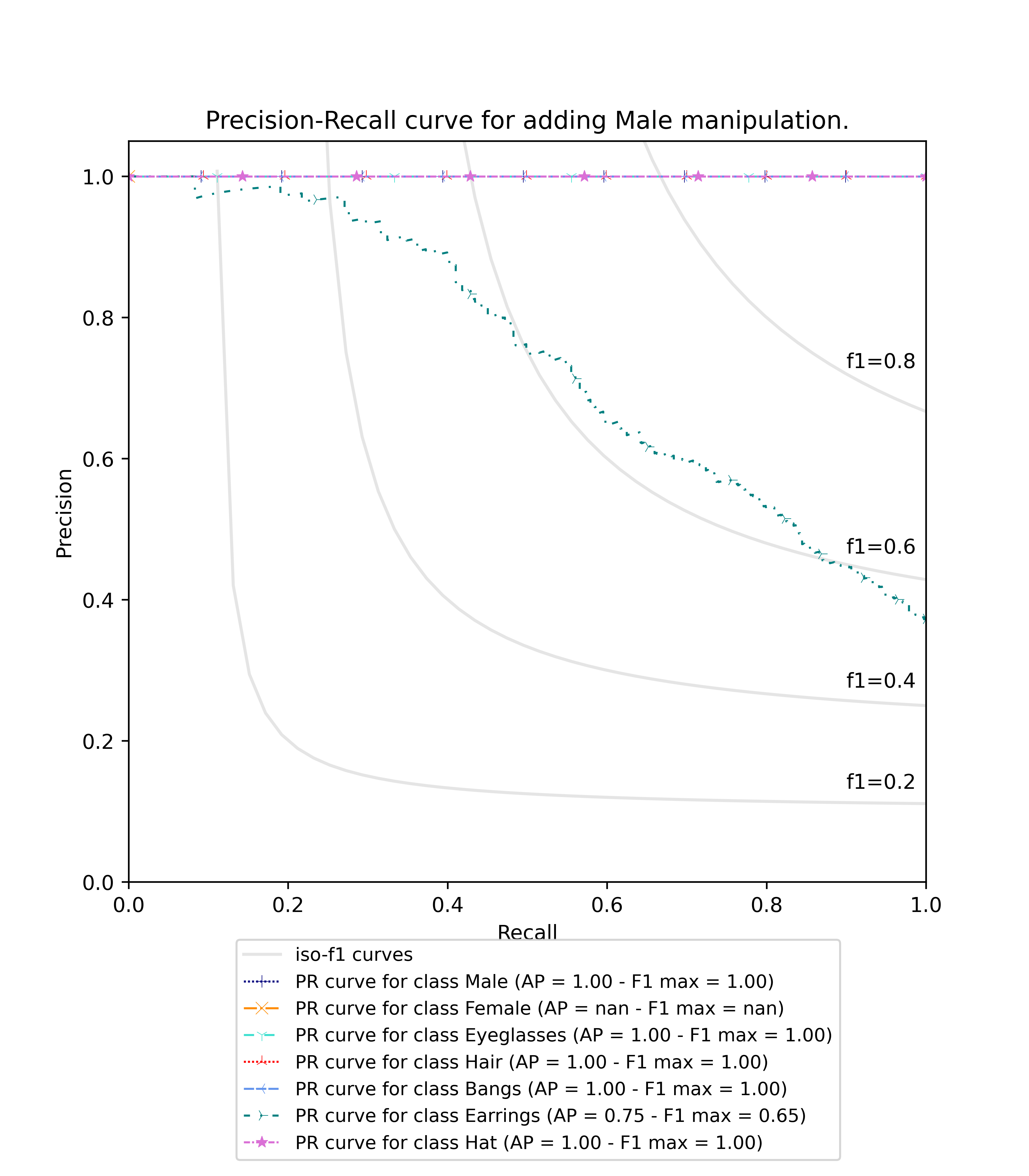} & \includegraphics[width=\textwidth]{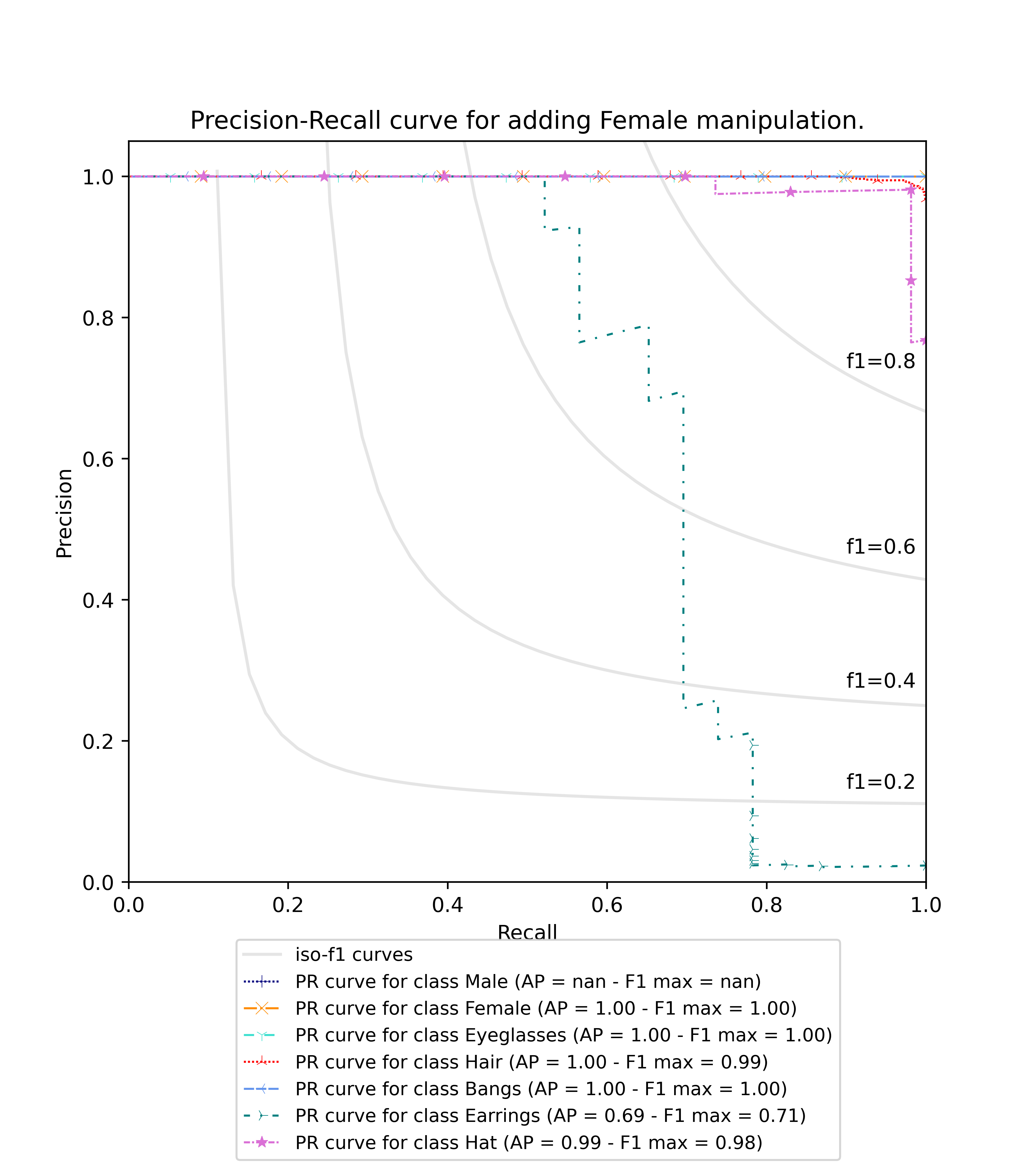} & \includegraphics[width=\textwidth]{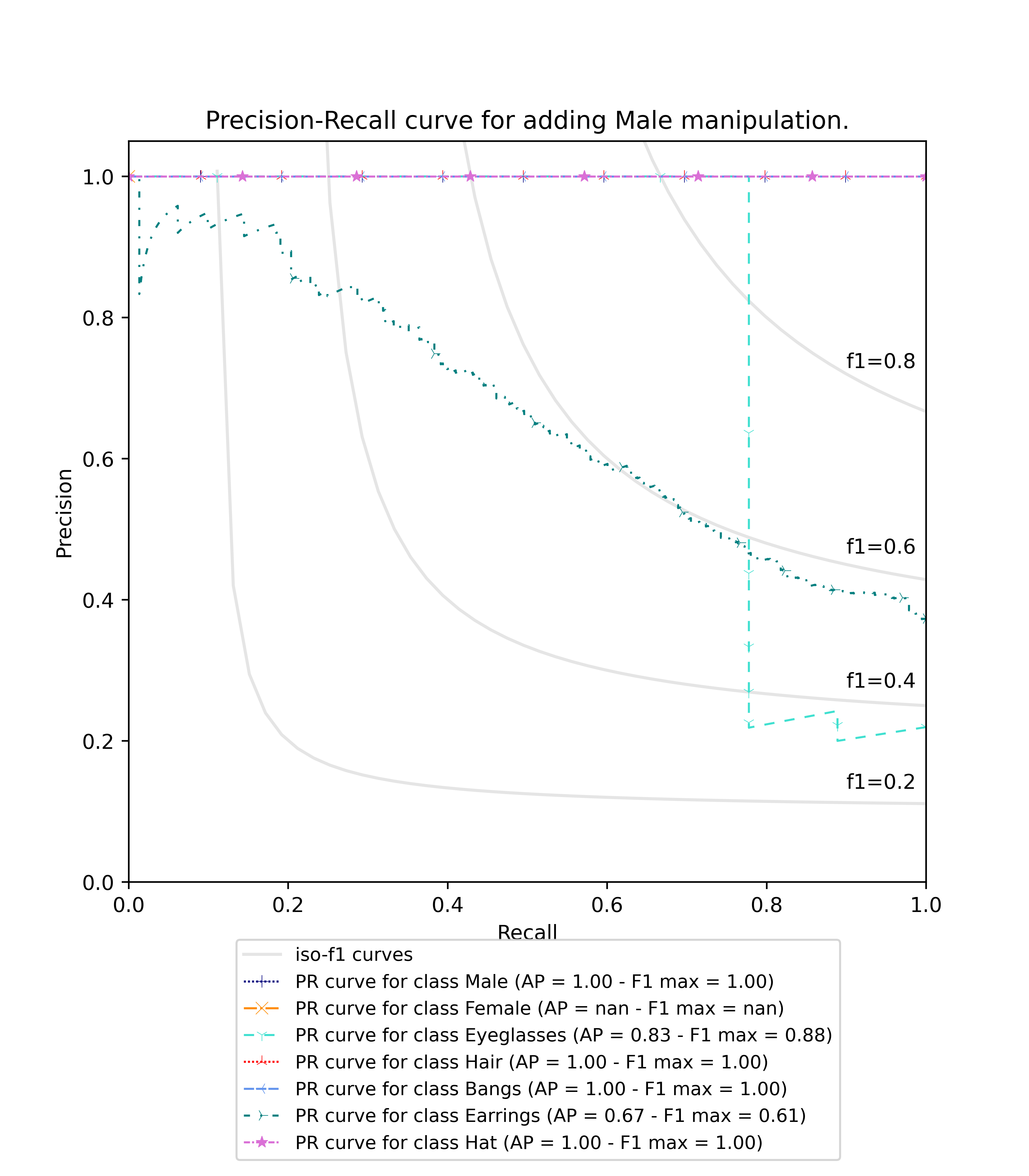} & \includegraphics[width=\textwidth]{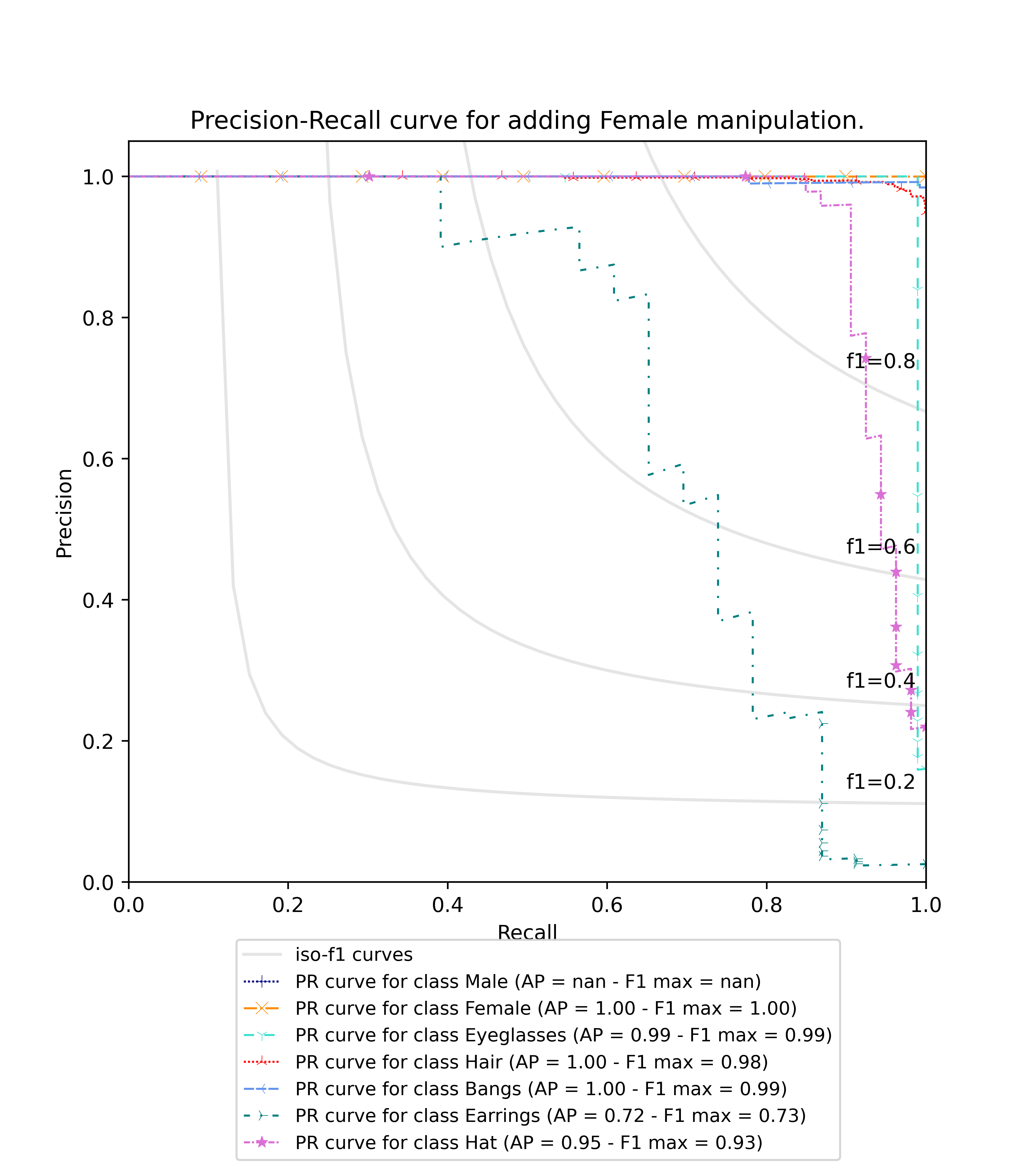} \\
  \includegraphics[width=\textwidth]{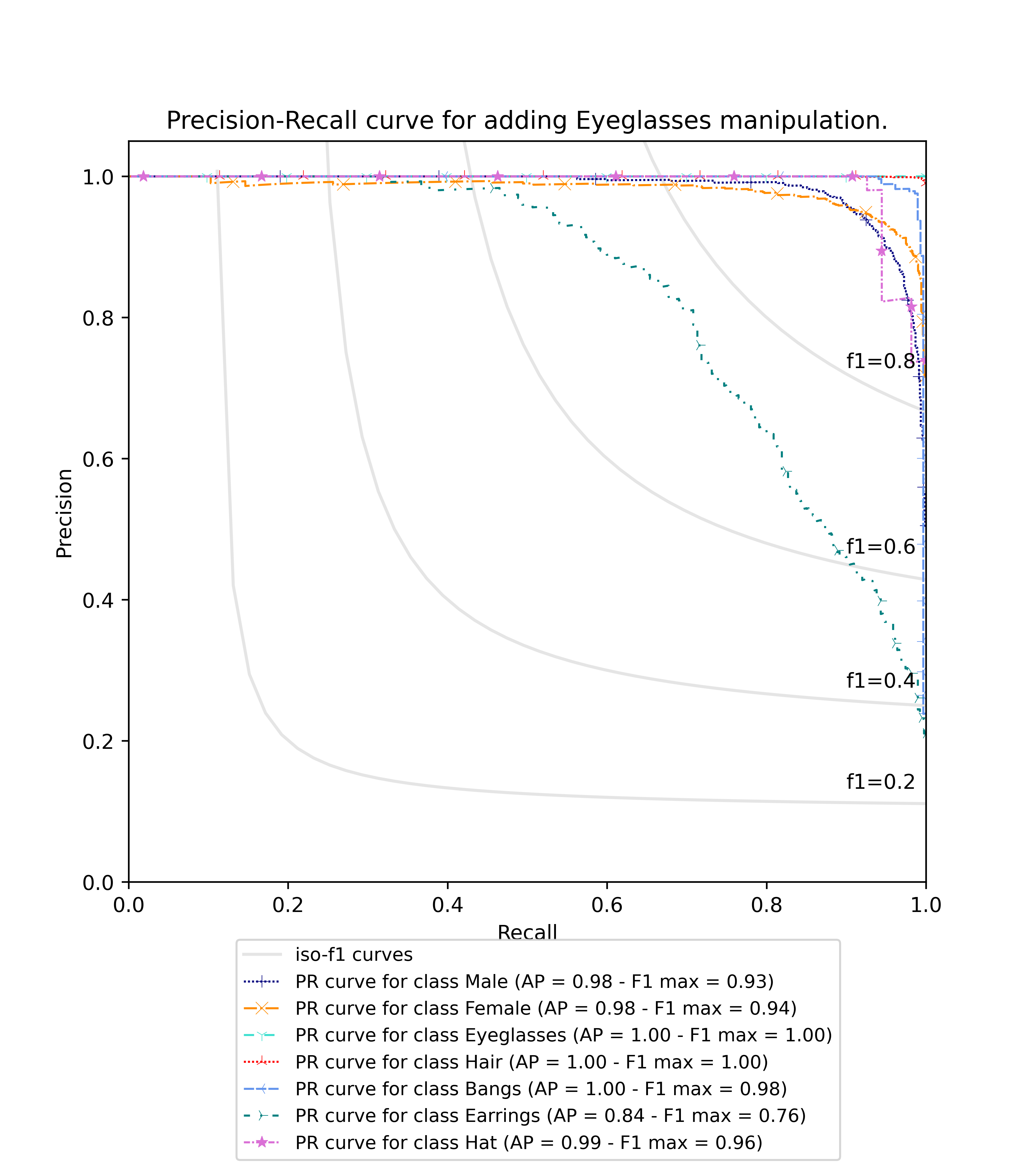} & \includegraphics[width=\textwidth]{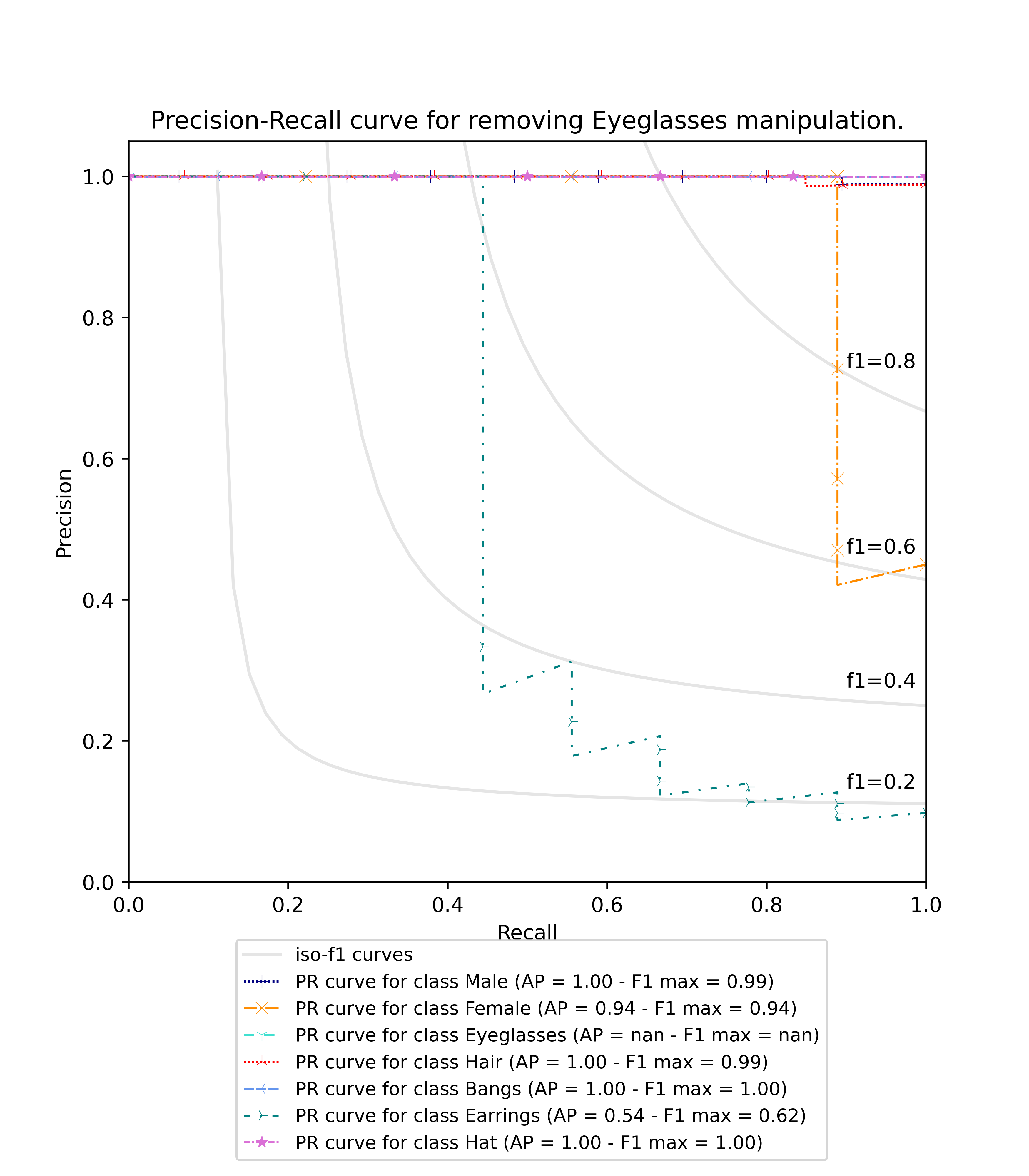} & \includegraphics[width=\textwidth]{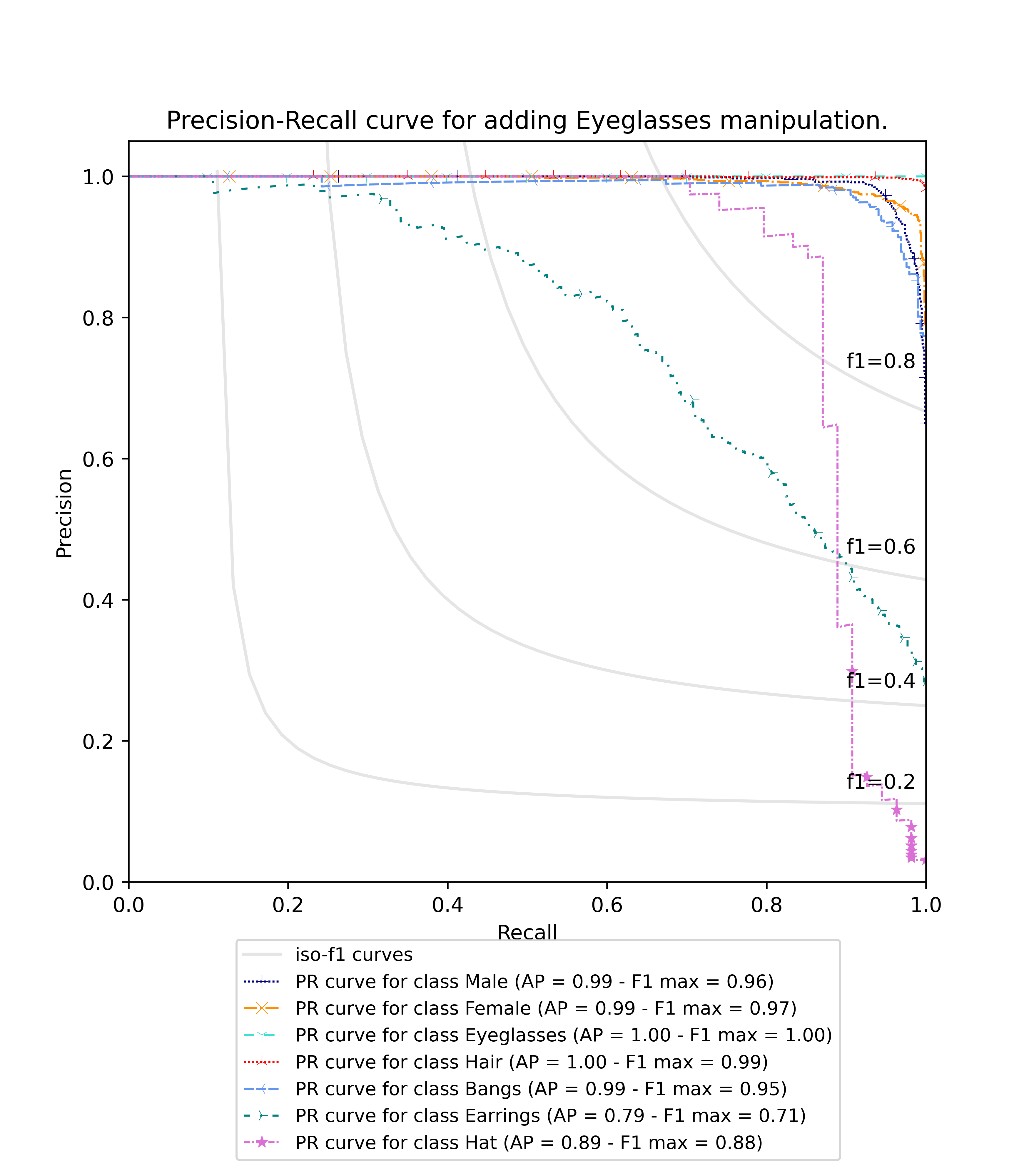} & \includegraphics[width=\textwidth]{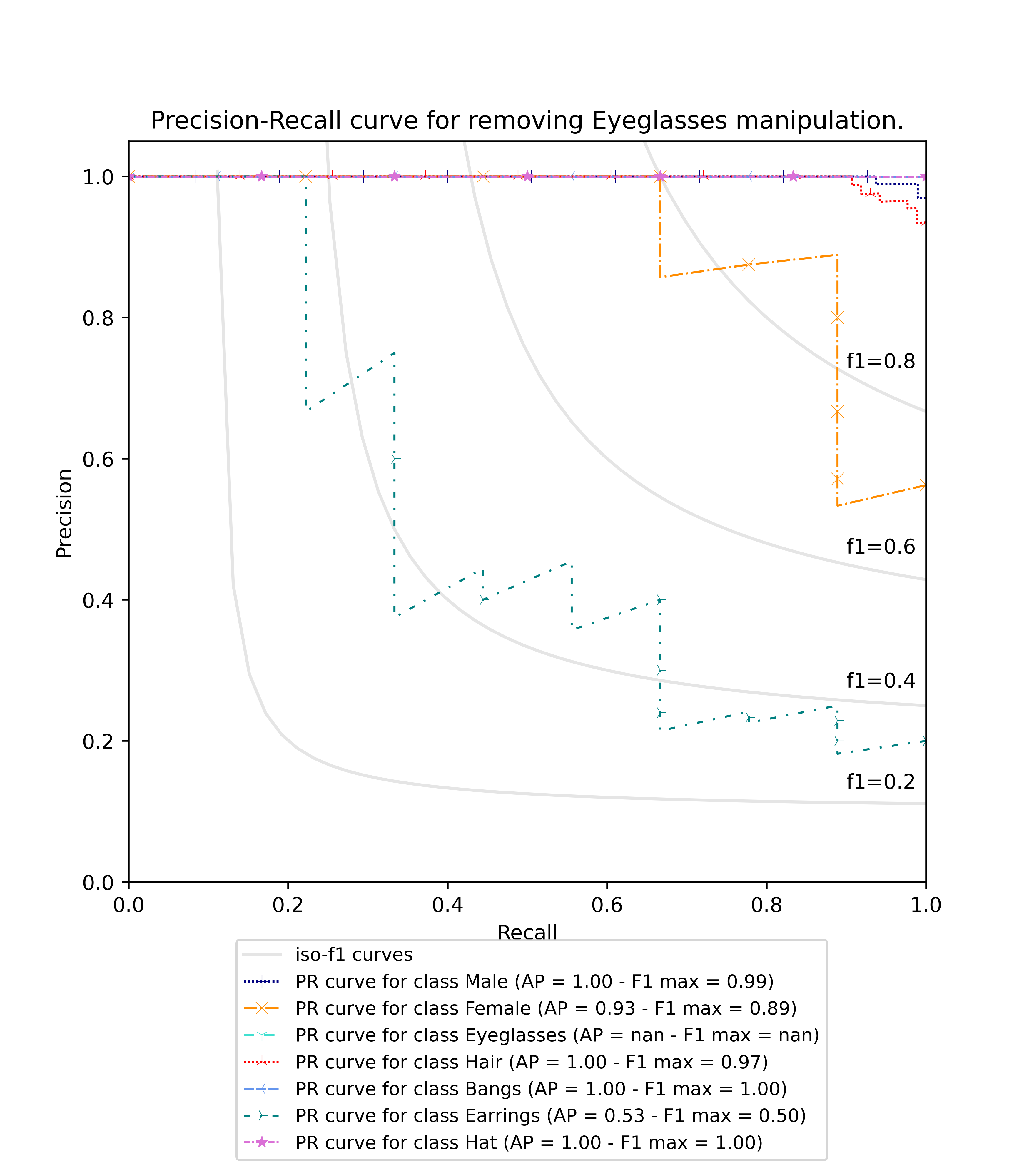} \\
  \includegraphics[width=\textwidth]{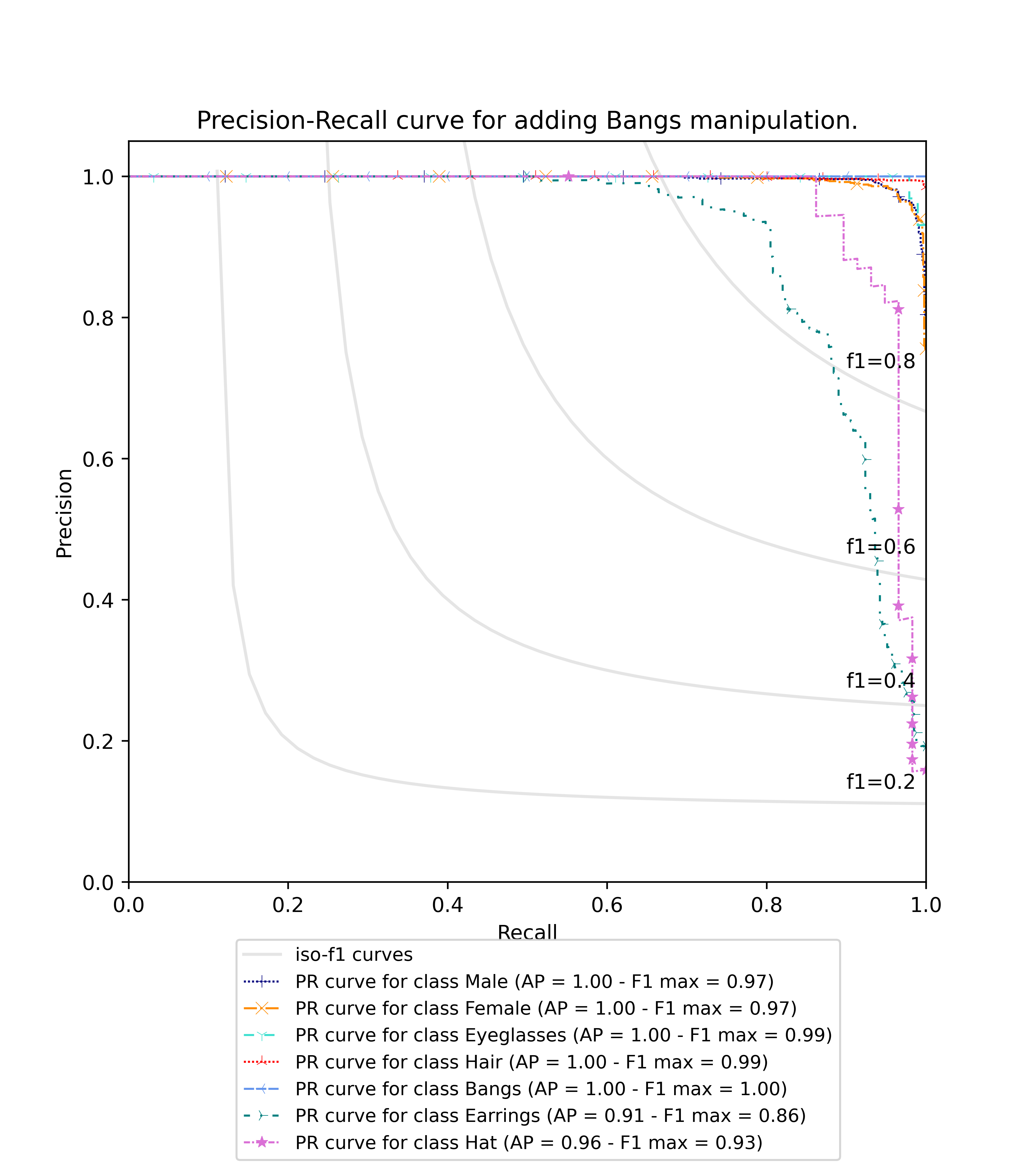} & \includegraphics[width=\textwidth]{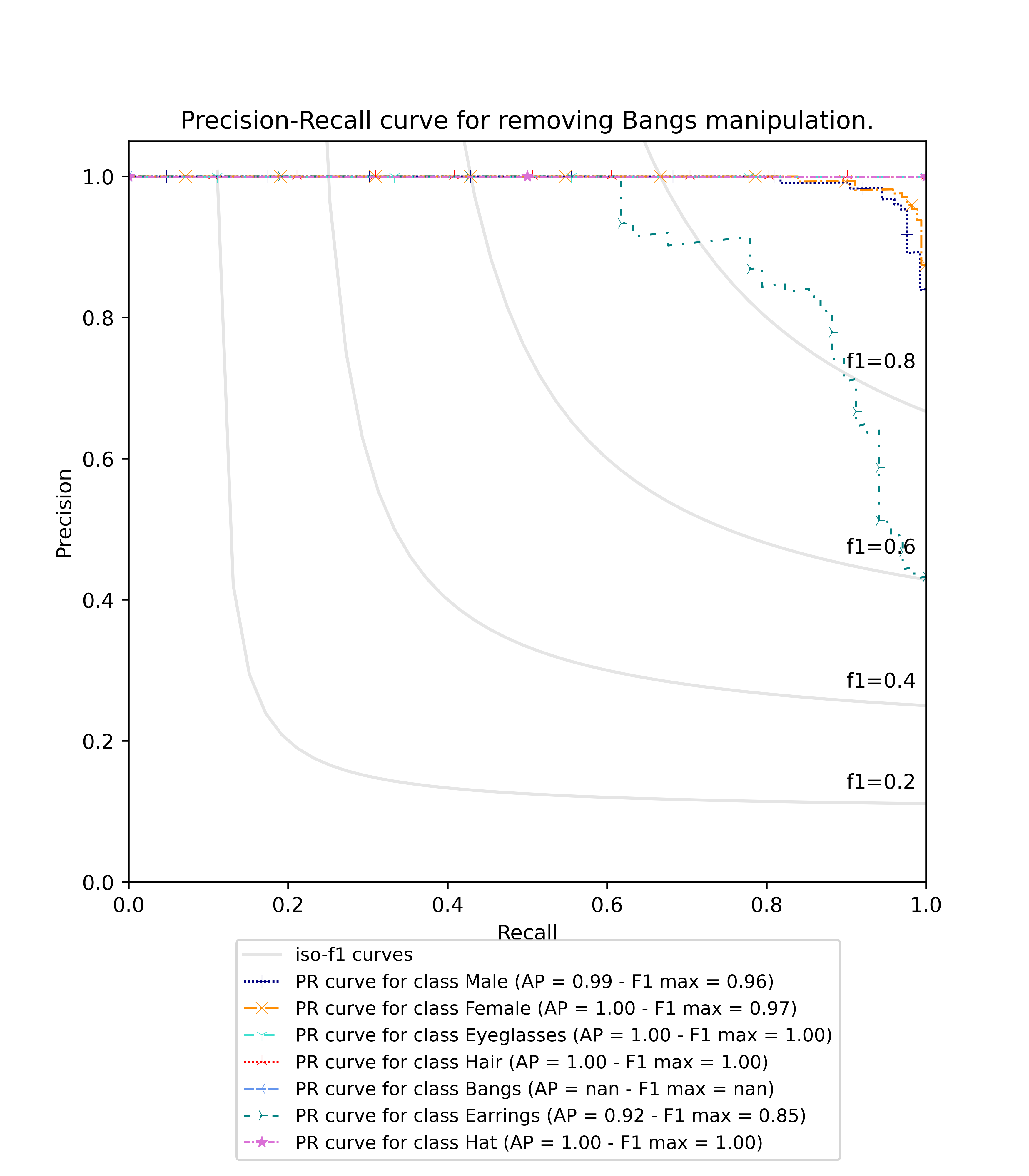} & \includegraphics[width=\textwidth]{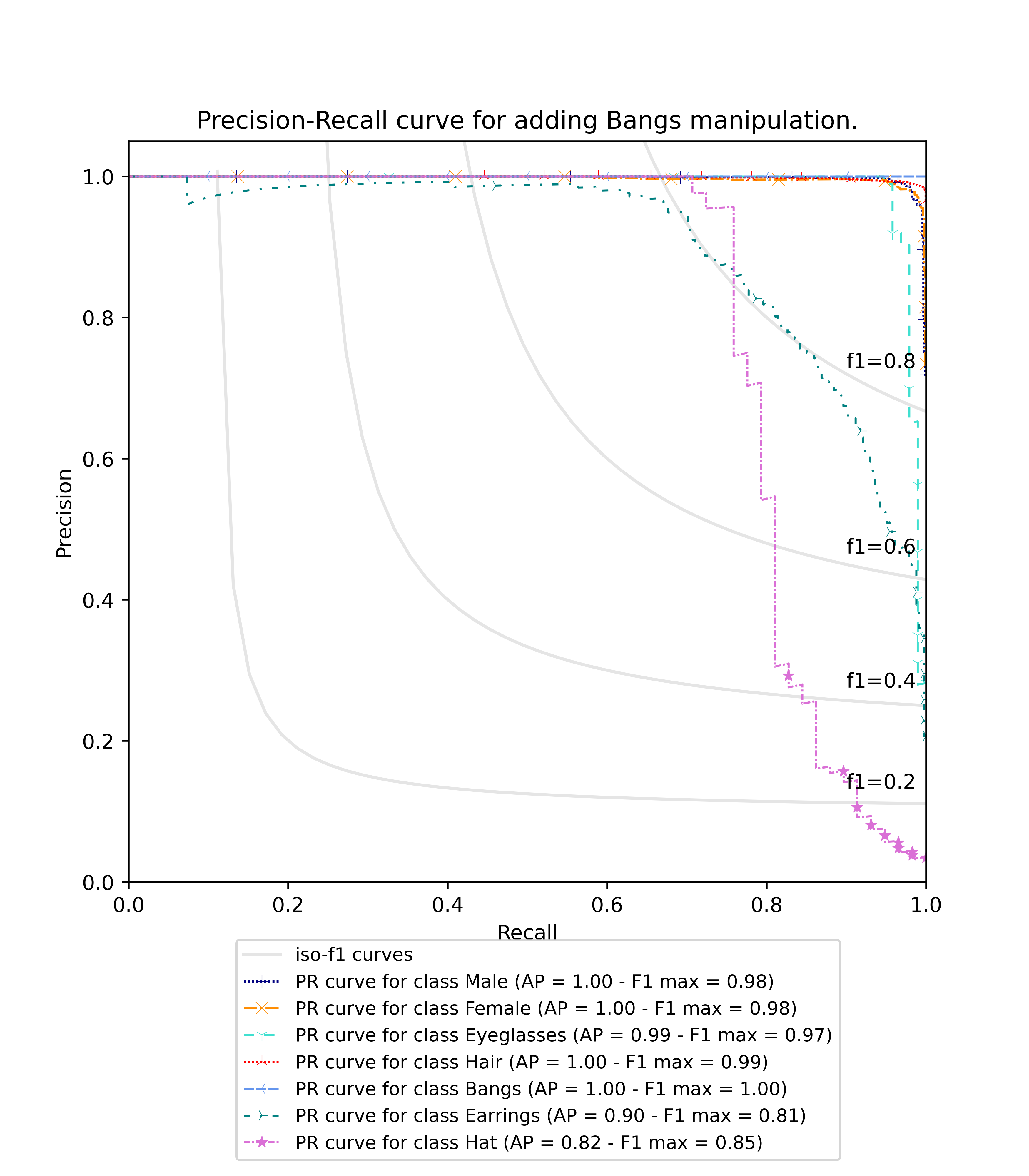} & \includegraphics[width=\textwidth]{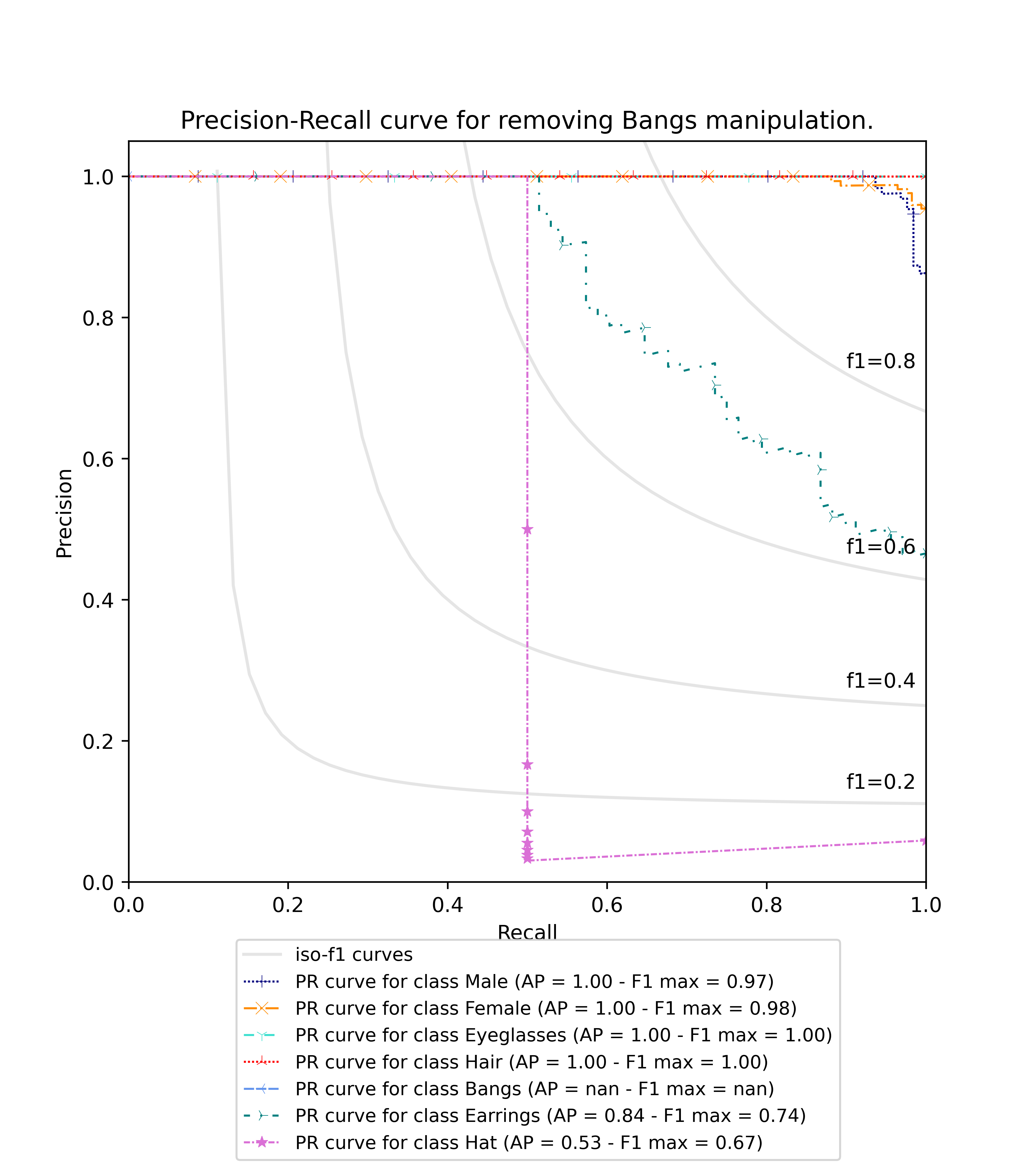} \\
  \hline
\end{tabular}
}
\end{center}
   \caption{\textbf{Precision vs Recall curves for each semantic manipulation for latent styles (left) and reference styles (right).} We study how the prediction of the remaining attributes is affected independently. Please zoom in for better details. Note that NaN in the legend of the images means the there were no images in the test set for the particular manipulation, \eg Male manipulation starts from the Female Test set and transform them to Male, so no Male in the entire set, and hence NaN in the score.}
\label{figure:sem_pr1}
\end{figure*}

\begin{figure*}[t]
\begin{center}
\centering
\resizebox{\linewidth}{!}{
\begin{tabular}{|c|c|||c|c|}
  \hline
  \includegraphics[width=\textwidth]{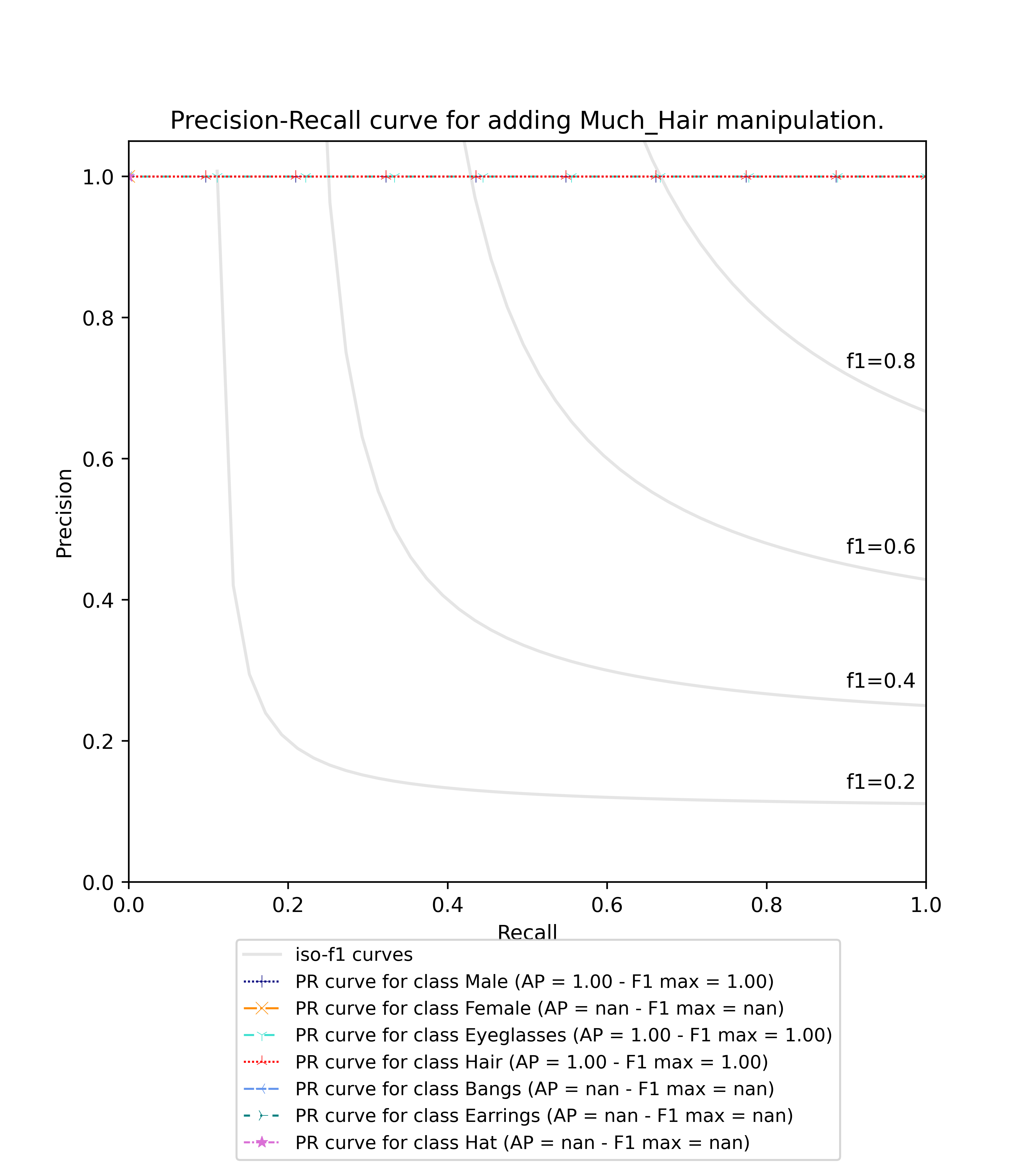} & \includegraphics[width=\textwidth]{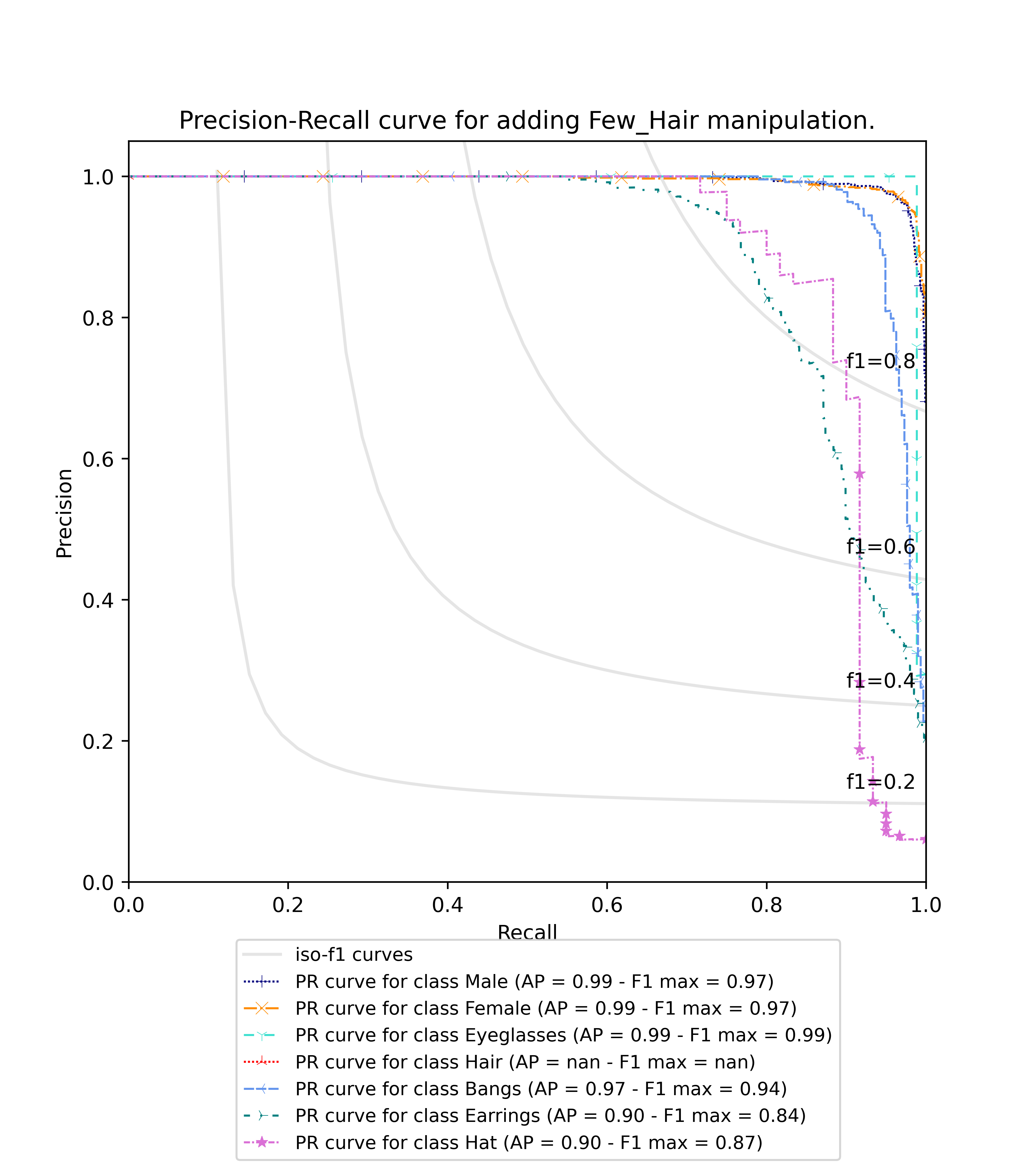} & \includegraphics[width=\textwidth]{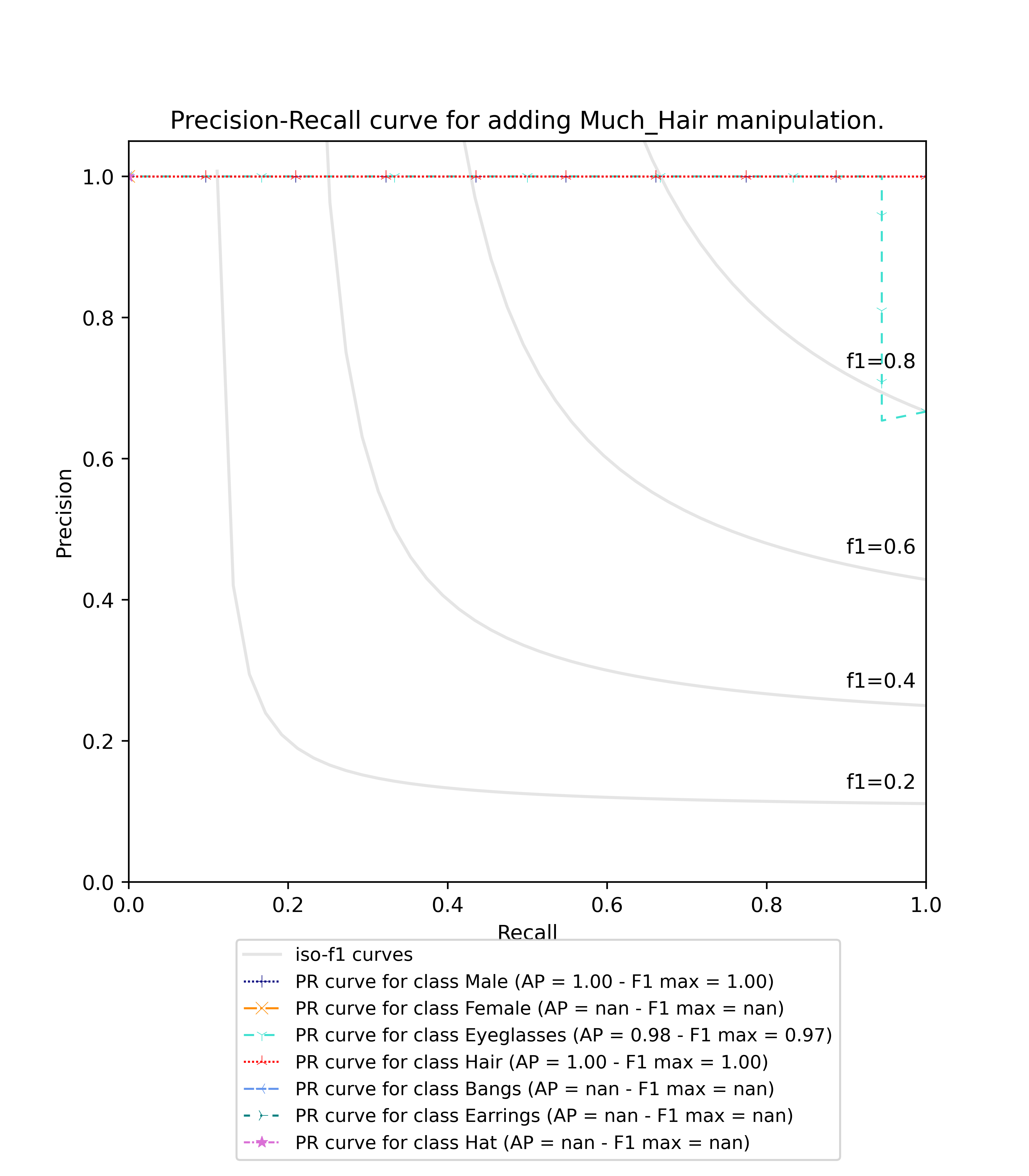} & \includegraphics[width=\textwidth]{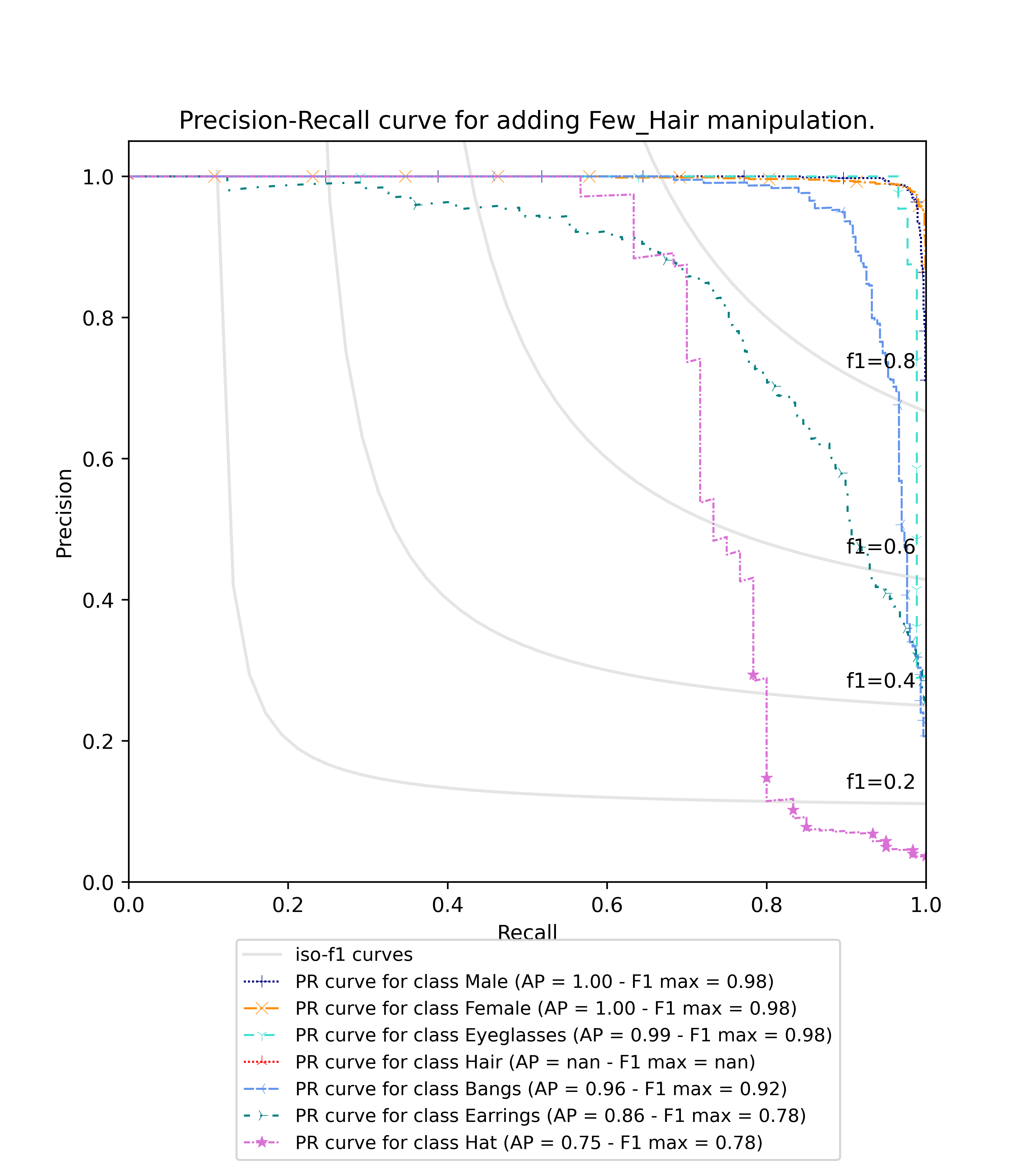} \\ \includegraphics[width=\textwidth]{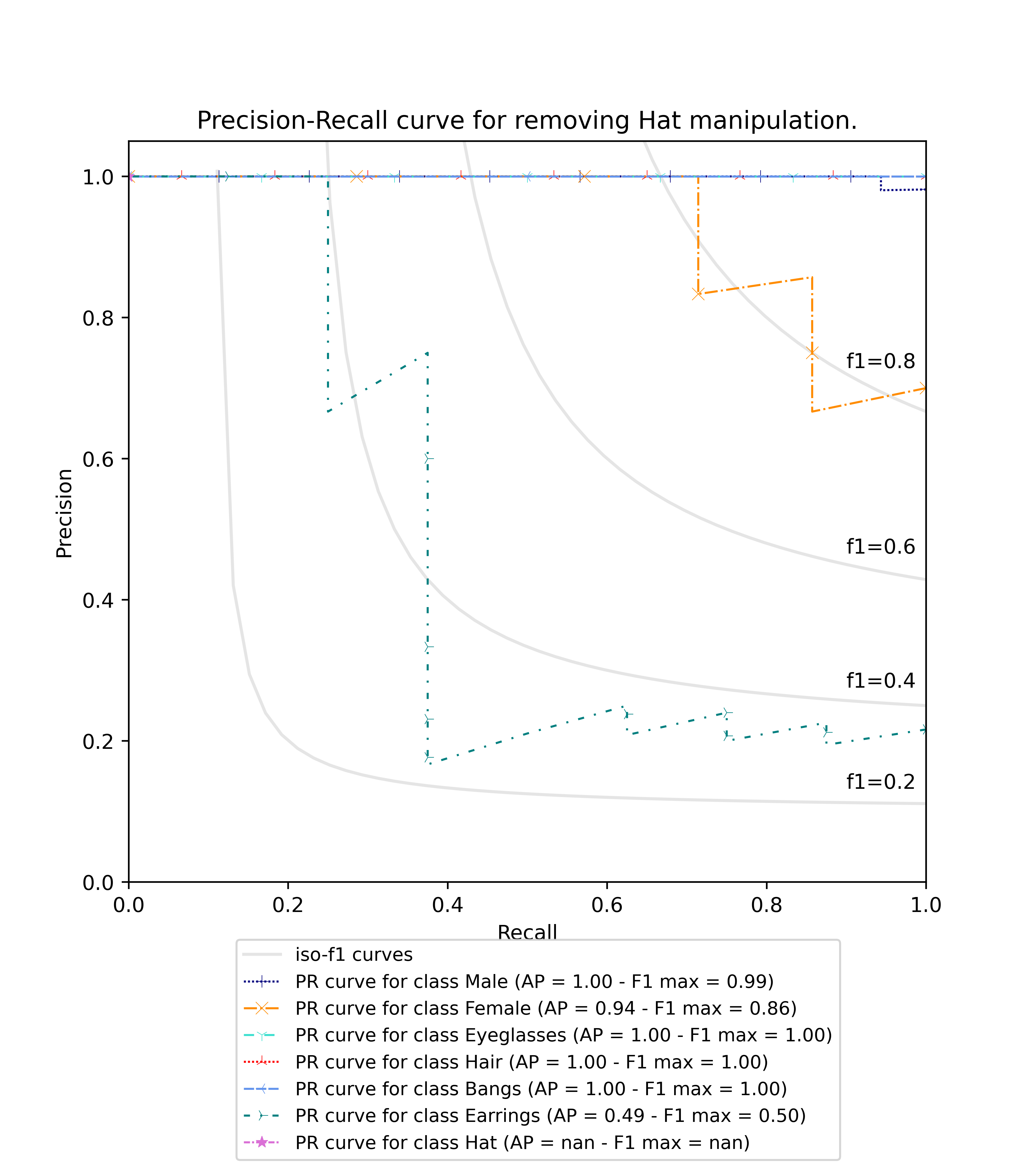} &
  \includegraphics[width=\textwidth]{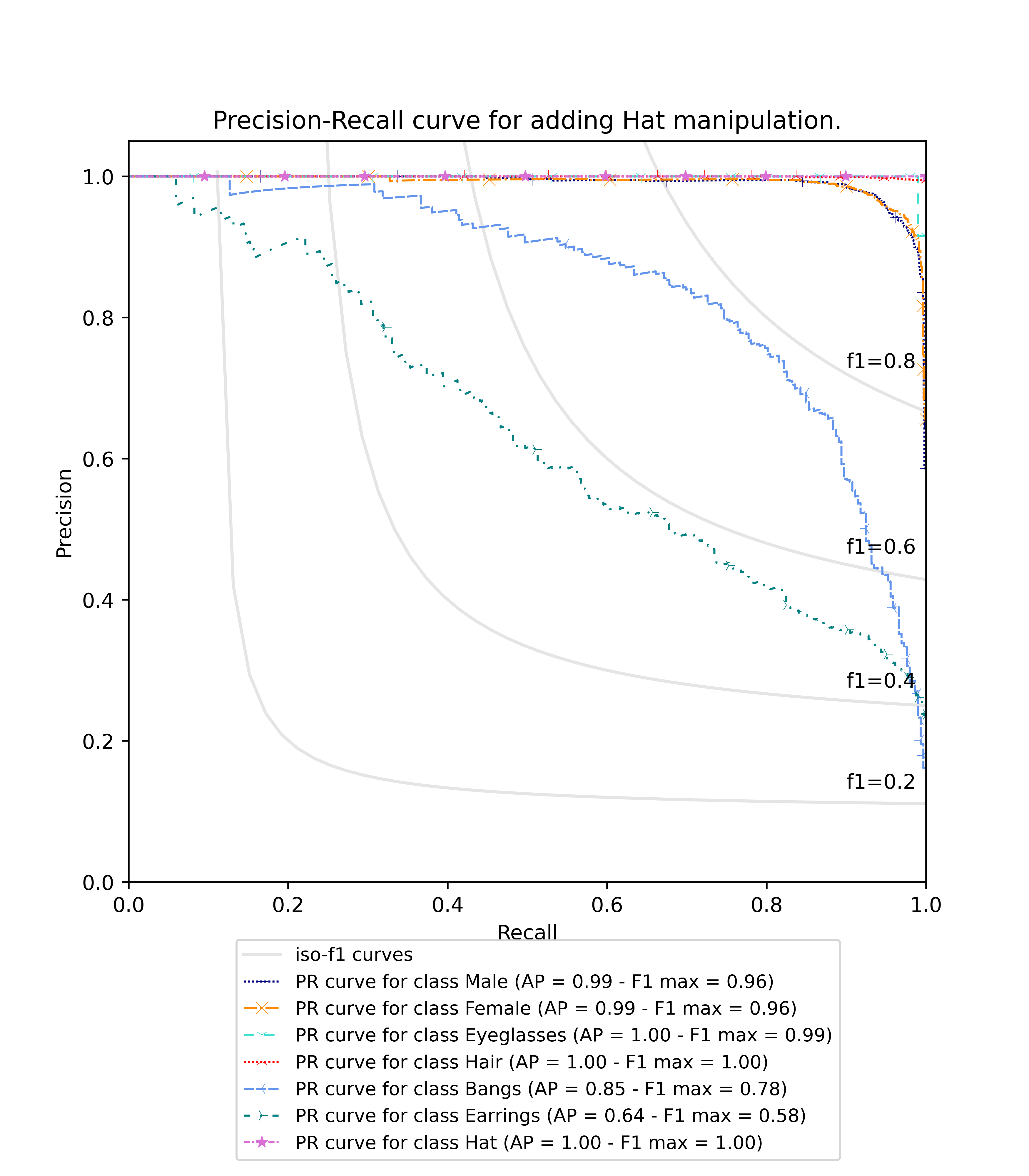} & \includegraphics[width=\textwidth]{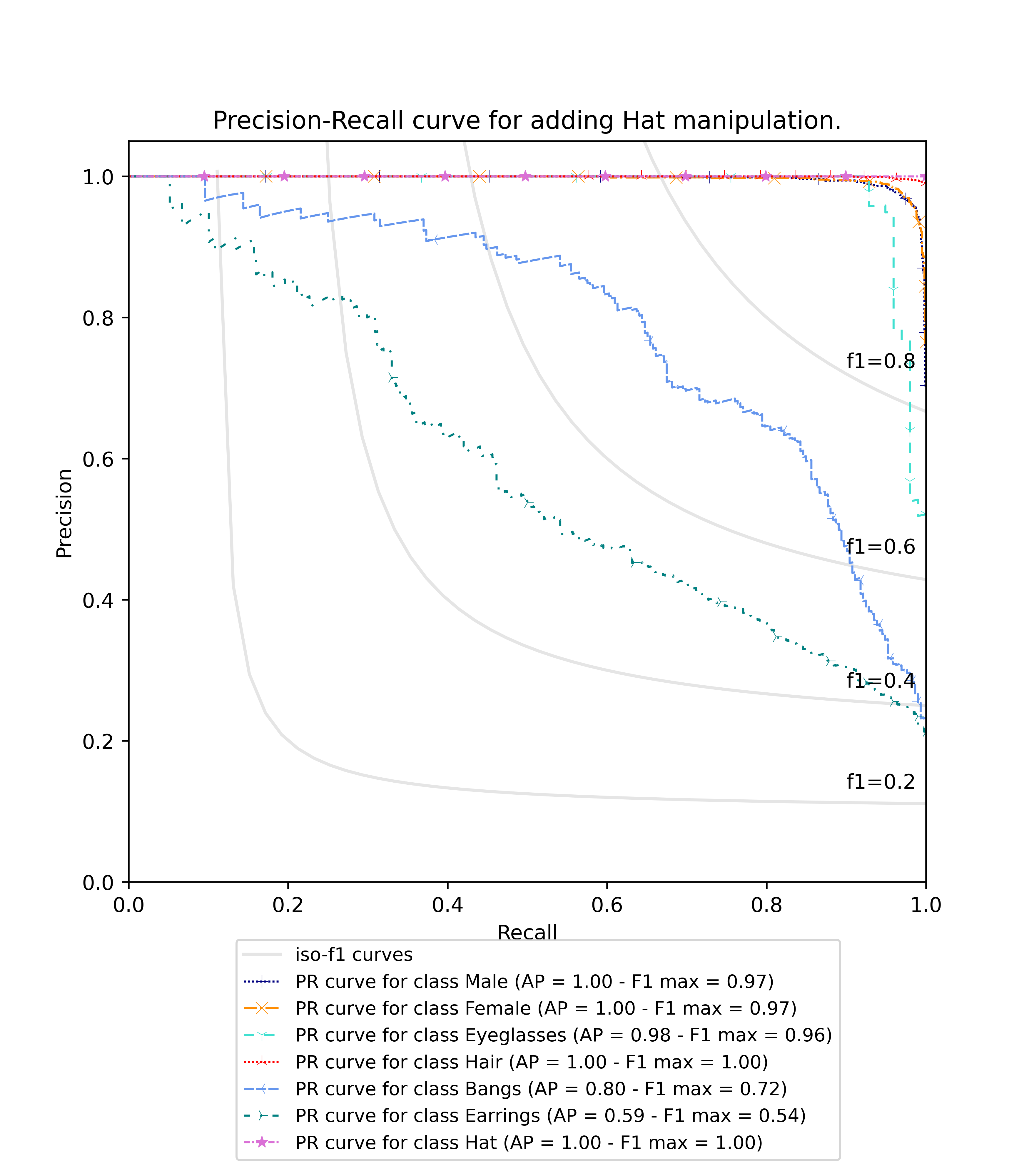} & \includegraphics[width=\textwidth]{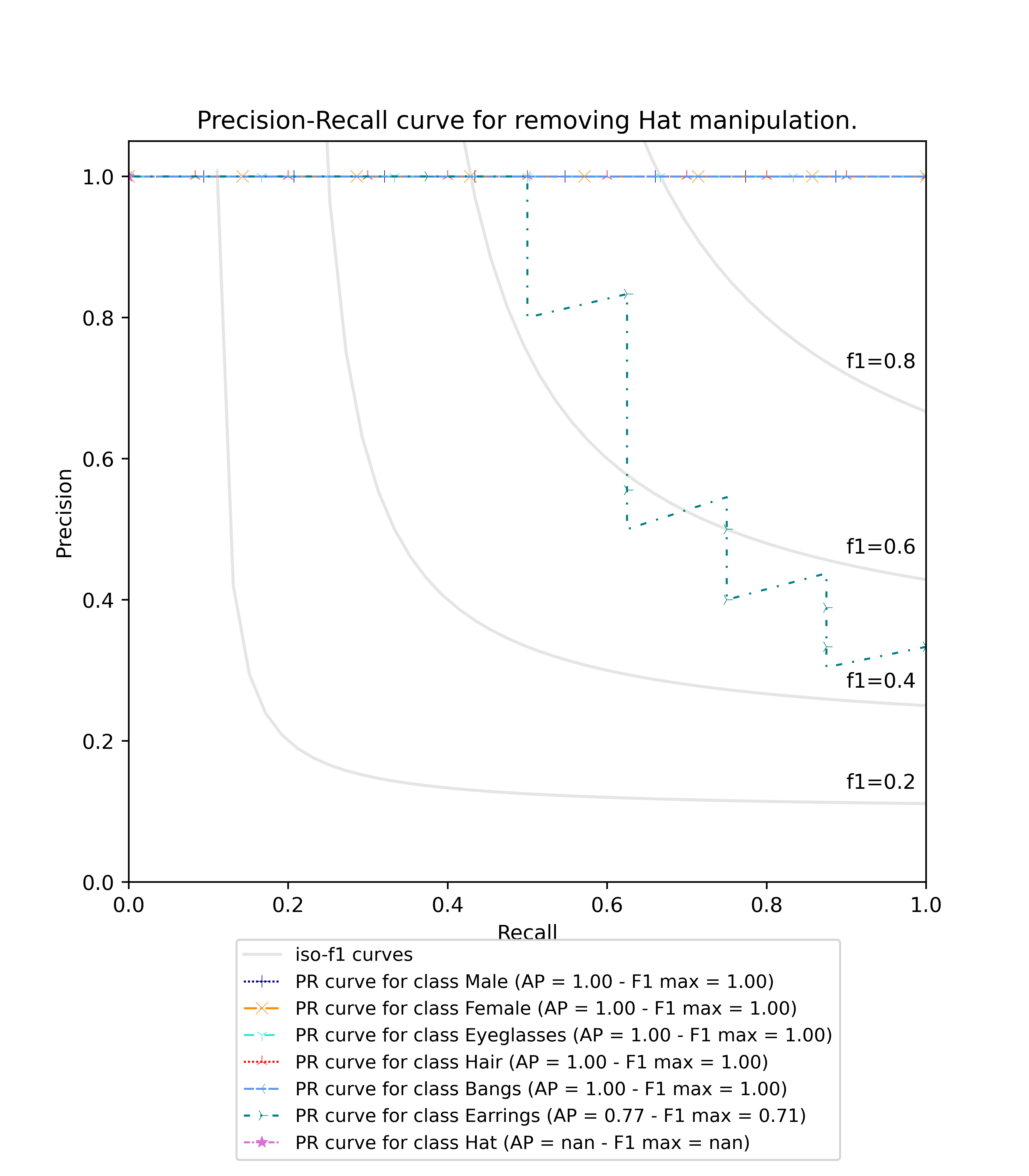} \\
  \includegraphics[width=\textwidth]{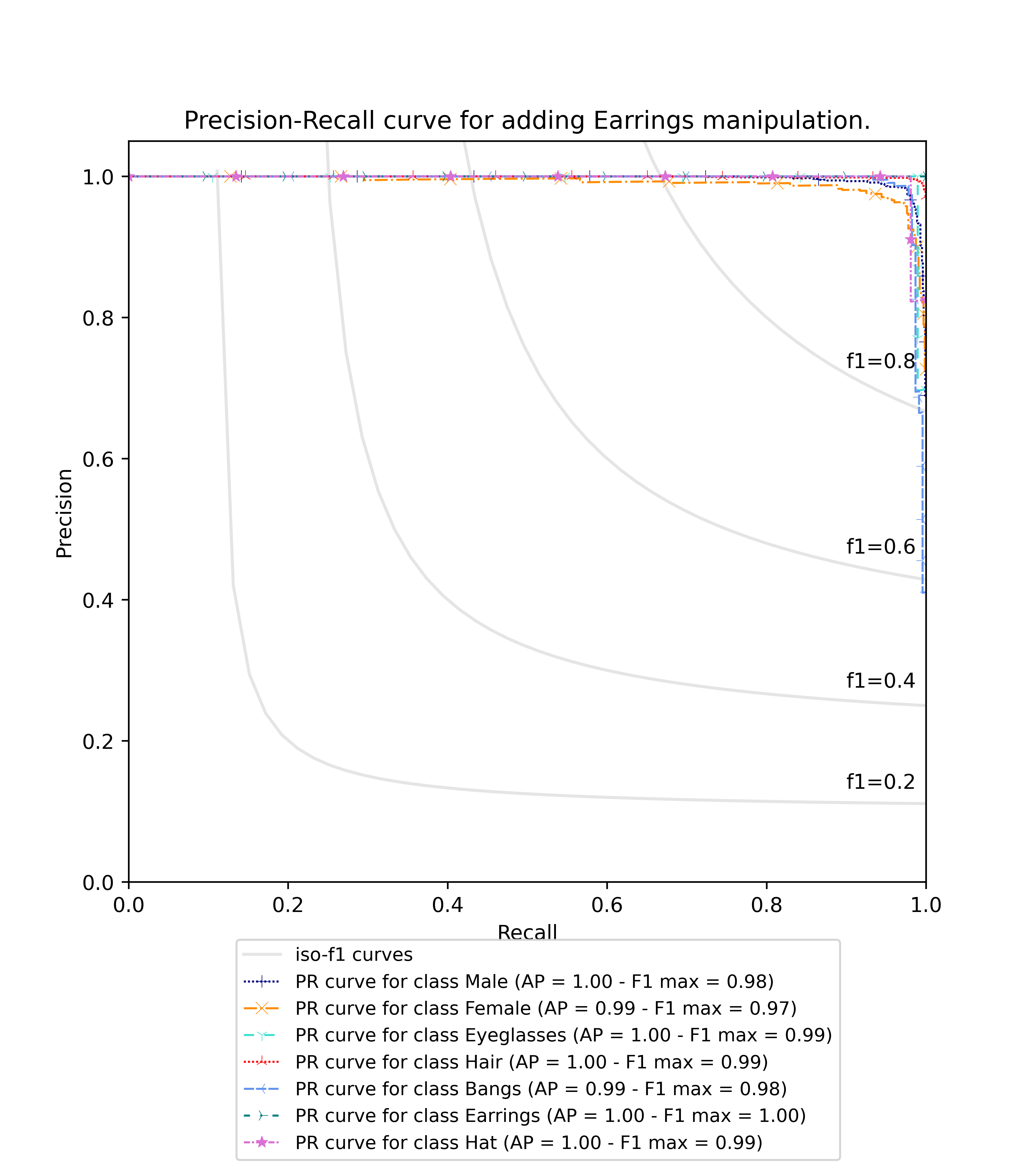} & \includegraphics[width=\textwidth]{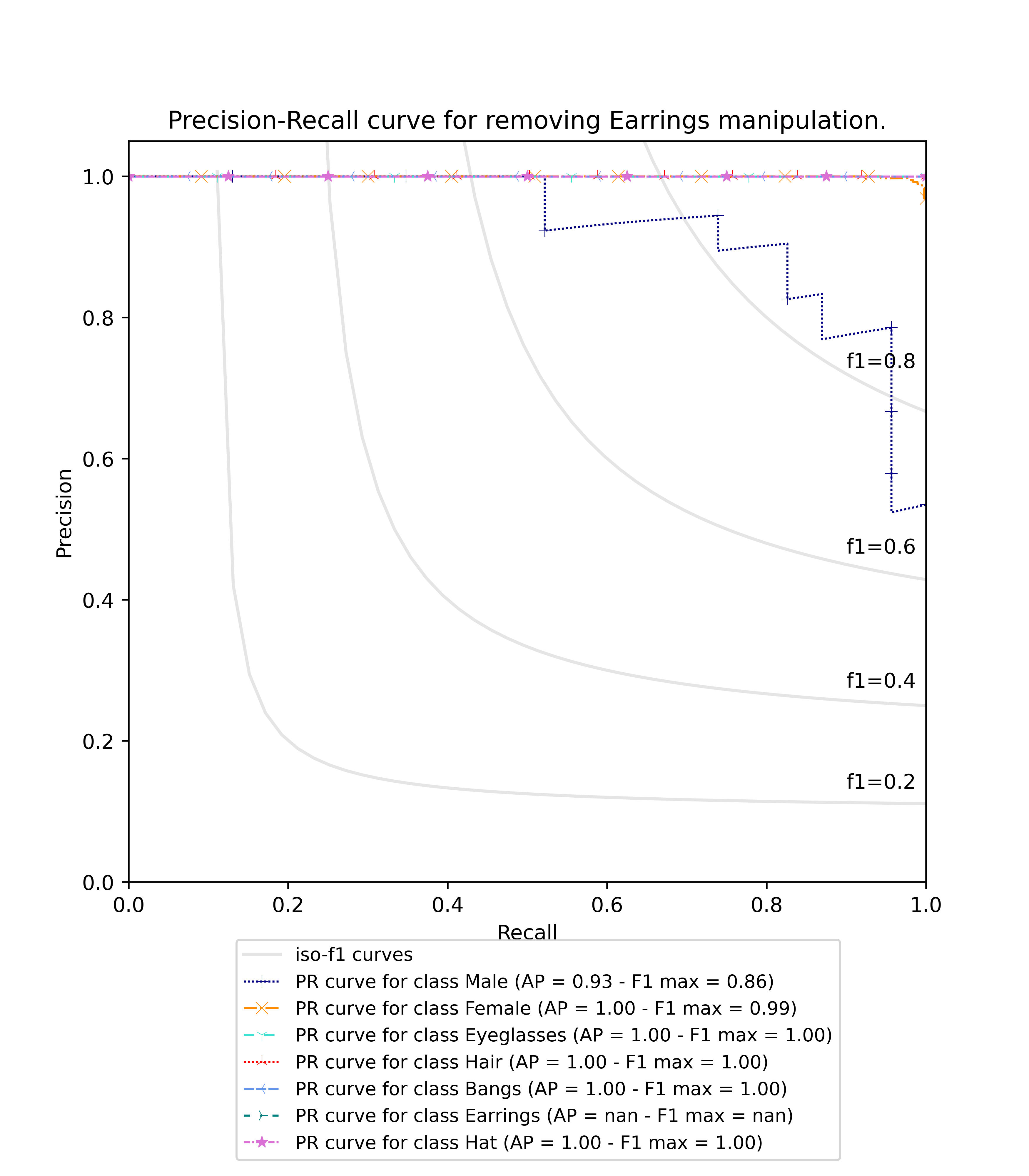} & \includegraphics[width=\textwidth]{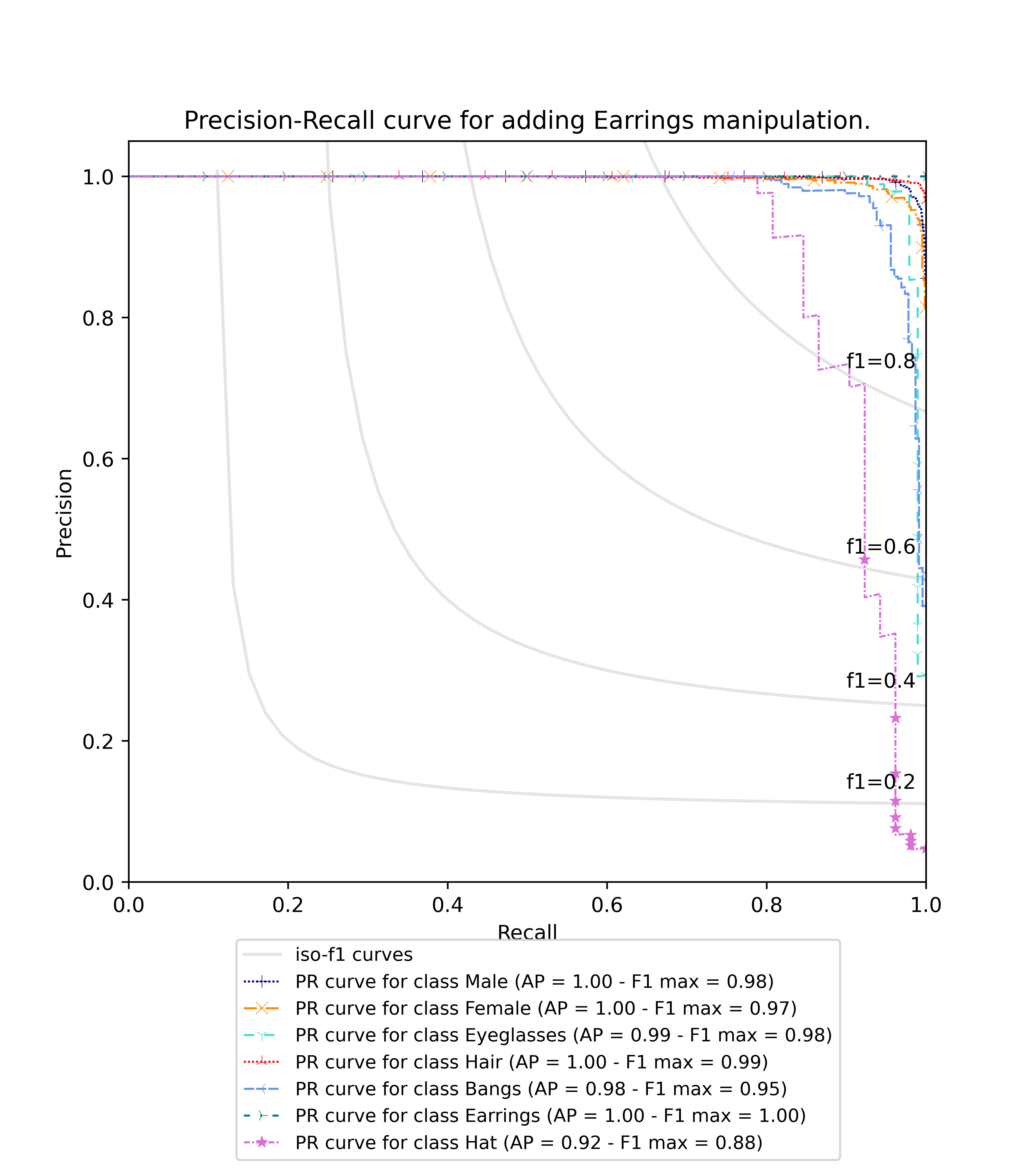} & \includegraphics[width=\textwidth]{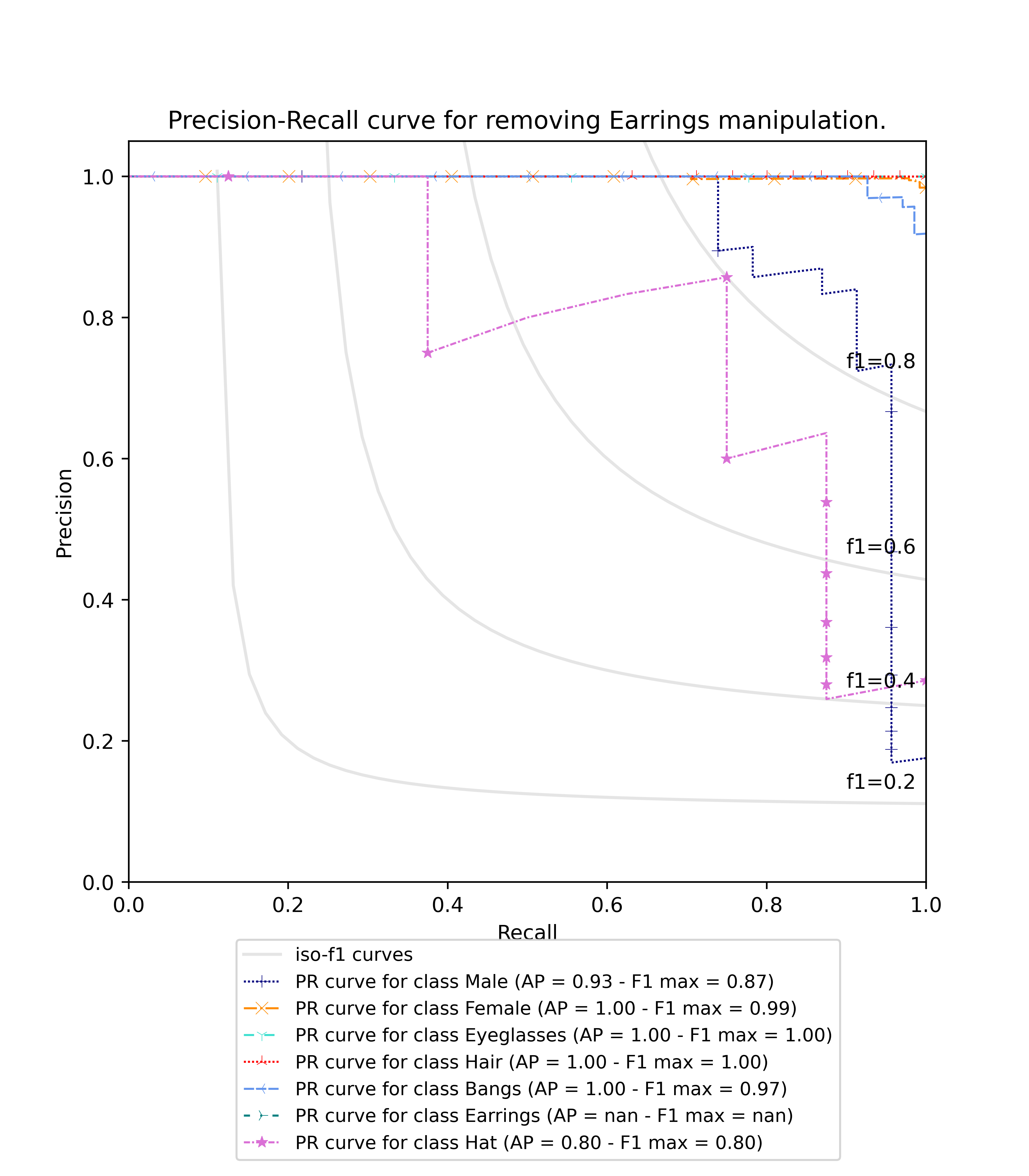} \\
  \hline
\end{tabular}
}
\end{center}
   \caption{\textbf{Precision vs Recall curves for each semantic manipulation for latent styles (left) and reference styles (right).} We study how the prediction of the remaining attributes is affected independently. Please zoom in for better details. Note that NaN in the legend of the images means the there were no images in the test set for the particular manipulation, \eg Much Hair (opposite of bald label) manipulation starts from the Bald label in the Test set and transform them to not Bald, and as only Males with no Earrings and no Hats have bald label in the CelebA-HQ dataset, hence NaN score for Female, Earrings and Hat.}
\label{figure:sem_pr2}
\end{figure*}

Furthermore, to study the disentanglement during the transformation, we compute Precision vs Recall curves for each attribute manipulation (see~\fref{figure:sem_pr1} and~\ref{figure:sem_pr2}).

\section{Additional Results for Image Synthesis}

\subsection{Semantic mask input size}
We study the influence of the mask size as starting point of the generator synthesis (see~\tref{table:mask_size}).
\label{appendix:additional_syn_mask}
\begin{table}[t]
\begin{center}
\begin{tabular}{|c||c|c|}
\hline
\multicolumn{3}{|c|}{FFHQ~\cite{karras2019style}} \\
\hline
Mask Size & FID$\downarrow$  & Diversity$\uparrow$ \\
\hline
4  & 13.40 & 0.42 $\pm$ 0.04 \\
8  & 12.10 & 0.41 $\pm$ 0.04 \\
16  & 12.56 & 0.42 $\pm$ 0.04 \\
32  & 15.12 & 0.42 $\pm$ 0.04 \\
\hline
\end{tabular}
\caption{\textbf{Quantitative assessment for the size of the mask during image synthesis.}}
\label{table:mask_size}
\end{center}
\end{table}

\section{Face Reenactment}
\label{appendix:reenactment}
\begin{figure*}[t]
\begin{center}
\resizebox{\linewidth}{!}{
\begin{tabular}{ccc}
    \includegraphics[width=0.33\textwidth]{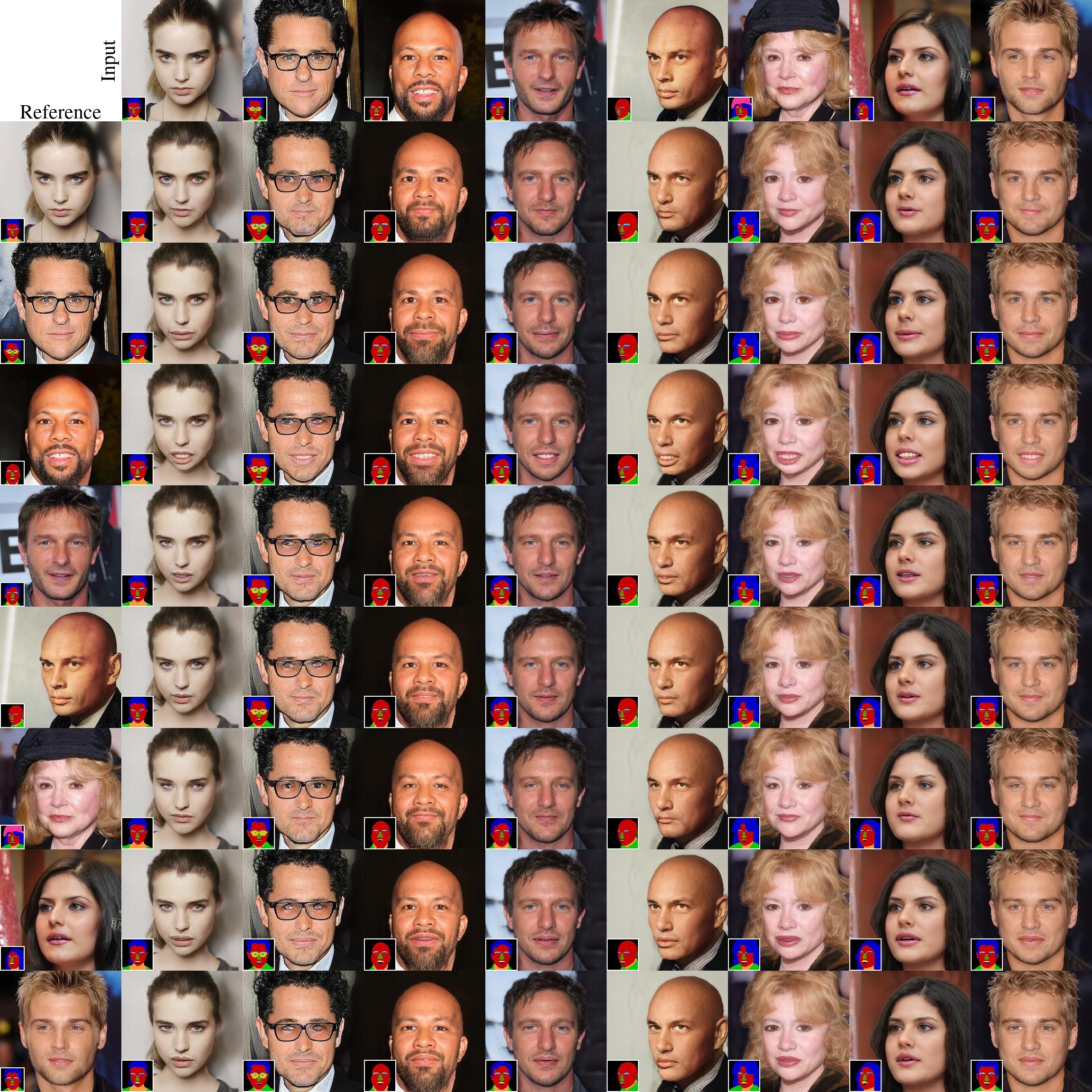} & 
    \includegraphics[width=0.33\textwidth]{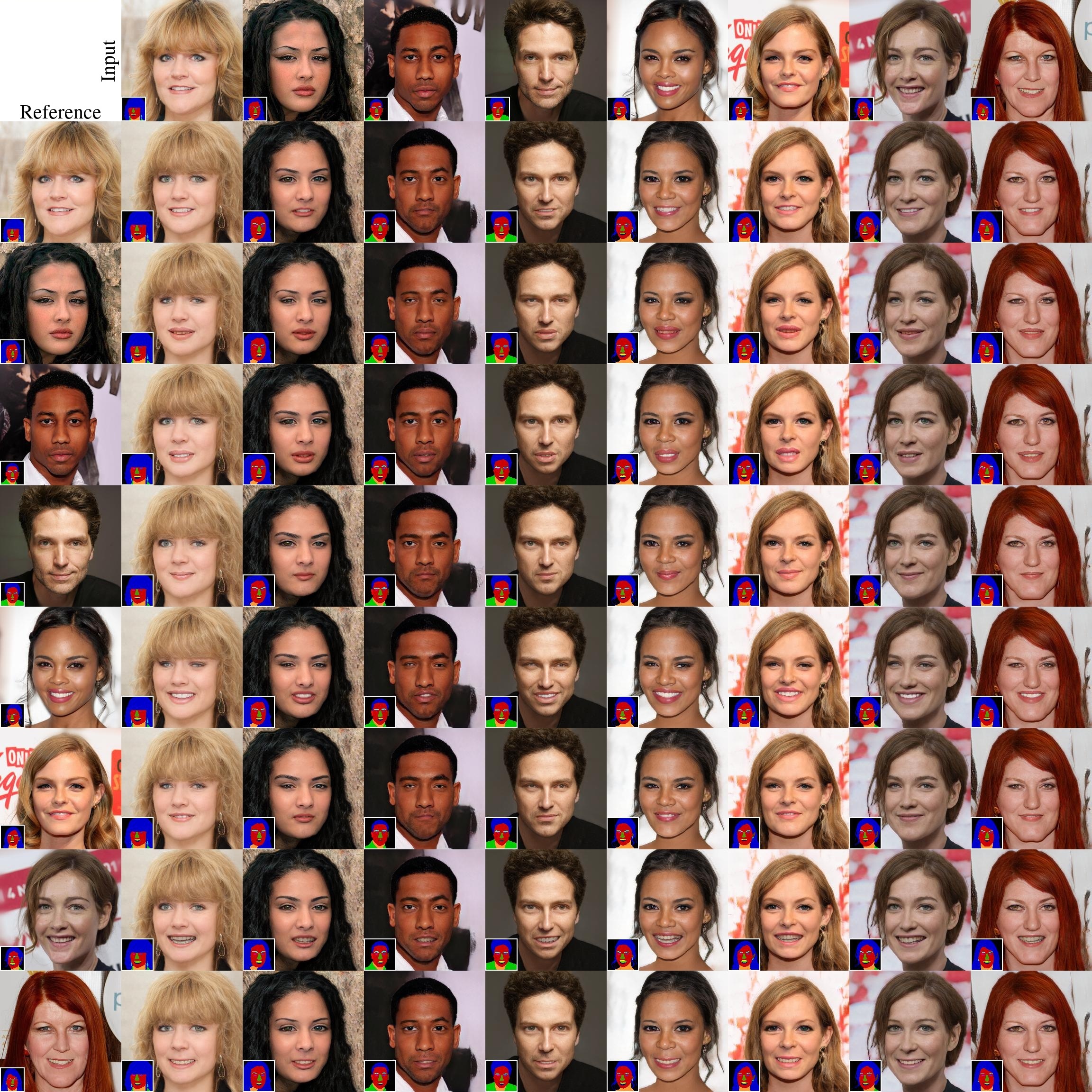} &
    \includegraphics[width=0.33\textwidth]{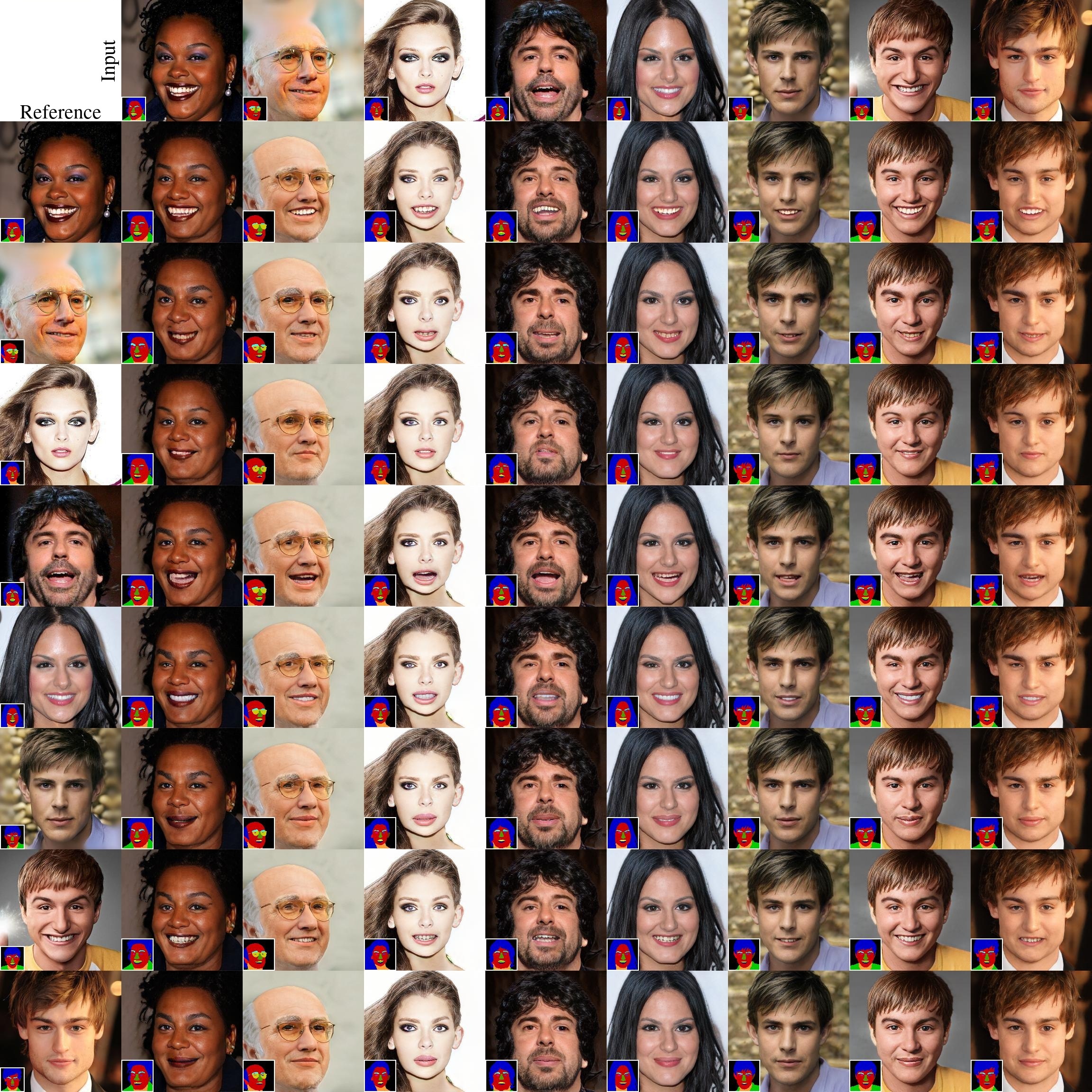} \\
\end{tabular}
}
\end{center}
\caption{\textbf{Face Reenactment Qualitative Results}. By only manipulation the semantic space and keeping the RGB information of the input, we can puppeteer the input with respect to a reference image. Zoom in for better details.}
\label{figure:reenactment}
\end{figure*}

Face Reenactment results are possible with a slight modification of our system. In this case, there are no domains, so we entirely rely on the transformation of certain regions of the face and keeping unaltered those regions do not related to the puppeteering. In~\fref{figure:reenactment} we show qualitative results.

\end{appendix}

\end{document}